\documentclass[10pt,twocolumn,letterpaper]{article}
\PassOptionsToPackage{dvipsnames,table,xcdraw}{xcolor}

\usepackage{cvpr}      
\definecolor{cvprblue}{rgb}{0.21,0.49,0.74}
\usepackage[pagebackref,breaklinks,colorlinks,allcolors=cvprblue]{hyperref}

\usepackage{microtype}
\usepackage{graphicx}
\usepackage{indentfirst}
\usepackage{booktabs} 
\usepackage{tabularx}
\usepackage{diagbox} 
\usepackage{overpic}
\usepackage{adjustbox}
\usepackage{pgfplots}
\usepackage{caption}
\usepackage{tikz}  
\usepackage{subcaption}

\usepackage{amsmath}
\usepackage{amssymb}
\usepackage{mathtools}
\usepackage{amsthm}
\usepackage{bm}  
\usepackage{multirow}

\def\paperID{9634} 
\def\confName{CVPR}
\def\confYear{2026}

\title{FreqAdapt: Frequency-Adaptive Processing for RAW Object Detection}

\author{
    Hanxi Li$^{1,2}$, Huiling Li$^3$\\
    $^1$University of Science and Technology China, $^2$Li Auto Inc, $^3$Hunan University\\
    {\tt\small lihanxi@mail.ustc.edu.cn, lhl@hnu.edu.cn}\\
}

\begin{document}
\maketitle
\begin{abstract}

Existing object detection methods predominantly utilize sRGB inputs, which are compressed from RAW sensor data using Image Signal Processors (ISP) originally designed for visualization purposes. Compared to RGB images, RAW images possess favorable noise characteristics and richer information representation, which are crucial for object detection, particularly under challenging conditions such as adverse weather or low-light environments. In this paper, we propose FreqAdapt, a lightweight module for adaptive RAW data enhancement in the frequency domain. Unlike traditional spatial domain processing methods, FreqAdapt innovatively maps ISP operations to the Fourier frequency domain and performs domain separation based on the physical properties of ISP operations, ensuring each operation is performed in its most suitable domain. Meanwhile, through an adaptive frequency domain encoder that jointly analyzes amplitude spectrum, phase spectrum, and RAW image features, we provide global context for ISP parameter prediction and employ a learnable fusion mechanism to achieve adaptive feature enhancement. Extensive experiments on multiple datasets with diverse lighting and weather conditions demonstrate that FreqAdapt achieves state-of-the-art performance while maintaining lightweight efficiency and good physical interpretability. Furthermore, our module can be seamlessly incorporated into existing object detection frameworks, providing a novel solution for visual perception tasks in the RAW domain. 

\end{abstract}    
\section{Introduction}
\label{sec:intro}

\begin{figure}[t]
    \centering
    \begin{subfigure}[b]{0.96\linewidth}
        \centering
        \includegraphics[width=\linewidth]{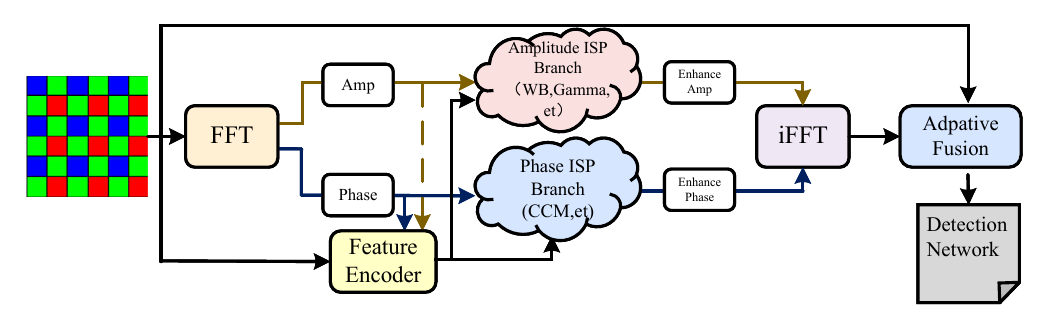}
        \caption{A simplified illustration of FreqAdapt.}
        \label{Freqintro:a}
    \end{subfigure}
    \begin{subfigure}[b]{0.96\linewidth}
        \centering
        \includegraphics[width=\linewidth]{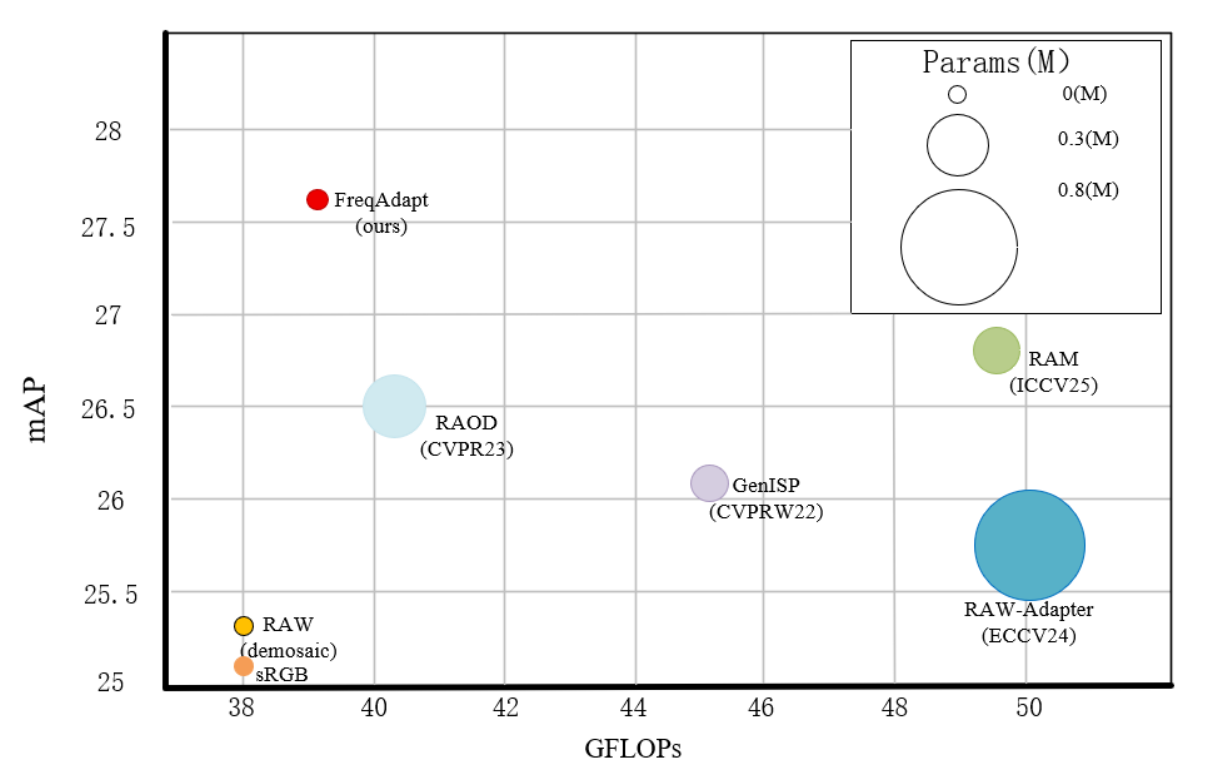}
        \caption{Performance comparison of different method.}
        \label{Freqintro:b}
    \end{subfigure}
    \caption{(a) Illustration of the core framework of our proposed method, highlighting the frequency-domain processing pipeline and domain separation strategy. (b) Comparison of detection performance (mAP) and computational efficiency (parameters) with existing RAW processing methods.}
    \label{Freqintro}
    \vspace{-4mm}
\end{figure}

Most computer vision tasks today are trained and evaluated on standard RGB (sRGB) images, largely due to the widespread availability of public datasets such as COCO \cite{lin2014microsoft} and ImageNet \cite{deng2009imagenet}. The manually designed Image Signal Processing (ISP) pipeline converts high bit-depth RAW sensor data into 8-bit sRGB images, aiming to produce images that offer superior quality for human visual perception \cite{wu2019visionisp,karaimer2016software}. It typically consists of a sequence of operations such as demosaicing, white balance adjustment, color correction, tone mapping, denoising, sharpening, and gamma correction \cite{ramanath2005color}. However, certain steps within the ISP may introduce artifacts or degradation in image quality \cite{guo2024learning}, which can adversely impact the performance of downstream high-level vision tasks. 
For instance, demosaicing can introduce color artifacts and blur fine patterns, tone mapping compresses the dynamic range causing loss of details in shadows and highlights, and aggressive sharpening can amplify noise and create halo artifacts around edges. 
These limitations become particularly pronounced under challenging conditions such as low-light environments or adverse weather, where preserving the original sensor information is crucial for robust object detection.

Recent research has demonstrated the advantages of utilizing RAW images for computer vision tasks \cite{guo2025dark,wang2024adaptiveisp,xie2025simrod}. Unlike processed RGB images, RAW data preserves the complete sensor information with higher bit-depth, wider dynamic range, and physically meaningful noise characteristics \cite{wei2020physics,wei2021physics,jin2023lighting}. This rich information representation has proven beneficial not only for low-level vision tasks such as super-resolution and denoising \cite{kerepecky2021d3net,lecouat2022high,liang2020raw}, but increasingly for high-level tasks including object detection and segmentation \cite{chen2023instance,maxwell2024logarithmic,xu2023toward,yoshimura2023dynamicisp}. By directly processing RAW data, systems can potentially reduce complexity, lower latency, and cut computational costs—critical advantages for real-time applications such as autonomous driving \cite{li2024efficient}.


However, existing RAW-based object detection methods face several limitations. Most approaches require complex model architectures and training procedures that cannot be easily integrated as plug-and-play modules\cite{cui2025raw,wang2024adaptiveisp}. They employ computationally expensive parameter searching algorithms and multi-stage training strategies\cite{morawski2022genisp,yoshimura2023dynamicisp} , while some methods require additional reference data such as normal-light RGB images or camera metadata during training \cite{cui2021multitask,zhang2024isp}. While these methods demonstrate promising results on benchmark datasets, they tend to be computationally expensive and introduce unnecessary design complexities, making them impractical for real-world deployment where efficiency and simplicity are crucial.






In this paper, we present FreqAdapt, a novel frequency-domain adaptive processing module for RAW image enhancement. Our key insight is that different ISP operations exhibit distinct mathematical properties in the frequency domain that can be leveraged for more efficient and effective processing. Specifically, the Fourier transform \cite{brigham1988fast,oppenheim2005importance} decomposes images into two orthogonal components: the amplitude spectrum encoding energy distribution and contrast information \cite{xu2020learning,chen2021amplitude}, and the phase spectrum preserving structural and edge information \cite{schwartz2018deepisp}. This decomposition naturally aligns with the physical characteristics of ISP operations: as illustrated in \cref{Freqintro}, intensity-related operations such as gamma correction and white balance can be formulated as multiplicative transformations on the amplitude spectrum while preserving phase information intact; whereas structure-related operations like Color Correction Matrix (CCM) and detail enhancement primarily modify inter-channel relationships, manifesting as phase adjustments in the frequency domain \cite{he2024enhancing,chen2021amplitude}.

Through this physics-inspired domain separation strategy, we avoid the inherent information loss in spatial domain processing. Meanwhile, we design an adaptive frequency encoder that jointly analyzes amplitude spectrum, phase spectrum, and raw image statistics to provide global context for ISP parameter prediction, and employ a learnable fusion mechanism to achieve adaptive feature enhancement. Extensive experiments demonstrate that FreqAdapt achieves state-of-the-art performance on multiple challenging datasets while maintaining efficiency. Furthermore, FreqAdapt can be seamlessly integrated as a plug-and-play module into existing detection frameworks, providing a principled yet practical solution for RAW-based visual perception tasks.
 To summarize, our main contributions are as follows:
\begin{itemize}

\item  We propose FreqAdapt, a lightweight and novel frequency-domain adaptive processing module that performs domain separation based on the mathematical properties of ISP operations in Fourier space, achieving efficient parallel processing while maintaining physical interpretability.
\item  We design an adaptive parameter prediction mechanism that jointly analyzes amplitude spectrum, phase spectrum, and RAW statistical features to dynamically adjust ISP operation parameters, ensuring robust performance under diverse lighting conditions and scene types.
\item  We demonstrate superior performance with fewer parameters than existing methods on multiple RAW detection benchmark datasets.

\end{itemize}

\section{Related Work}
\label{sec:related}

\subsection{Image Signal Processor}

Image Signal Processing (ISP) transforms RAW sensor data into visually appealing RGB images through a series of operations. Traditional ISP pipelines typically consist of sequential steps including demosaicing, white balance, color correction, tone mapping, denoising, sharpening, and gamma correction \cite{delbracio2021mobile,karaimer2016software,nishimura2018automatic,ramanath2005color}. While these manually designed components produce high-quality images for human perception \cite{dai2020awnet,ignatov2019aim}, the sequential processing paradigm can introduce cumulative errors and information loss \cite{gamrian2025beyond,guo2024learning}, which adversely affects downstream vision tasks.
Recent advances in deep learning have led to various approaches to enhance or replace traditional ISPs \cite{kim2023paramisp,schwartz2018deepisp,xing2021invertible,chen2018learning}, primarily following two main directions. The first direction focuses on RAW-to-RGB mapping, where end-to-end learning methods directly map RAW to RGB images using neural networks \cite{schwartz2018deepisp, ignatov2020replacing, zhang2021learning}, trained on paired RAW-RGB datasets to generate visually appealing images that align with human perception. The second emerging research direction focuses on directly leveraging RAW data to enhance downstream vision tasks, where differentiable ISP modules enable joint optimization with task-specific objectives \cite{tseng2019hyperparameter,qin2023learning,yu2021reconfigisp}. However, such deep network models are often large and computationally intensive, which limits their practicality in real-world deployment, particularly on resource-constrained devices.

\begin{figure*}[t]
    \centering
    \centerline{\includegraphics[width=0.96\linewidth]{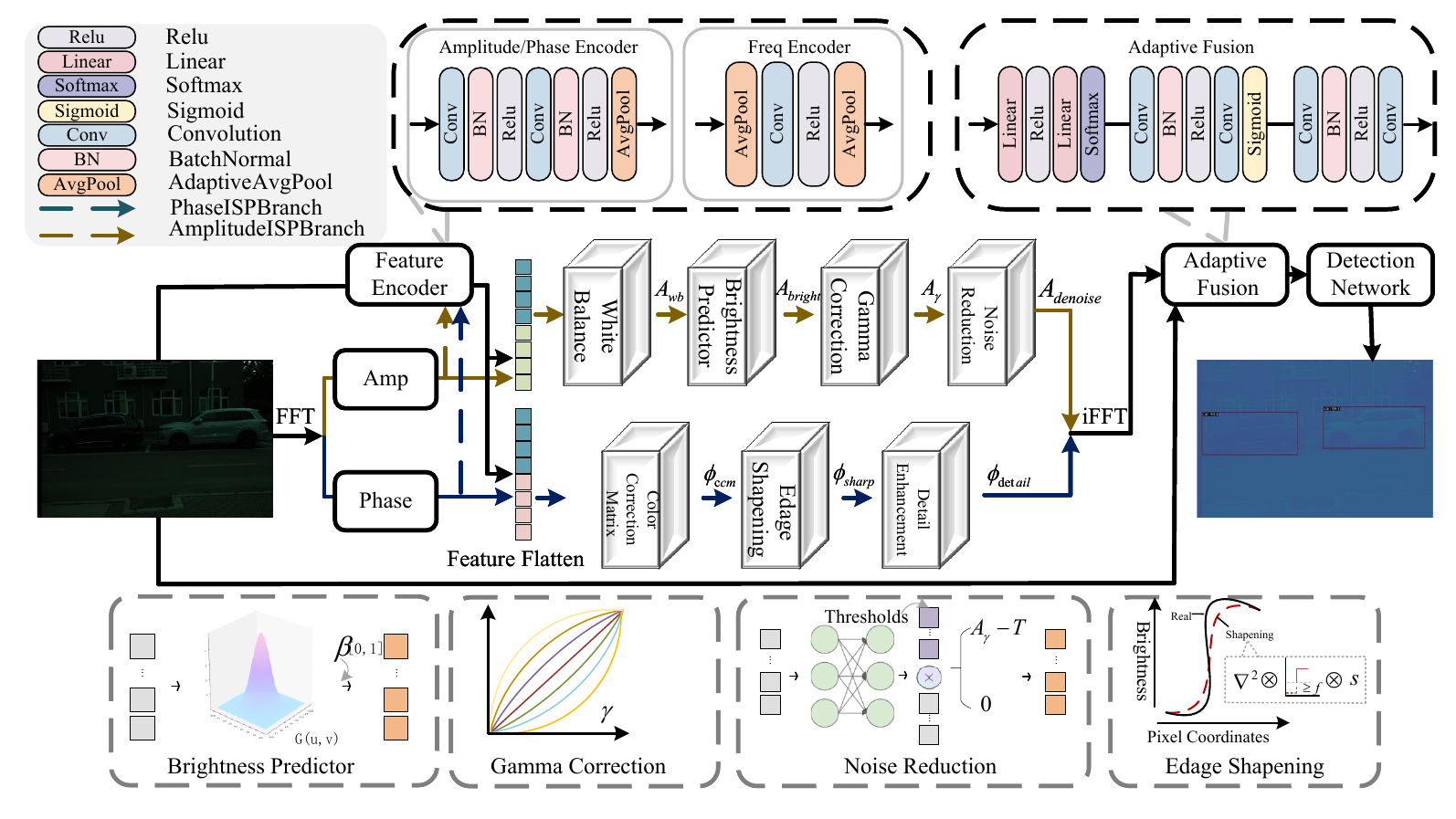}}
    \caption{Architecture of FreqAdapt. Our framework decomposes RAW images into amplitude and phase spectra via FFT, processes them through domain-specific ISP branches, and reconstructs the enhanced image via IFFT. The Frequency RAW Encoder jointly analyzes amplitude spectrum, phase spectrum, and RAW image features to provide global context for adaptive parameter prediction in subsequent ISP operations. The Amplitude ISP Branch handles intensity operations (white balance, brightness, gamma, noise reduction), while the Phase ISP Branch manages structural operations (color correction, sharpening, detail enhancement). The Adaptive Fusion Module combines frequency-enhanced and original images. }
    \label{freqadapt-framework}
    \vspace{-4mm}
\end{figure*}

\subsection{RAW Object Detection}

RAW images have demonstrated significant potential for improving object detection performance beyond sRGB, particularly under challenging conditions such as low-light and adverse weather \cite{cui2025raw,guo2025dark,li2025towards,gamrian2025beyond}. Unlike processed RGB images, RAW data preserves complete sensor information with higher bit-depth, wider dynamic range, and physically meaningful noise characteristics.
Existing RAW object detection methods can be broadly divided into three categories. First, early works perform vision tasks directly on RAW images \cite{9035647} without modeling camera noise, which limits Performance. Second, ISP parameter retuning adapts conventional pipelines to pre-trained sRGB detectors by adjusting hyper-parameters via evolutionary algorithms \cite{mosleh2020hardware} or gradient-based optimization \cite{yu2021reconfigisp}; recent methods such as DynamicISP \cite{yoshimura2023dynamicisp} and AdaptiveISP \cite{wang2024adaptiveisp} further realize per-image dynamic control to handle varying conditions. Third, end-to-end approaches jointly optimize the ISP and the detector under the detection loss \cite{buckler2017reconfiguring}, with neural networks generating task-specific ISP parameters, as in GenISP \cite{morawski2022genisp} and RAW-Adapter \cite{cui2025raw}.


Despite progress, these solutions largely retain sequential, spatial-domain processing, which can cause cumulative information loss and often entails complex training and high computational cost. Our FreqAdapt addresses these limitations by operating in the frequency domain, aligning ISP functions with amplitude (intensity) and phase (structure) components. This design mitigates the information loss inherent in spatial-domain processing, while remaining lightweight and plug-and-play for seamless integration with existing detection frameworks, delivering superior performance with practical efficiency.

\section{Method}
\label{sec:method}

We propose a Fourier Adaptive Module that separates amplitude and phase components through Fourier transform, executing different ISP functions in appropriate domains based on their characteristics to achieve adaptive RAW image enhancement. This section first introduces the fundamentals of Fourier transform and our design motivation, then details the network architecture.

\subsection{Fourier Transform and Motivation}

The Fourier transform converts images from the spatial domain to the frequency domain, providing a novel perspective for image processing. For an input image $\mathbf{I} \in \mathbb{R}^{C \times H \times W}$, the 2D discrete Fourier transform is defined as:
\begin{equation}
\mathcal{F}(\mathbf{I})(u,v) = \sum_{x=0}^{H-1} \sum_{y=0}^{W-1} \mathbf{I}(x,y) \cdot e^{-2\pi i(\frac{ux}{H} + \frac{vy}{W})}
\end{equation}

The Fourier transform decomposes the image into amplitude spectrum $\mathbf{A}$ and phase spectrum $\mathbf{\Phi}$:
\begin{equation}
\mathcal{F}(\mathbf{I}) = \mathbf{A} \cdot e^{i\mathbf{\Phi}}
\end{equation}
where $\mathbf{A} = |\mathcal{F}(\mathbf{I})|$ represents the amplitude and $\mathbf{\Phi} = \angle\mathcal{F}(\mathbf{I})$ represents the phase.
The amplitude spectrum primarily contains brightness, contrast, and texture information, while the phase spectrum encodes structural patterns, edges, and spatial relationships.  This decomposition allows us to design specialized processing branches for each component, applying ISP operations where they are most effective.  


Based on these characteristics, our motivations for designing the Fourier Adaptive Module include:





 Frequency-domain correspondence of ISP functions: Different ISP operations have clear correspondences in the frequency domain. White balance and gamma correction primarily affect the amplitude spectrum; sharpening and detail enhancement mainly adjust high-frequency phase components; noise predominantly exists in the high-frequency portion of the amplitude spectrum. Operating in frequency domain enables direct and precise control over specific frequency bands through simple multiplicative masks. This correspondence enables us to design targeted processing strategies, ensuring each operation is applied in its most suitable domain.

Global receptive field: The Fourier transform inherently provides a global receptive field, with each frequency component containing information from the entire image. In contrast, spatial domain convolutions require stacking multiple layers to achieve large receptive fields with complexity $\mathcal{O}(k^2 \cdot H \cdot W)$, while frequency domain operations only require $\mathcal{O}(H \cdot W)$ after the initial FFT transformation.

 Computational efficiency: Many operations that require iterative computation in the spatial domain can be implemented through simple element-wise operations in the frequency domain. For example, convolution corresponds to multiplication in the frequency domain, and global color adjustments that would require processing every pixel in spatial domain become simple channel-wise operations in frequency domain, significantly reducing computational complexity.

\subsection{Network Framework}

Based on the above analysis, we design the network framework shown in \cref{freqadapt-framework}, consisting of four main components: Frequency RAW Encoder, Amplitude ISP Branch, Phase ISP Branch, and Adaptive Fusion Module.




\subsubsection{Frequency RAW Encoder}

The Frequency RAW Encoder serves as the feature extraction backbone that analyzes the input from multiple perspectives to guide adaptive ISP parameter prediction. Its primary purpose is to understand the global characteristics of the image—such as overall brightness distribution, noise level, color cast, and structural complexity—which are essential for determining appropriate enhancement strategies.

The encoder processes three complementary inputs: the amplitude spectrum capturing brightness and contrast information, the phase spectrum encoding structural patterns and edges, and the raw image providing spatial context. 
These multi-modal features are combined through convolutional networks and adaptive pooling to generate a global feature vector that serves as a conditioning signal for all ISP operations. This design ensures each ISP function receives appropriate information—white balance utilizes amplitude-based color distribution while sharpening leverages phase-based edge information—enabling coordinated, image-specific parameter prediction across all processing branches.

\subsubsection{Amplitude ISP Branch}

The Amplitude ISP Branch specializes in brightness-related enhancements. We process the amplitude spectrum through a cascade of ISP operations, each targeting specific image characteristics.


\textbf{White Balance.} We correct color cast by applying channel-wise scaling factors to the amplitude spectrum:
\begin{equation}
\mathbf{A}_{wb} = \mathbf{W} \odot \mathbf{A} , \quad \text{where} \quad \mathbf{W} = \text{diag}(w_r, w_g, w_b)
\end{equation}
The scaling factors $[w_r, w_g, w_b]$ are constrained to prevent extreme color shifts, where $\mathbf{W} \in \mathbb{R}^{3 \times 3}$ is a diagonal matrix with channel-specific gains.




\textbf{Brightness Enhancement.} Global brightness is adjusted by modulating low-frequency components:
\begin{equation}
\mathbf{A}_{bright} = \mathbf{A}_{wb} \cdot (1 + \beta \cdot \mathbf{M}_{low})
\end{equation}
where $\mathbf{M}_{low}(u,v) = \exp(-\alpha(u^2 + v^2))$ is a Gaussian mask centered at zero frequency, and $\beta \in [0,1]$ is the predicted brightness factor.

\textbf{Gamma Correction.} We apply adaptive non-linear transformation for contrast enhancement:
\begin{equation}
\mathbf{A}_{\gamma} = \mathbf{A}_{bright}^{\gamma}
\end{equation}



\textbf{Noise Reduction.} We employ frequency-adaptive thresholding to suppress noise while preserving details:
\begin{equation}
\mathbf{A}_{denoise}(u,v) = \max(\mathbf{A}_{\gamma}(u,v) - T(u,v), 0)
\end{equation}
The threshold $T(u,v)$ varies across frequency bands:
\begin{equation}
T(u,v) = T_{low} \cdot \mathbf{M}_{low} + T_{mid} \cdot \mathbf{M}_{mid} + T_{high} \cdot \mathbf{M}_{high}
\end{equation}
where $[T_{low}, T_{mid}, T_{high}]$ are adaptively predicted based on noise level estimation.

Finally, a learned modulation network refines the processed amplitude spectrum:
\begin{equation}
\mathbf{A}_{final} = \mathbf{A}_{denoise} \odot \mathbf{M}_{amp}
\end{equation}
where $\mathbf{M}_{amp}$ is an attention mask learned to selectively enhance or suppress different frequency components based on the denoised amplitude patterns. 



\subsubsection{Phase ISP Branch}

The Phase ISP Branch manipulates phase spectrum to enhance structural details and color relationships, leveraging the fact that phase encodes edge and texture information.

\textbf{Color Correction Matrix.} We approximate color transformation by modulating phase relationships:
\begin{equation}
\mathbf{\Phi}_{ccm} = \mathbf{\Phi} + \alpha_{ccm} \cdot \text{Reshape}(\mathbf{C} \cdot \text{Flatten}(\mathbf{\Phi}))
\end{equation}
where $\mathbf{C} \in \mathbb{R}^{3 \times 3}$ is the predicted correction matrix with $\alpha_{ccm}$ controlling the strength to maintain stability.



\textbf{Edge Sharpening.} High-frequency phase components are selectively enhanced to sharpen edges:
\begin{equation}
\mathbf{\Phi}_{sharp} = \mathbf{\Phi}_{ccm} + s \cdot \mathbf{M}_{high} \odot \nabla^2(\mathbf{\Phi}_{ccm})
\end{equation}
where $\nabla^2$ denotes the Laplacian operator for edge detection, $\mathbf{M}_{high}$ is a high-frequency mask, and $s$ is the adaptive sharpening strength.




\textbf{Detail Enhancement.} Fine texture details in mid-to-high frequencies are enhanced through adaptive phase modulation:
\begin{equation}
\mathbf{\Phi}_{detail} = \mathbf{\Phi}_{sharp} + \alpha \cdot \mathbf{M}_{detail} \odot (\mathcal{F}_{detail}(\mathbf{\Phi}_{sharp}))
\end{equation}
where $\mathcal{F}_{detail}$ is a shallow CNN for texture enhancement, $\alpha$ controls enhancement strength, and $\mathbf{M}_{detail}$ is a band-pass mask targeting texture frequencies.


Phase values are regularized to ensure validity:
\begin{equation}
\mathbf{\Phi}_{final} = \text{Wrap}(\mathbf{\Phi}_{detail}, [-\pi, \pi])
\end{equation}

\subsubsection{Adaptive Fusion Module}

The Adaptive Fusion Module intelligently combines the frequency-enhanced result with the original input to preserve naturalness while incorporating enhancements.

First, we reconstruct the enhanced image via inverse FFT:
\begin{equation}
\mathbf{I}_{freq} = \text{IFFT}(\mathbf{A}_{final} \odot e^{i\mathbf{\Phi}_{final}})
\end{equation}


The final output is obtained through learned adaptive fusion:
\begin{equation}
\mathbf{I}_{output} = \mathbf{I}_{fused} \odot \mathbf{M}_{att} + \mathbf{I}_{raw} \odot (1 - \mathbf{M}_{att})
\end{equation}
where $\mathbf{I}_{fused} = w_{freq} \cdot \mathbf{I}_{freq} + w_{raw} \cdot \mathbf{I}_{raw}$ with adaptive weights $[w_{freq}, w_{raw}]$, and $\mathbf{M}_{att} = \sigma(\text{CNN}_{att}([\mathbf{I}_{freq}, \mathbf{I}_{raw}]))$ is a spatial attention map that selectively preserves enhancements.

The entire framework is end-to-end differentiable and trained with detection loss, enabling joint optimization of frequency-domain ISP operations and downstream detection performance. Further details on the model architecture are available in  the supplementary material.

\section{Experiments}
\label{sec:exp}
\begin{table*}[!ht]
    \centering
    \caption{Performance comparison across different datasets. Results are
reported using mean Average Precision (mAP) and mAP at 50\% IoU (mAP$_{50}$). The highest result is highlighted in bold, while the secondhighest is marked with an underline.
}
    \resizebox{0.95\textwidth}{!}{%
    \begin{tabular}{lcccccccccc}\toprule
    
    \multirow{2}{*}{Method} & \multicolumn{2}{c}{LOD-Dark} & \multicolumn{2}{c}{LOD-Normal} & \multicolumn{2}{c}{NOD-Nikon} & \multicolumn{2}{c}{NOD-Sony} & \multicolumn{2}{c}{AROD}\\
    & mAP & mAP$_{50}$ & mAP & mAP$_{50}$ & mAP & mAP$_{50}$ & mAP & mAP$_{50}$ & mAP & mAP$_{50}$\\\midrule
    
    RAW & 24.4& 47.2& 31.4& 58.2& 26.3& 49.7& 25.8& 50.6& 18.7& 28.6\\
    sRGB & 24.6& 48.3& 31.7& 58.7& 26.6& 51.5& 25.5& 50.3& 18.4& 27.9\\
    DynamicISP \cite{yoshimura2023dynamicisp} & 26.5& 50.5& 33.2& 61.6& 27.8& 53.2& 26.7& 52.6& 19.1& 29.7\\
    GenISP \cite{morawski2022genisp} & 25.1& 48.4& 34.0& \underline{62.5}& 27.4& 53.5& 26.4& 51,8& 19.3& 30.8\\
    RAOD \cite{xu2023toward} & 26.3& 50.8& 34.5& 62.1& 28.4& 53.7& \underline{27.1}& \textbf{53.9}& 19.6& 31.4\\
        RAW-Adapter \cite{cui2025raw} & 26.4& \underline{51.7}& 32.8& 59.9& 28.2& 53.9& 26.1& 53.5& -& -\\
    RAM \cite{gamrian2025beyond} & \underline{27.0}& 51.2& \underline{34.6}& \underline{62.5}& \underline{28.7}& \underline{54.0}& 26.2& 53.2& \underline{20.6}& \underline{32.9}\\
 FreqAdapt-T& 26.5& 50.5& 32.4& 57.6& 28.1& 53.1& 26.6& 50.9& 20.3&32.1\\
    \textbf{FreqAdapt} & \textbf{27.2}& \textbf{52.6}& \textbf{35.1}&\textbf{62.7}& \textbf{29.4}& \textbf{54.8}&\textbf{ 27.3}& \underline{53.7}& \textbf{21.3}& \textbf{33.8}\\
    
    \bottomrule
 & & & & & & & & & &\\
    \end{tabular}%
    }
    \label{tab:performance_comparison1}
\end{table*}

\subsection{Datasets}

Extensive experiments were conducted on multiple RAW object detection datasets to evaluate and compare the proposed method with existing approaches.

\textbf{LOD:} The LOD dataset \cite{hong2021crafting} contains 2,230 14-bit lowlight RAW images categorized into eight object classes. This dataset is designed for detecting multiple categories of objects in low-light indoor and outdoor environments. It includes long (LOD-Normal) and short (LOD-Dark) exposure images of the same scenes. The object classes are: car, motorcycle, bicycle, chair, dining table, bottle, TV, and bus.

\textbf{NOD:} The NOD dataset \cite{morawski2022genisp} includes 14-bit RAW outdoor images taken under low-light conditions. It includes 7,200 images, with 3,200 captured by the Sony RX100 VII (NOD-Sony) and 4,000 by the Nikon D750 (NOD-Nikon).The dataset is annotated with bounding boxes for 46,000 instances of people, bicycle, and car classes.

\textbf{AROD:} The AROD dataset \cite{li2025towards} consists of 7,785 high-resolution RAW images with 135,601 annotated instances across 62 object categories. The dataset captures diverse indoor and outdoor scenes under 9 distinct lighting and weather conditions, including normal light, low-light, overexposure, fog, rain, and snow. Each RAW image is provided with its corresponding sRGB version, enabling evaluation of both RAW and sRGB detection methods under challenging environmental conditions.

\subsection{Implementation Details}

To better demonstrate the advantages of RAW in extreme scenarios and align with practical applications, we downsampled all dataset inputs to approximately 2K resolution. To validate the effectiveness of the proposed method across different detectors and backbones, we employed a two-stage Faster R-CNN detector \cite{ren2015faster} with ResNet50 \cite{he2016deep} as the backbone for the LOD and NOD datasets. For the AROD dataset, we utilized a two-stage Cascade R-CNN \cite{cai2018cascade} detector  with ConvNeXt-T \cite{liu2022convnet} as the backbone. Computational efficiency evaluation experiments employed a single-stage RetinaNet detector \cite{lin2017focal}. Our implementation is based on MMDetection \cite{chen2019mmdetection}, with all models trained from scratch and evaluated on NVIDIA A100 GPUs. Additional details regarding experimental settings can be found in the supplementary material.


\subsection{Comparison with State-of-the-Art Methods}

We conduct comprehensive experimental comparisons between FreqAdapt and various state-of-the-art methods in terms of both performance and efficiency. The baseline methods include RGB images converted from RAW using the open-source RawPy \cite{riechert2014raw} library with default ISP settings, and RAW images processed through simple bilinear interpolation for demosaicing. We compare against several recent RAW processing methods: DynamicISP \cite{yoshimura2023dynamicisp}, which dynamically adjusts ISP parameters based on input characteristics; GenISP \cite{morawski2022genisp} and RAOD \cite{xu2023toward}, which generate task-specific ISP parameters using convolutional neural networks; RAW-Adapter \cite{cui2025raw}, which adapts RAW images to the RGB domain through learnable adapter modules; and RAM \cite{gamrian2025beyond}, which processes ISP functions in parallel rather than sequentially. Additionally, we introduce FreqAdapt-T (Tiny), a lightweight variant of FreqAdapt with reduced parameters (detailed architecture specifications are provided in the supplementary material). All methods were evaluated using identical data augmentation strategies and training configurations to ensure fair comparison.





\subsubsection{Performance Analysis}


As shown in \cref{tab:performance_comparison1}, FreqAdapt achieves state-of-the-art performance across all evaluated datasets. On the challenging LOD-Dark dataset, our method attains 27.2 mAP and 52.6 mAP$_{50}$, surpassing the second-best method RAM by 0.2\% and 1.4\% respectively. The improvement is more pronounced on LOD-Normal (35.1 mAP, +0.5\% over RAM), demonstrating the robustness of our approach under varying illumination conditions. 
Cross-camera generalization is validated through NOD dataset evaluation, where FreqAdapt achieves 29.4 mAP on NOD-Nikon (+0.7\% over RAM) and 27.3 mAP on NOD-Sony (+1.1\% over RAM). This consistent improvement across different camera manufacturers confirms our frequency-domain processing's capability to handle diverse sensor characteristics and noise patterns inherent to various imaging systems. On the challenging AROD dataset, FreqAdapt achieves 21.3 mAP and 33.8 mAP$_{50}$, exceeding RAM by 0.7\% and 0.9\% respectively, demonstrating the effectiveness of our method in adapting RAW data for downstream detection tasks across extreme scenarios.

The consistent improvements across all datasets validate our key insights regarding frequency-domain processing advantages. Through amplitude-phase decoupling, FreqAdapt performs targeted enhancements by separately processing amplitude and phase components, avoiding interference between different image characteristics. Unlike fixed ISP pipelines or learning-based methods with limited adaptability, our frequency-domain approach enables fine-grained, content-aware adjustments that better preserve task-relevant information. Furthermore, the Frequency RAW Encoder efficiently captures global frequency characteristics, providing superior feature extraction compared to spatial-domain processing while maintaining computational efficiency.

\begin{figure*}[h]
    \centering

    \begin{minipage}[t]{0.01\textwidth}
        \centering
        \rotatebox{90}{\footnotesize LOD-Dark}
    \end{minipage}
    \begin{minipage}[t]{0.135\textwidth}
        \centering
        \includegraphics[width=\textwidth]{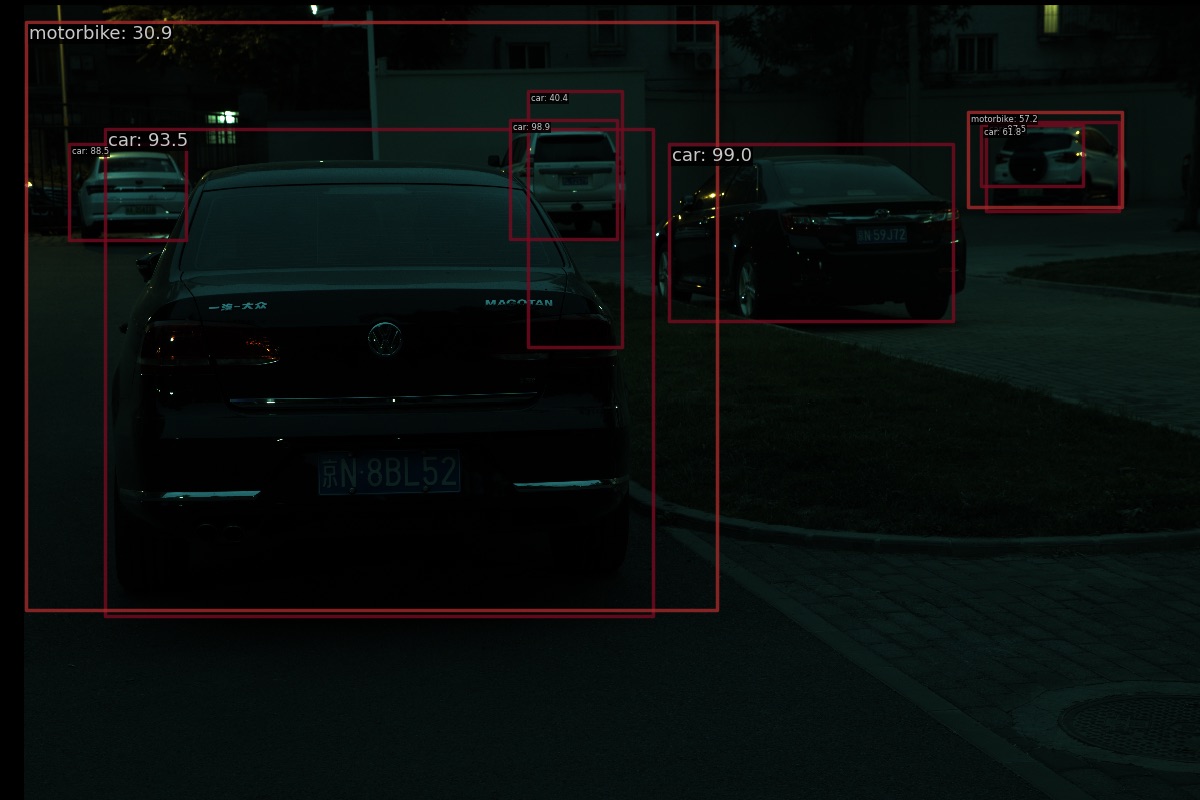}
    \end{minipage}
    \begin{minipage}[t]{0.135\textwidth}
        \centering
        \includegraphics[width=\textwidth]{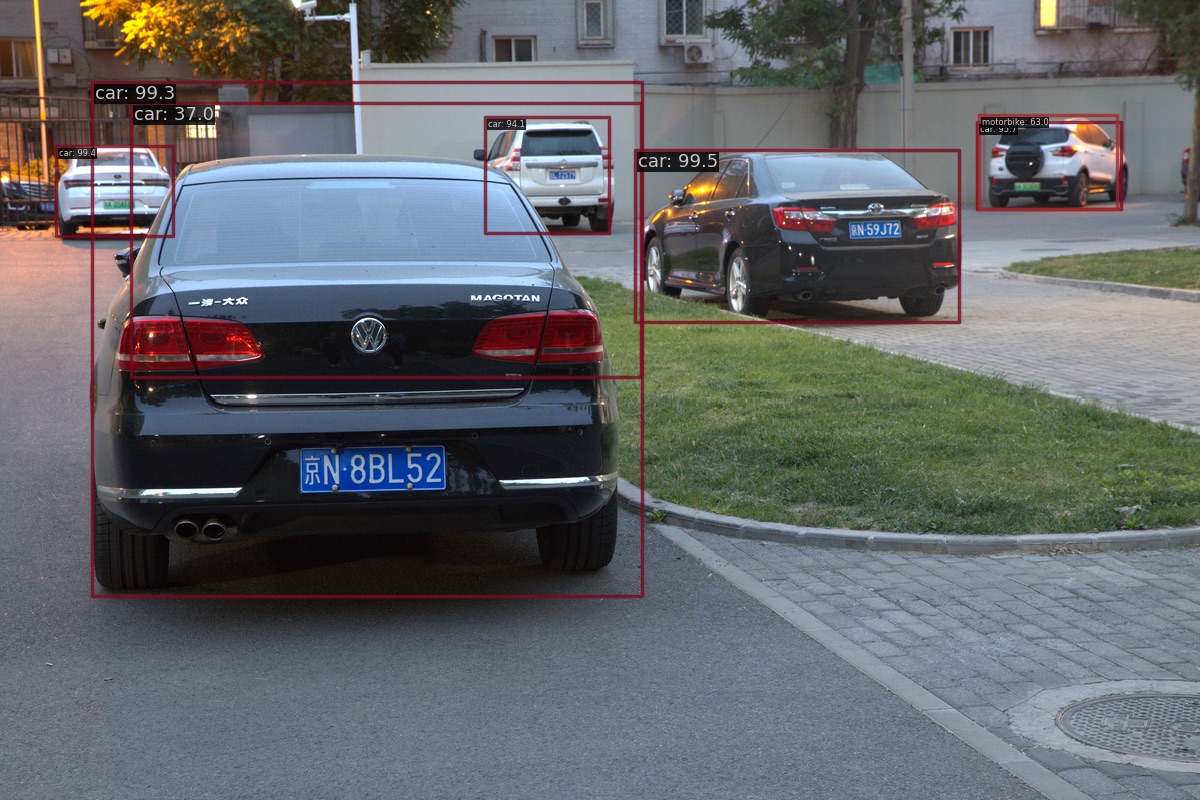}
    \end{minipage}
    \begin{minipage}[t]{0.135\textwidth}
        \centering
        \includegraphics[width=\textwidth]{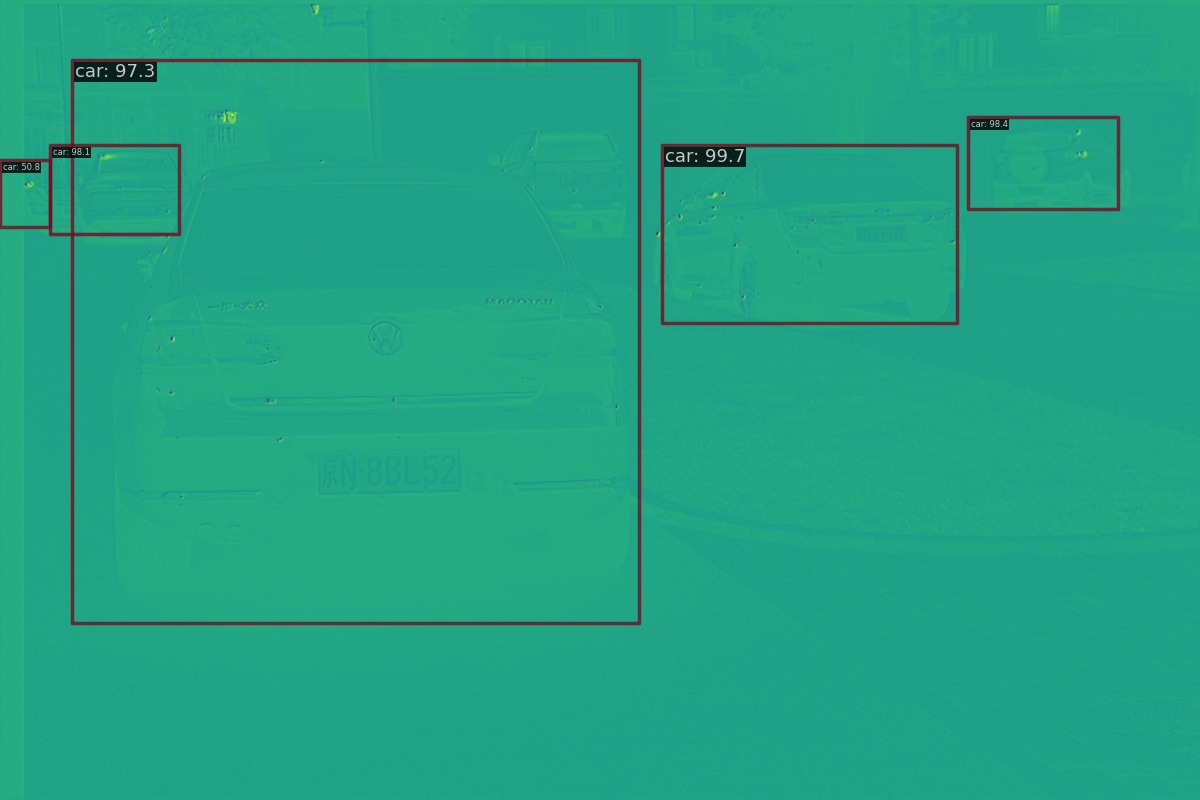}
    \end{minipage}
    \begin{minipage}[t]{0.135\textwidth}
        \centering
        \includegraphics[width=\textwidth]{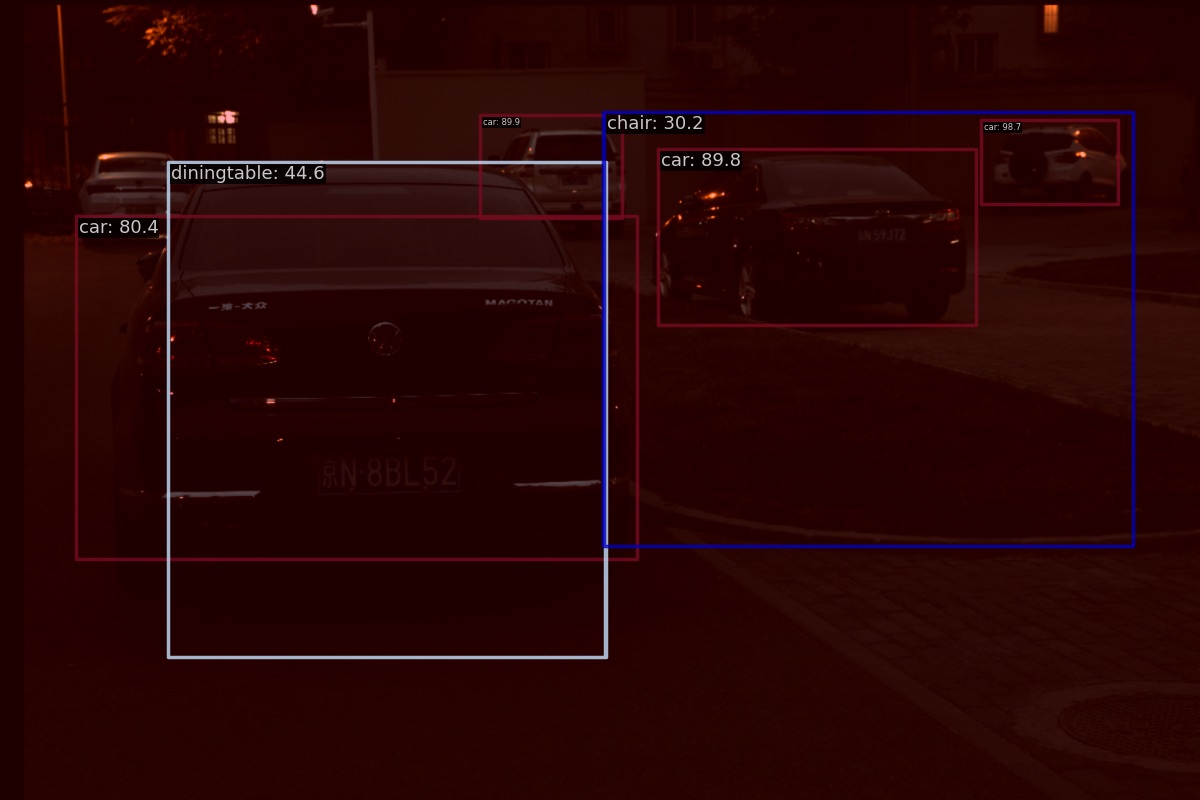}
    \end{minipage}
    \begin{minipage}[t]{0.135\textwidth}
        \centering
        \includegraphics[width=\textwidth]{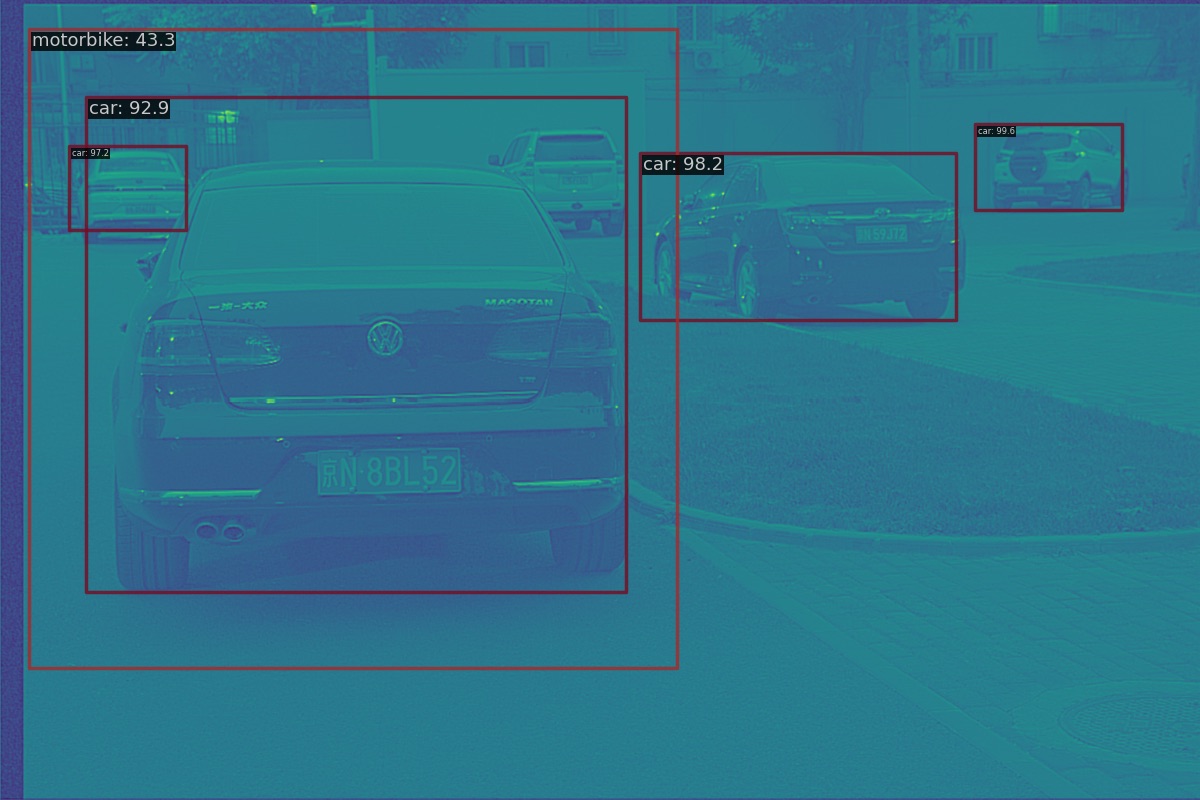}
    \end{minipage}
    \begin{minipage}[t]{0.135\textwidth}
        \centering
        \includegraphics[width=\textwidth]{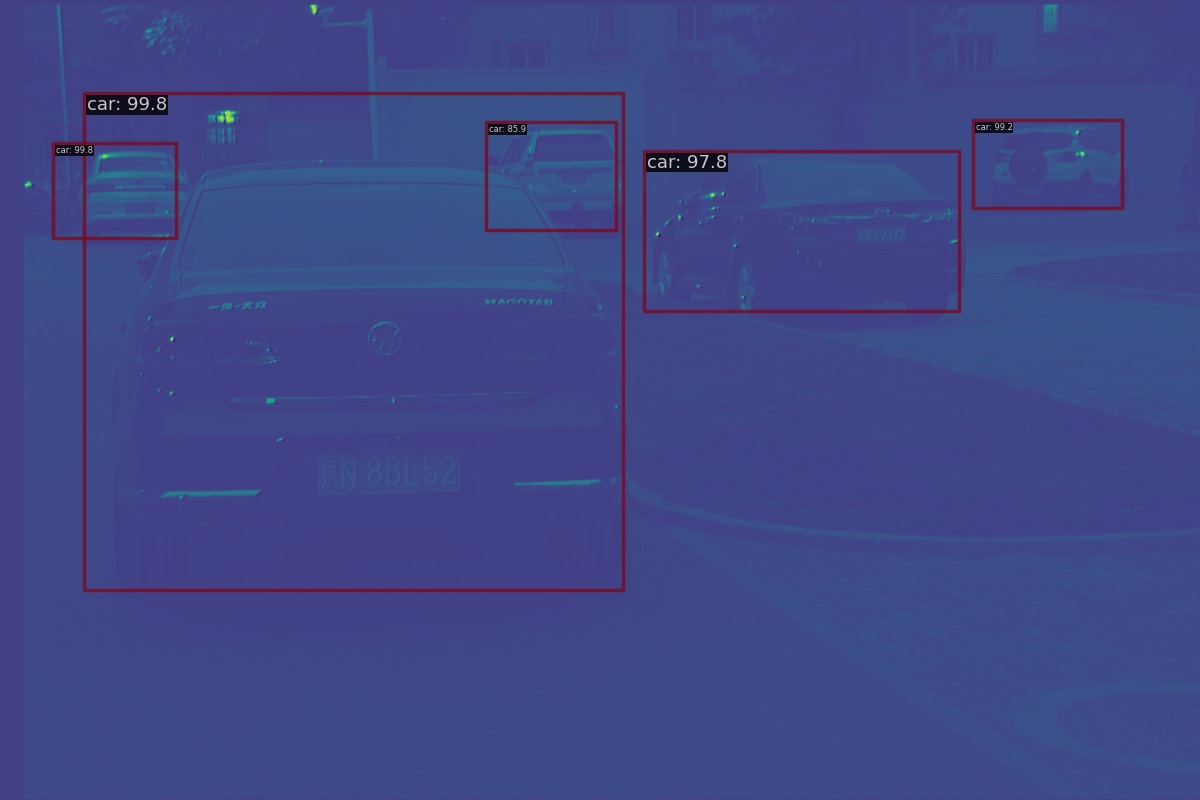}
    \end{minipage}
    \begin{minipage}[t]{0.135\textwidth}
        \centering
        \includegraphics[width=\textwidth]{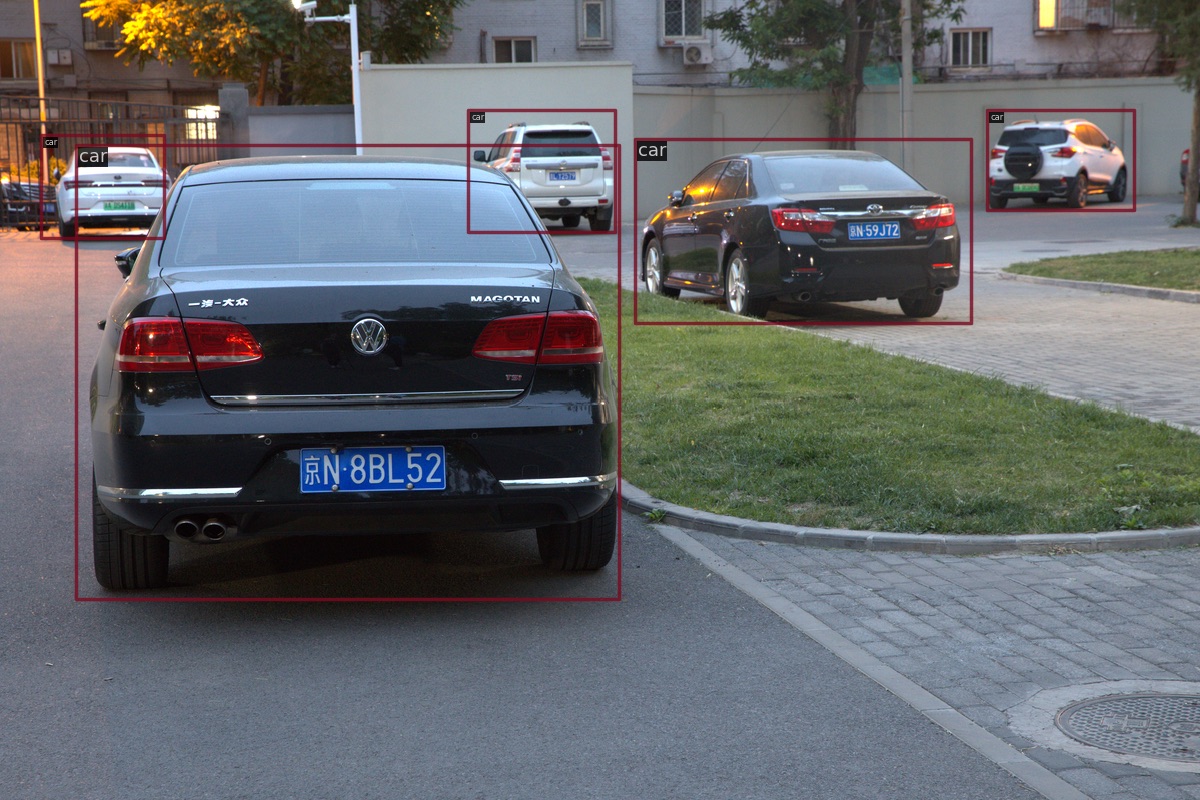}
    \end{minipage}

    \vspace{0.2em}

    \begin{minipage}[t]{0.01\textwidth}
        \centering
        \rotatebox{90}{\footnotesize NOD-Sony}
    \end{minipage}
    \begin{minipage}[t]{0.135\textwidth}
        \centering
        \includegraphics[width=\textwidth]{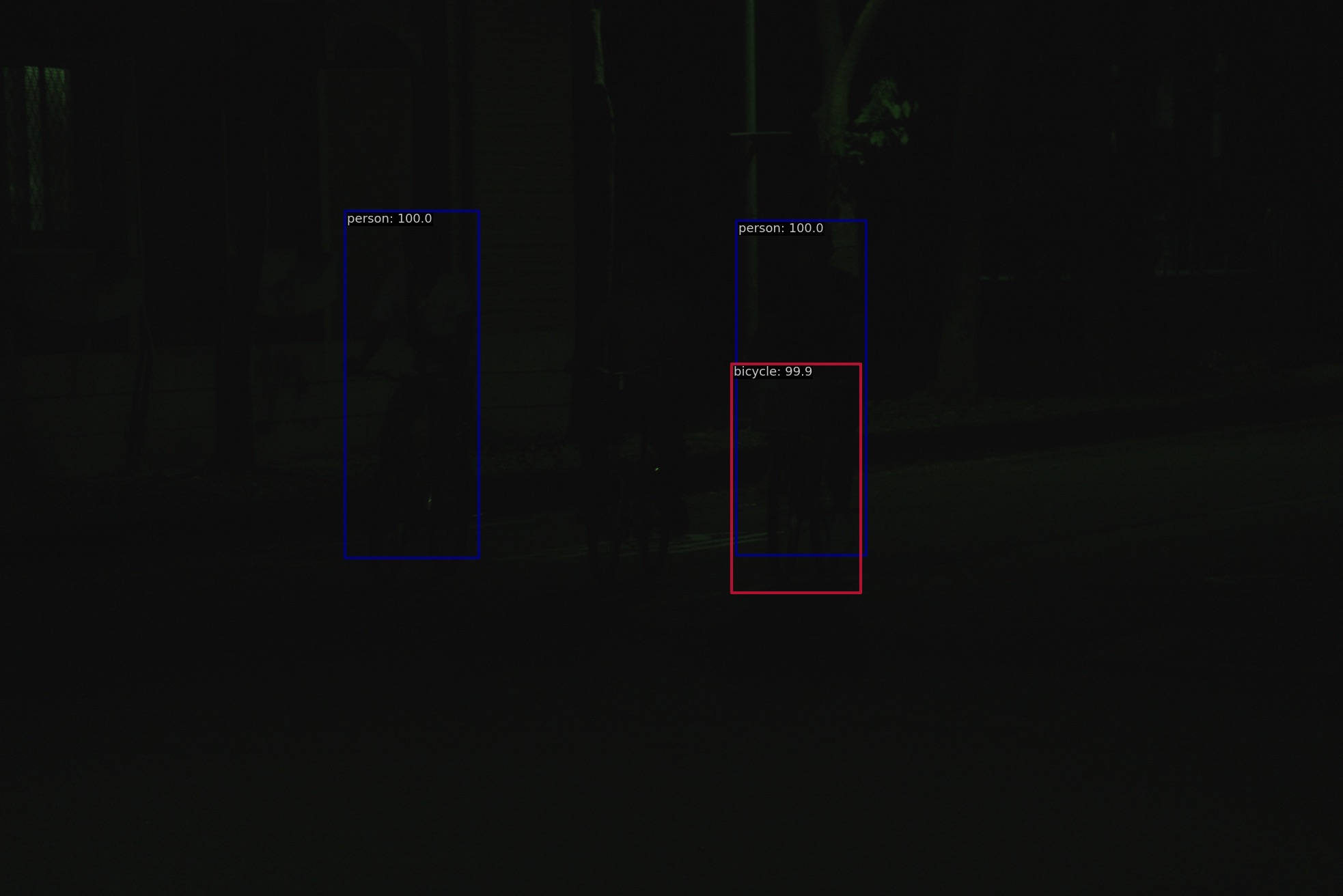}
    \end{minipage}
    \begin{minipage}[t]{0.135\textwidth}
        \centering
        \includegraphics[width=\textwidth]{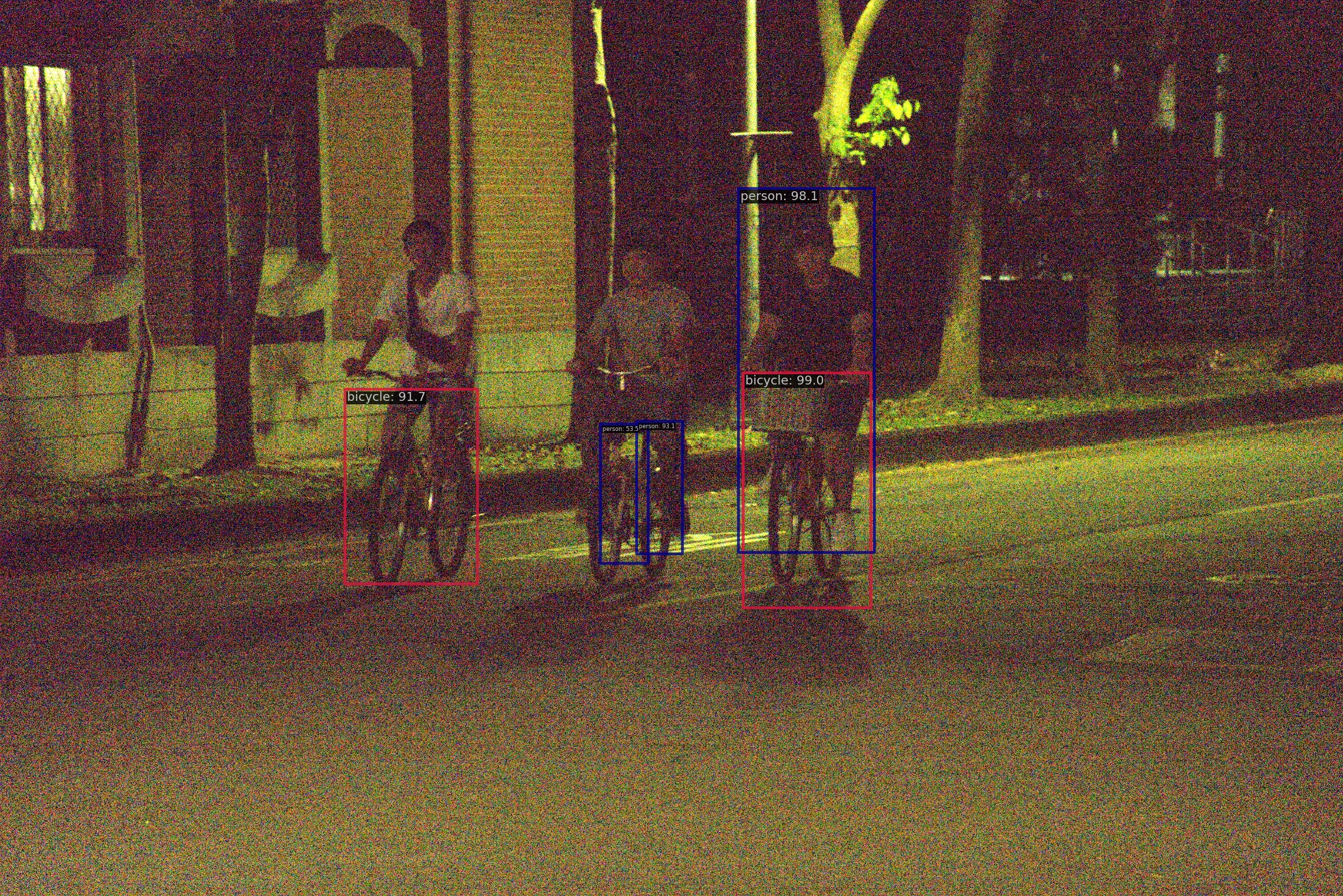}
    \end{minipage}
    \begin{minipage}[t]{0.135\textwidth}
        \centering
        \includegraphics[width=\textwidth]{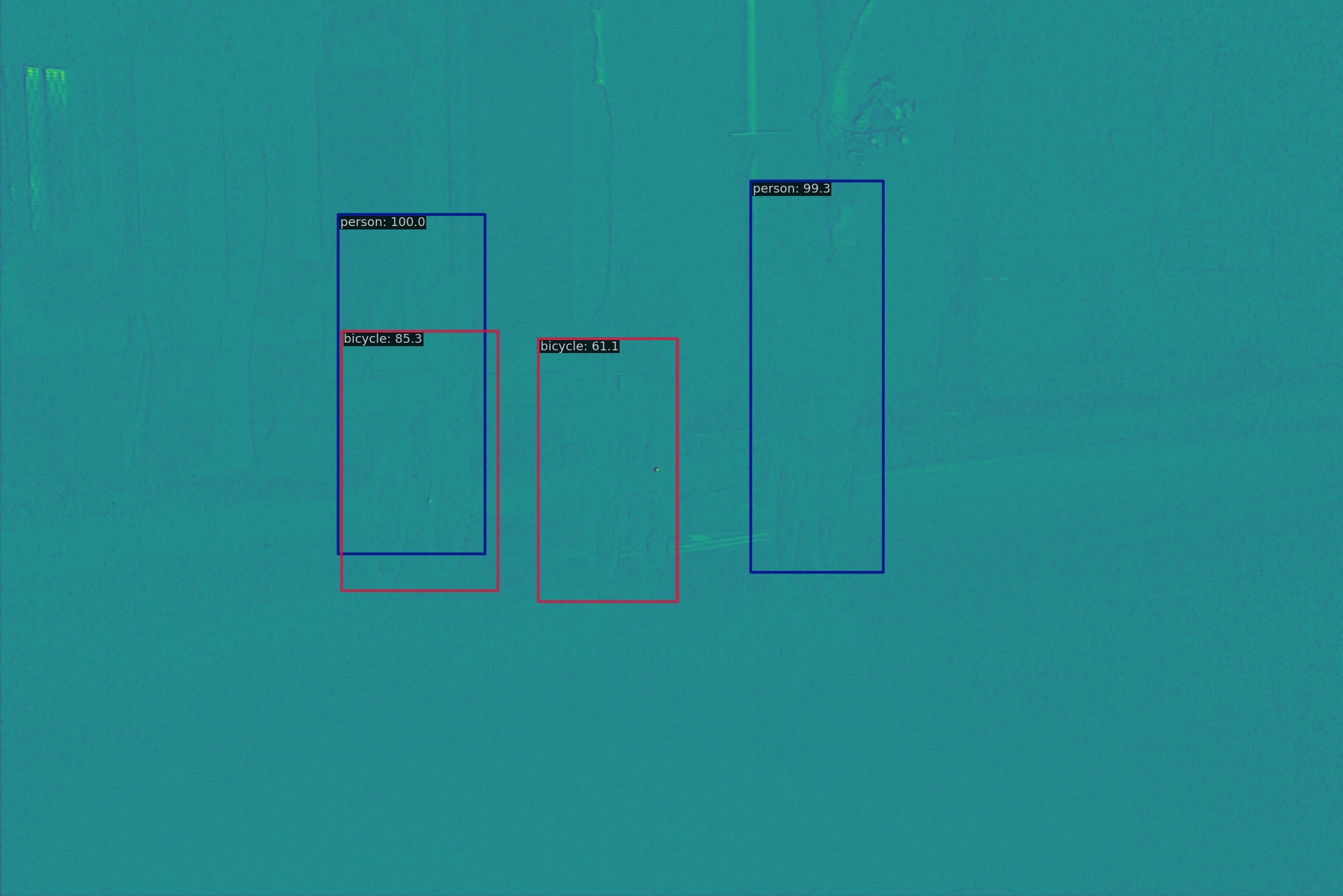}
    \end{minipage}
    \begin{minipage}[t]{0.135\textwidth}
        \centering
        \includegraphics[width=\textwidth]{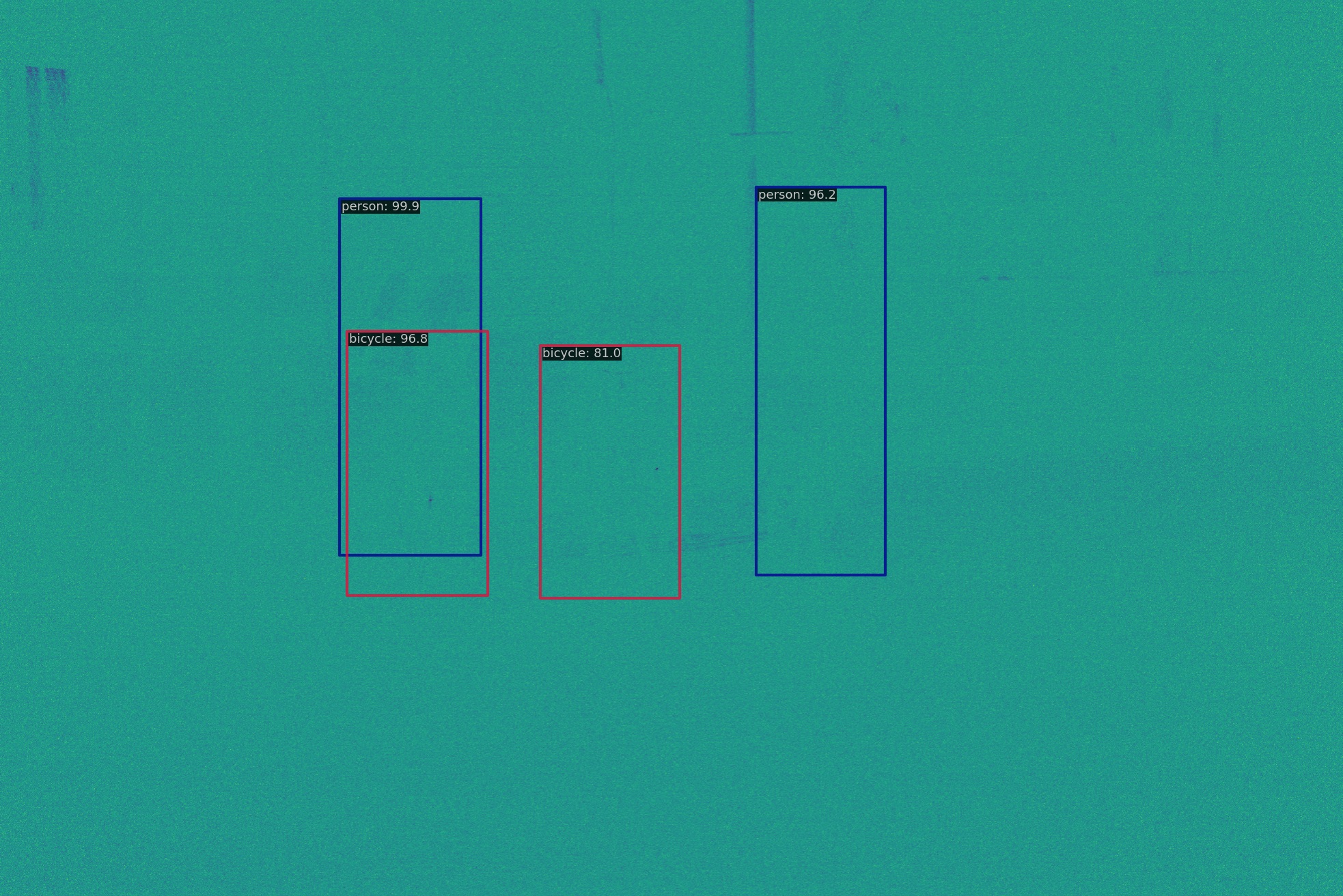}
    \end{minipage}
    \begin{minipage}[t]{0.135\textwidth}
        \centering
        \includegraphics[width=\textwidth]{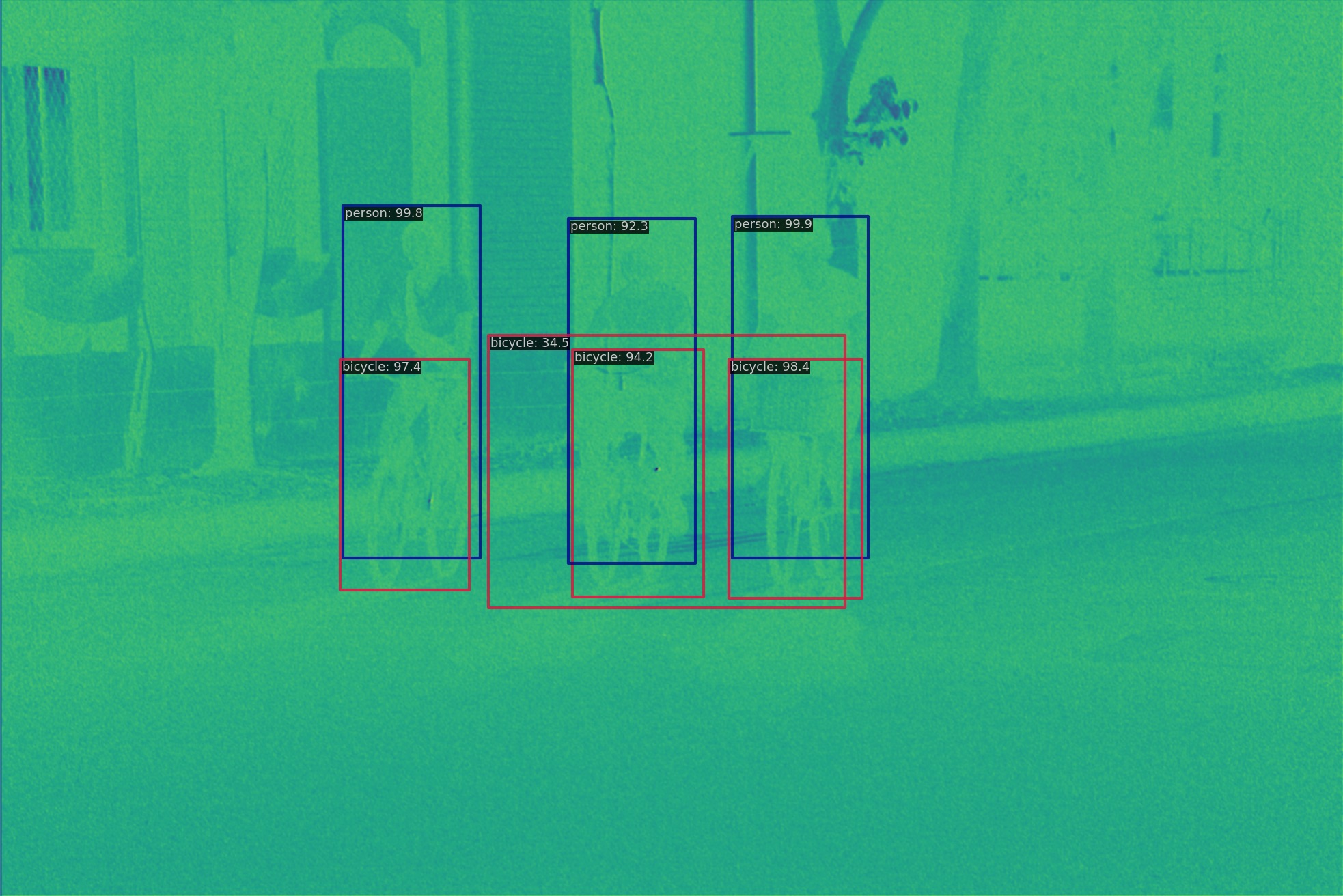}
    \end{minipage}
    \begin{minipage}[t]{0.135\textwidth}
        \centering
        \includegraphics[width=\textwidth]{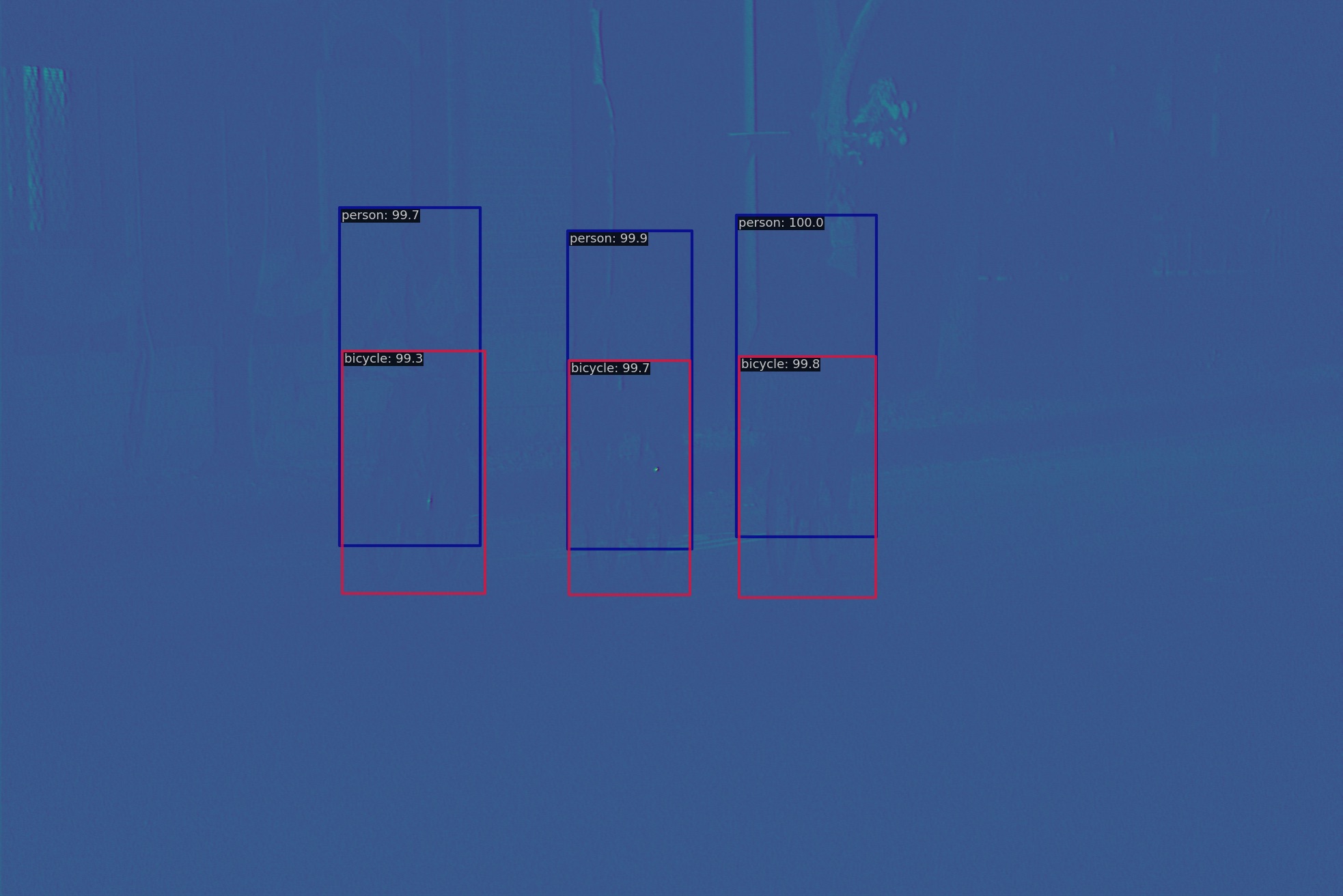}
    \end{minipage}
    \begin{minipage}[t]{0.135\textwidth}
        \centering
        \includegraphics[width=\textwidth]{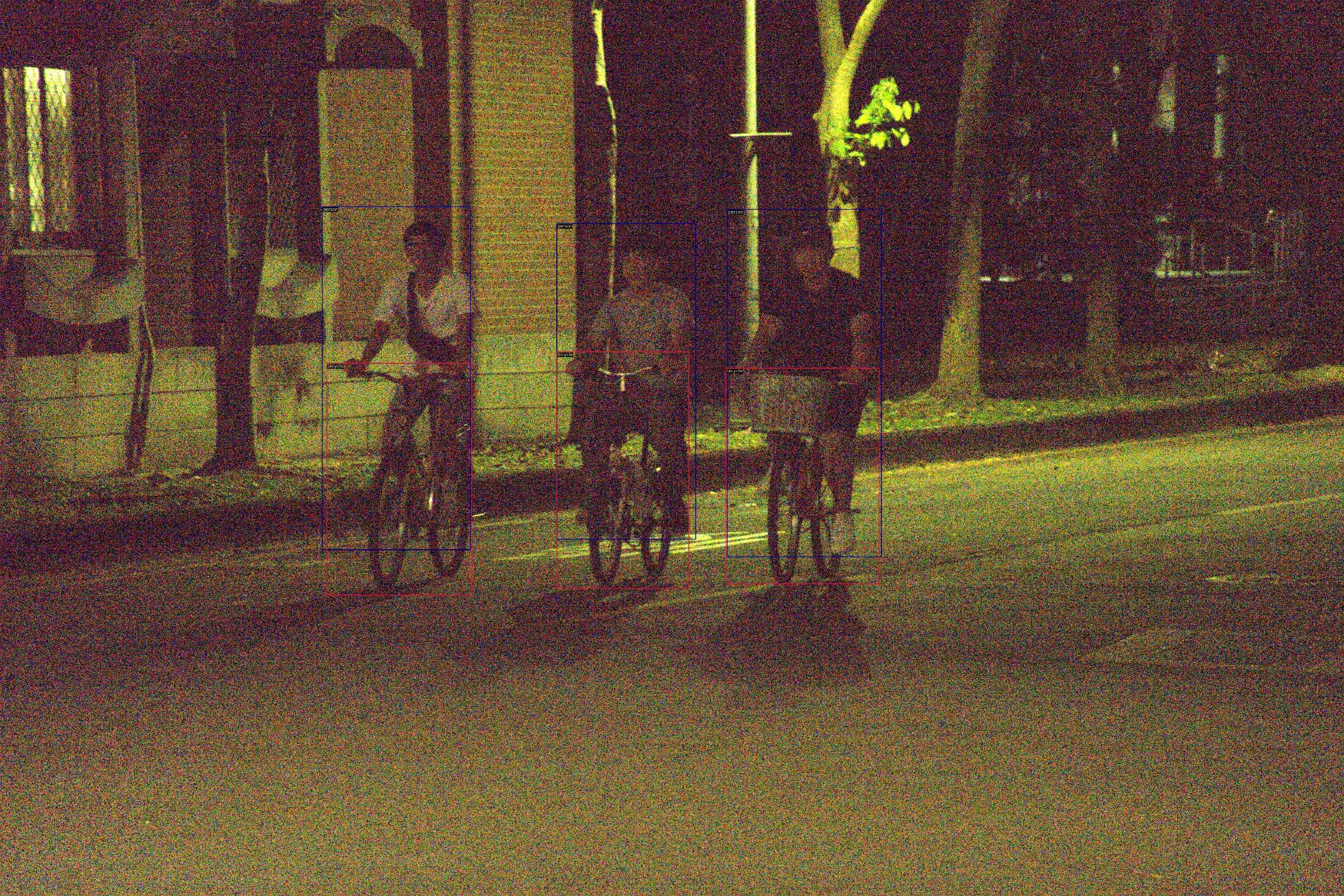}
    \end{minipage}

    \vspace{0.2em}

    \begin{minipage}[t]{0.01\textwidth}
        \centering
        \rotatebox{90}{\footnotesize AROD}
    \end{minipage}
    \begin{minipage}[t]{0.135\textwidth}
        \centering
        \includegraphics[width=\textwidth]{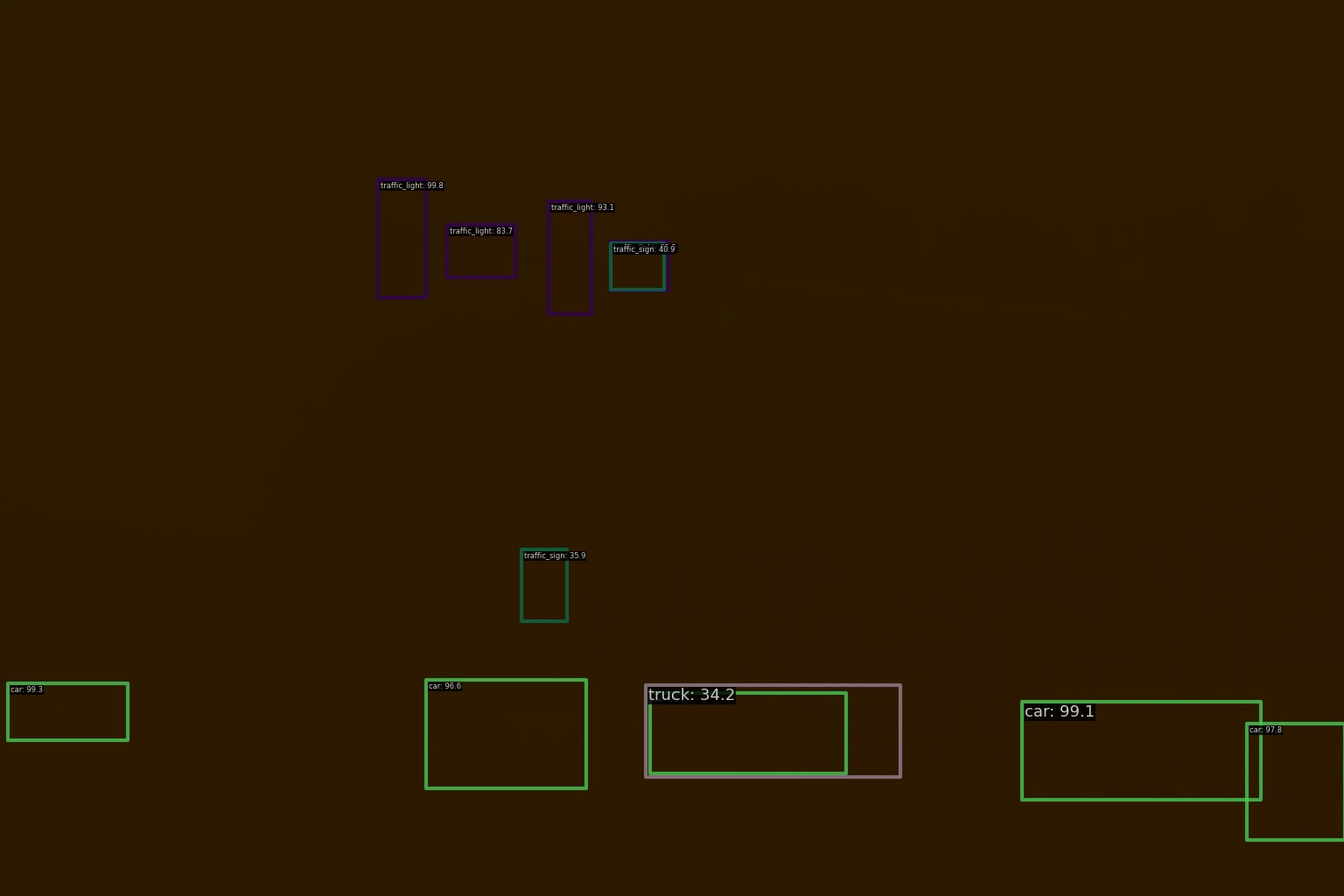}
        {\footnotesize (a) RAW}
    \end{minipage}
    \begin{minipage}[t]{0.135\textwidth}
        \centering
        \includegraphics[width=\textwidth]{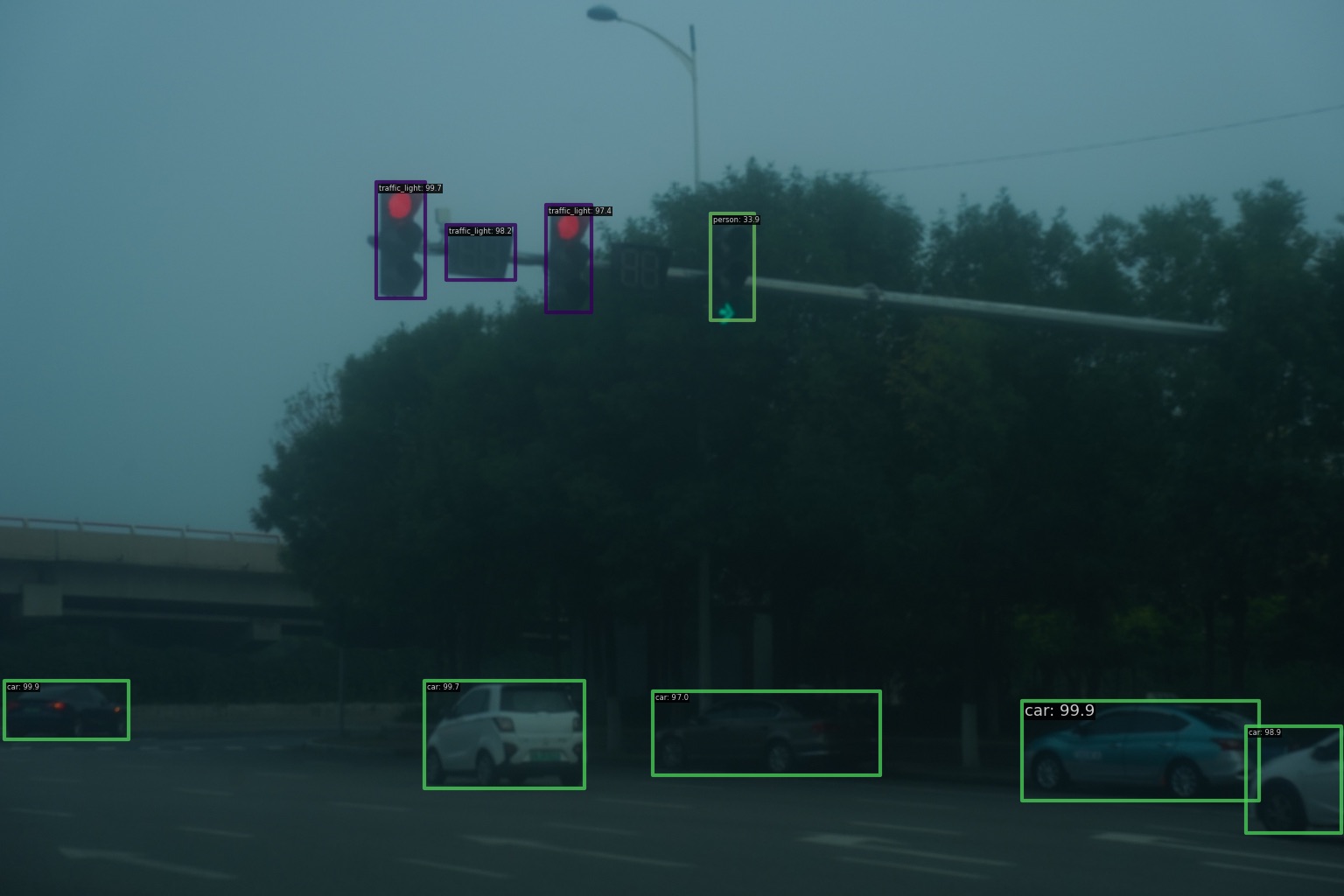}
        {\footnotesize (b) sRGB}
    \end{minipage}
    \begin{minipage}[t]{0.135\textwidth}
        \centering
        \includegraphics[width=\textwidth]{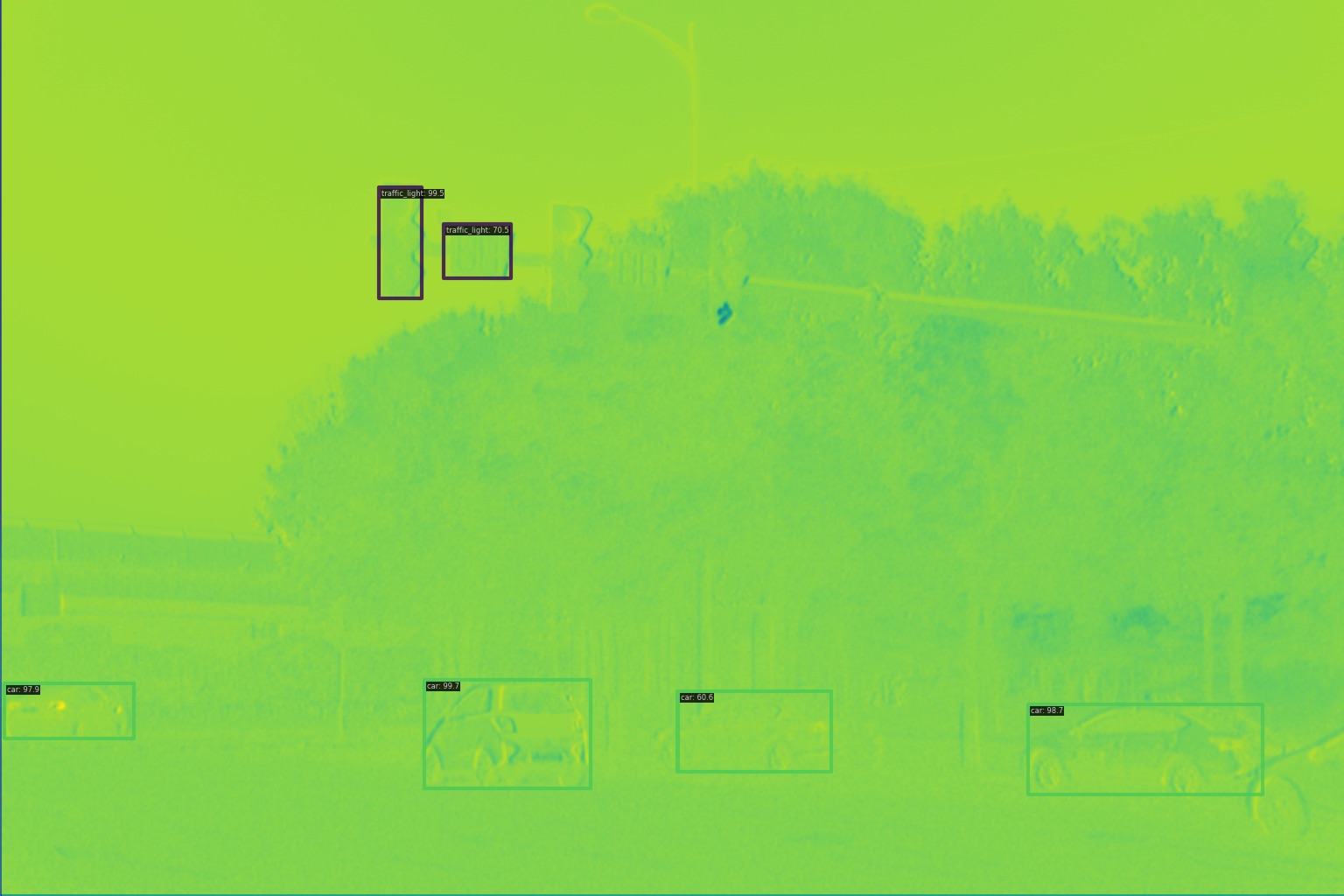}
        {\footnotesize (c) GenISP \cite{morawski2022genisp}}
    \end{minipage}
    \begin{minipage}[t]{0.135\textwidth}
        \centering
        \includegraphics[width=\textwidth]{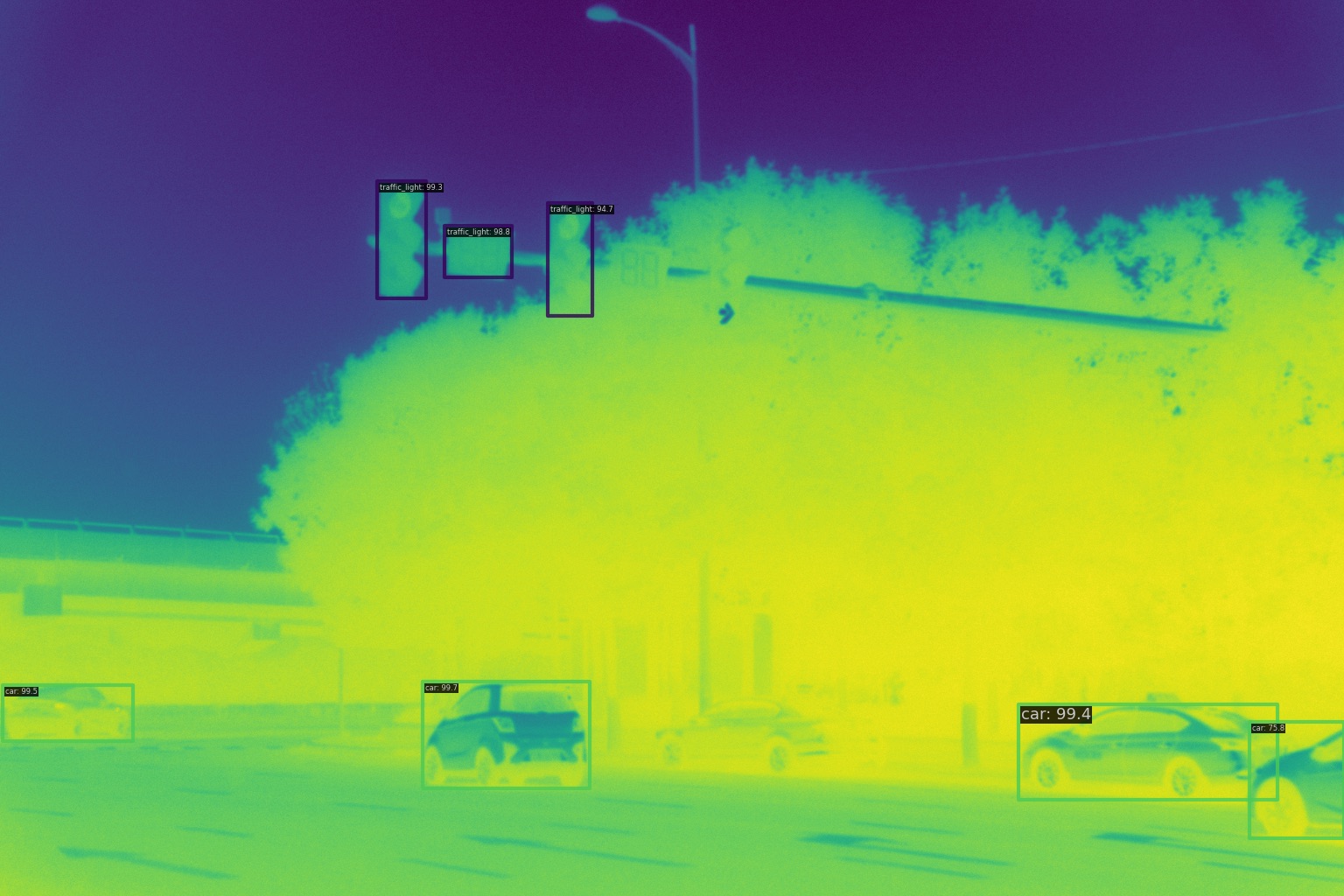}
        {\footnotesize (d) RAOD \cite{xu2023toward}}
    \end{minipage}
    \begin{minipage}[t]{0.135\textwidth}
        \centering
        \includegraphics[width=\textwidth]{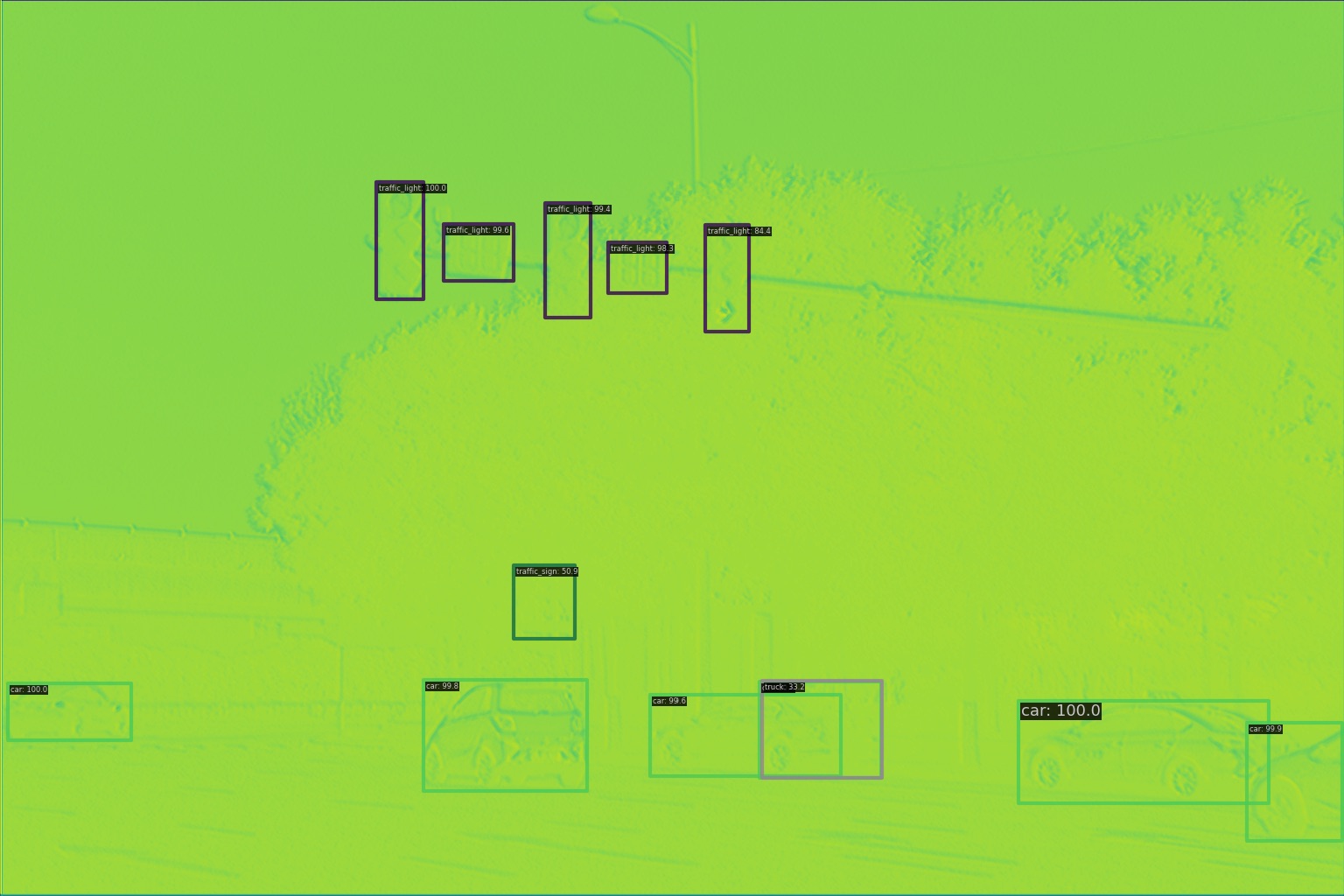}
        {\footnotesize (e) RAM \cite{gamrian2025beyond}}
    \end{minipage}
    \begin{minipage}[t]{0.135\textwidth}
        \centering
        \includegraphics[width=\textwidth]{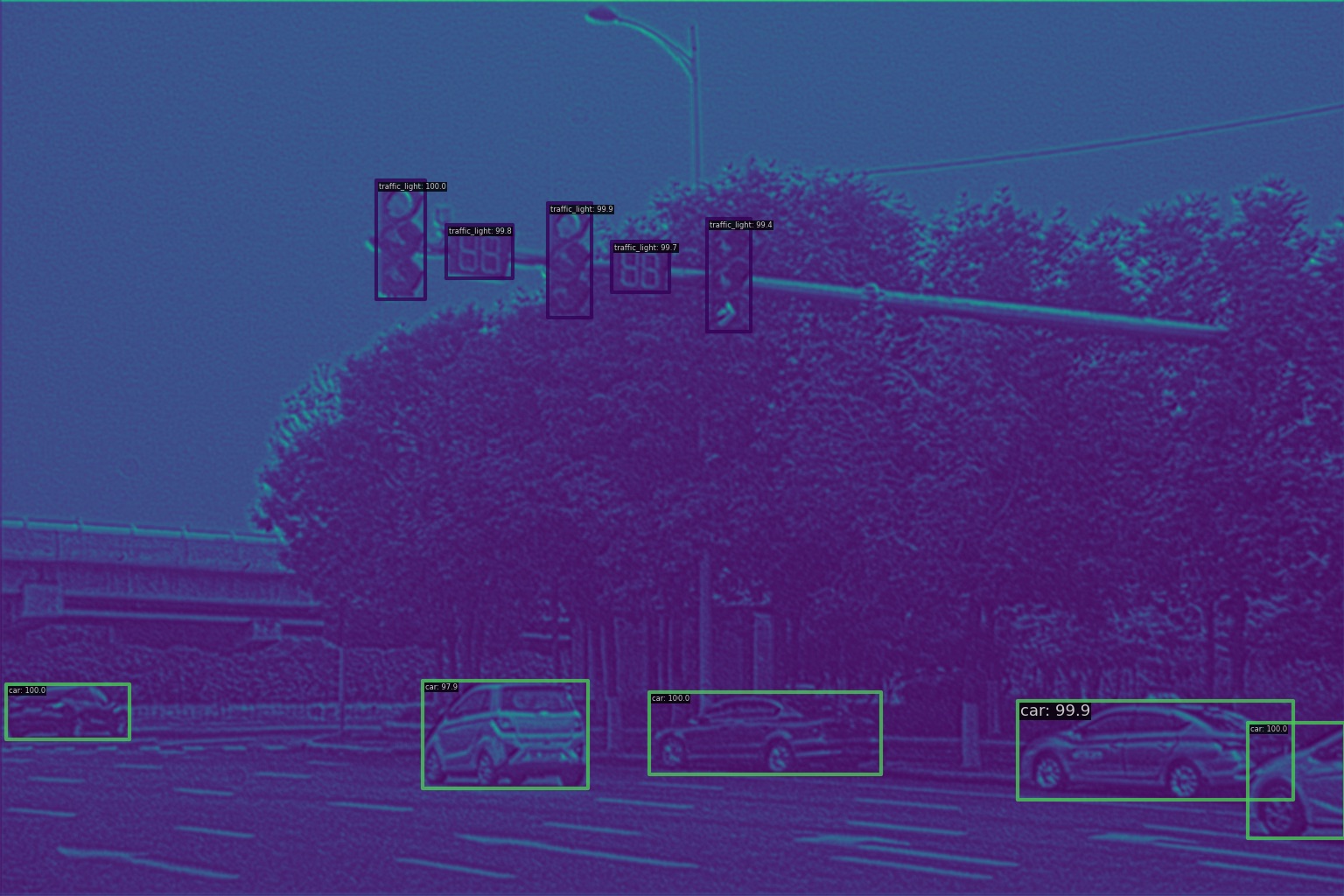}
        {\footnotesize (f) FreqAdapt}
    \end{minipage}
    \begin{minipage}[t]{0.135\textwidth}
        \centering
        \includegraphics[width=\textwidth]{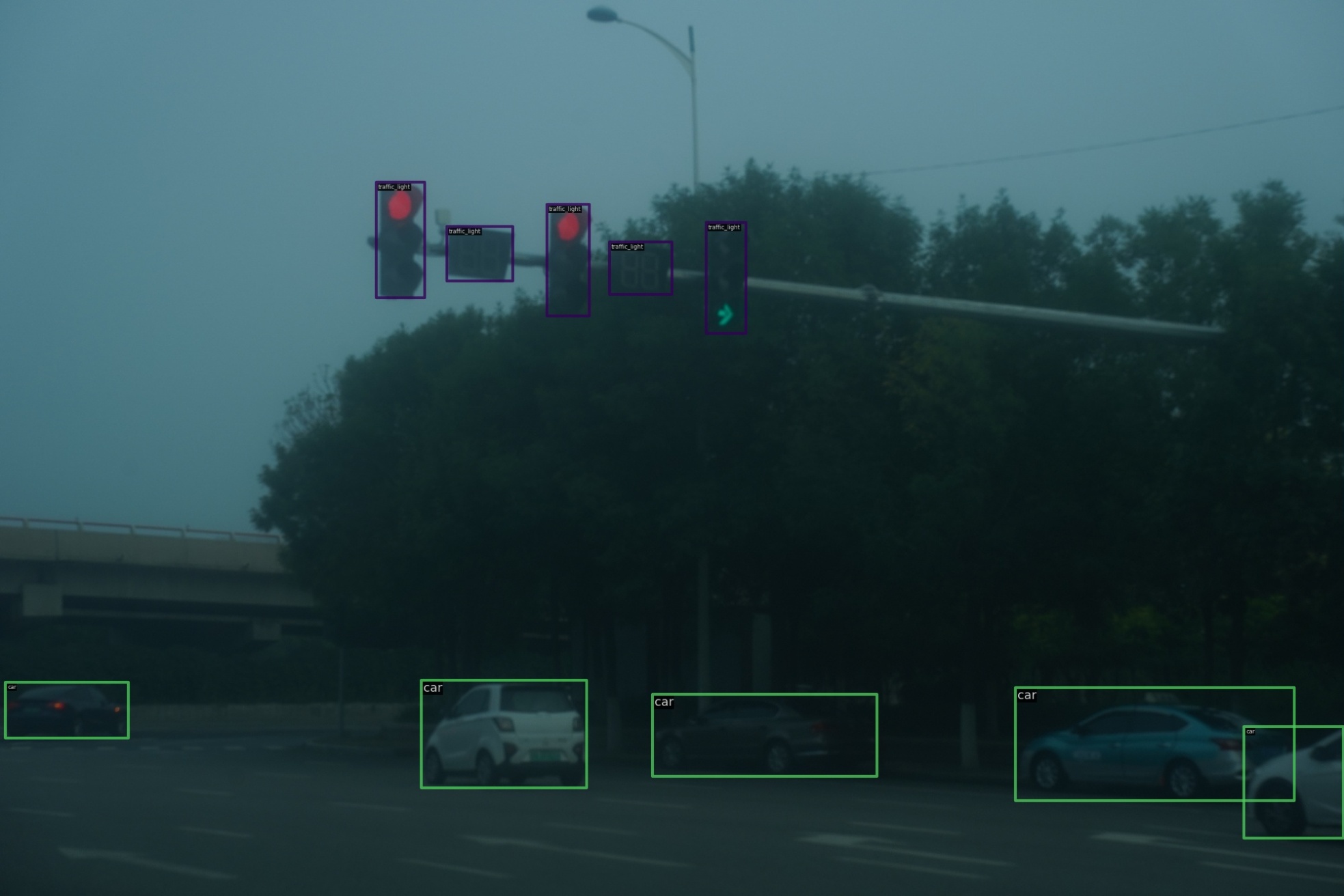}
        {\footnotesize (g) GT}
    \end{minipage}

    \caption{Visual comparison of detection results across different methods on challenging scenarios from LOD-Dark (low-light), NOD-Sony (nighttime), and AROD (adverse weather) datasets. Predictions with confidence scores above 0.3 are shown. FreqAdapt consistently outperforms existing methods with more accurate detections under challenging conditions.}

    \label{fig:object}
     \vspace{-4mm}
\end{figure*}


\subsubsection{Efficiency Analysis}

\begin{table}[!h]
    \centering
    \caption{Comparison of Model Size, GFLOPS, and mAP on the NOD-Sony. ‘Baseline’ represents the reference model, with values indicating the additional time cost and parameters introduced by each method.}
    \label{tab:comparison eff}
    \resizebox{0.48\textwidth}{!}{
    \begin{tabular}{lcccl}
    \toprule
        Method & Params (M) & GFLOPs  & mAP  &mAP$_{50}$\\ 
    \midrule
        RAW (ResNet18)& Baseline(19.81)& Baseline(38.21)& 25.3 &51.4\\
 RAW(ResNet50)& +16.56& +12.48&25.7 &52.8\\
        sRGB & -& -&  25.1&52.3\\
        RAOD \cite{xu2023toward} & +0.260& +2.73& 26.5&54.4\\
        RAM \cite{gamrian2025beyond} & +0.141& +10.93& 26.8&54.7\\
        RAW-Adapter \cite{cui2025raw} & +0.76& +12.58&  25.9&53.5\\
        GenISP \cite{morawski2022genisp} & +0.116& +6.86& 26.1 &53.4\\
 FreqAdapt-T & +0.014& +1.42& 26.7&54.1\\
        FreqAdapt & +0.019& +2.25& \textbf{27.6}&\textbf{55.2}\\
        \bottomrule
    \end{tabular}%
    }
\end{table}



As shown in \cref{tab:comparison eff}, the proposed FreqAdapt method requires only +0.019M additional parameters and +2.25 GFLOPs computational overhead, achieving 27.6 mAP—a +2.3\% improvement over the baseline RAW input (25.3 mAP). Compared to other learnable methods, FreqAdapt demonstrates remarkable efficiency: RAOD adds +0.260M parameters for 26.5 mAP and RAM requires +0.141M parameters for 26.8 mAP, while our method achieves superior performance using merely 7.3\% of RAOD's parameters and 13.5\% of RAM's parameters. This efficiency advantage stems from frequency-domain processing, which captures global image characteristics more compactly than spatial-domain approaches.

The comparison with ResNet50 further highlights our efficiency superiority. Upgrading from ResNet18 to ResNet50 introduces +16.56M parameters for only +0.4\% mAP gain, whereas FreqAdapt achieves +2.3\% improvement with less than 0.2\% of ResNet50's additional parameters. This demonstrates that frequency-domain processing provides a more parameter-efficient path to performance enhancement than simply scaling model capacity.

\subsubsection{Experiments with Frozen Detector}

In practical scenarios, RAW image datasets are often too limited to train large object detectors from scratch. A common practice involves utilizing pre-trained detectors trained on extensive datasets. To evaluate FreqAdapt's adaptability in such settings, we conducted experiments with frozen downstream detection networks, as shown in \cref{tab:comparison frozen}. We employed RetinaNet with ResNet18 backbone for evaluation due to its widespread adoption and computational efficiency.

FreqAdapt demonstrates superior adaptation capability, achieving 14.9 mAP on NOD-Nikon (+0.7 over RAM, +3.1 over GenISP) and 18.6 mAP on NOD-Sony (+2.1 over RAM, +0.4 over RAOD). The consistent improvements across both camera types validate our method's effectiveness in bridging the domain gap between RAW inputs and sRGB-trained features.
This superior performance stems from frequency-domain processing's ability to preserve both low-frequency content expected by pre-trained networks and high-frequency details crucial for detection. By separately optimizing amplitude and phase components, FreqAdapt maintains the statistical properties required by the frozen detector while enhancing task-relevant features, demonstrating its versatility across different training paradigms.

\begin{table}[!ht]
    \centering
    \caption{Comparison of different approaches to frozen downstream detection networks}
    \label{tab:comparison frozen}
    \resizebox{0.48\textwidth}{!}{%
    \begin{tabular}{lcccc}
\toprule
    & \multicolumn{2}{c}{NOD-Nikon} & \multicolumn{2}{c}{NOD-Sony} \\
    \cmidrule(lr){2-3} \cmidrule(lr){4-5}
    Method & mAP & mAP$_{50}$ & mAP & mAP$_{50}$ \\
    \midrule
    GenISP\cite{morawski2022genisp} & 11.8& 28.5& 13.8& 31.8\\
    RAOD \cite{xu2023toward} & 13.9& 32.9& 18.2& 38.7\\
    RAM \cite{gamrian2025beyond} & 14.2& 33.3& 16.5& 35.7\\
    FreqAdapt  & \textbf{14.9}& \textbf{33.8}& \textbf{18.6}& \textbf{39.2}\\ 
\bottomrule
    \end{tabular}%
    }
\end{table}

\subsubsection{Qualitative Analysis}

\cref{fig:object} presents qualitative detection results across three challenging RAW datasets (LOD-Dark, NOD-Sony, and AROD), demonstrating FreqAdapt's superior performance in diverse scenarios.   On LOD-Dark (first row), underexposure causes other methods to suffer from missed detections and false positives, while FreqAdapt successfully identifies objects with results closely matching the ground truth.  For NOD-Sony nighttime scenes (second row), where significant noise interference degrades detection performance, FreqAdapt benefits from its frequency-domain denoising capabilities to robustly detect pedestrians and cyclists that remain invisible to other approaches.   In the extreme conditions of AROD (third row), featuring foggy and low-contrast environments, FreqAdapt reliably detects vehicles at various distances and traffic lights, whereas competing methods exhibit notable failures in detecting distant or small objects.  The consistent superiority across all three datasets validates our frequency-domain processing's effectiveness in handling underexposure, noise, and low contrast—the primary challenges in RAW-based object detection.






\subsection{Ablation Studies}

We conduct extensive ablation studies to validate the design choices in FreqAdapt and analyze the contribution of each component.

\subsubsection{Comparison with Spatial Domain Processing}


To validate the effectiveness of frequency-domain processing over spatial-domain approaches, we implemented a spatial-domain variant using the same ISP functions but applied directly in the spatial domain.  The spatial version processes RAW images through sequential ISP operations including white balance, color correction, gamma correction, and brightness adjustment, with separate parameter prediction modules for each function. 

As shown in \cref{tab:spat vs freq}, FreqAdapt significantly outperforms the spatial approach, achieving 27.2 mAP (+2.3) and 52.6 mAP$_{50}$ (+4.2) on LOD-Dark, with more pronounced improvements on LOD-Normal at 35.1 mAP (+3.2) and 62.7 mAP$_{50}$ (+3.7). The superior performance stems from frequency-domain processing's inherent advantages: global context awareness through frequency components enables more effective parameter prediction, decoupled amplitude-phase processing prevents interference between brightness adjustments and structural enhancements, and frequency-domain operations naturally handle noise and artifacts more efficiently than spatial convolutions. 


\begin{table}[h]
\centering
\caption{Comparison of spatial and FreqAdapt on LOD-Dark and LOD-Normal datasets.}
\resizebox{0.45\textwidth}{!}{%
\begin{tabular}{lcccc}
\hline
\multirow{2}{*}{Method} & \multicolumn{2}{c}{LOD-Dark} & \multicolumn{2}{c}{LOD-Normal} \\
\cmidrule(lr){2-3} \cmidrule(lr){4-5}
 & mAP & mAP$_{50}$ & mAP & mAP$_{50}$ \\
\hline
spatial & 24.9& 48.4& 31.9& 59.0\\
FreqAdapt & \textbf{27.2} & \textbf{52.6} & \textbf{35.1} & \textbf{62.7} \\
\hline
\end{tabular}%
}
\label{tab:spat vs freq}
\end{table}

\subsubsection{Component Analysis of FreqAdapt}

To investigate the contribution of each module in FreqAdapt, we conducted systematic ablation experiments on the LOD-Dark dataset. As shown in \ref{tab:ablation}, we evaluate four key components through progressive addition.
Starting with only the Frequency RAW Encoder (24.6 mAP), we observe the baseline performance of frequency-domain feature extraction. Adding the Phase ISP Branch (25.4 mAP, +0.8) demonstrates modest improvement through structural enhancement. The most significant gain comes from incorporating the Amplitude ISP Branch (26.9 mAP, +1.5), highlighting the critical importance of brightness and contrast adjustments in low-light scenarios. Finally, the Adaptive Fusion Module brings the performance to 27.2 mAP (+0.3), effectively combining frequency-enhanced and original features.

The results reveal that while all components contribute positively, the Amplitude ISP Branch provides the largest individual contribution (+1.5 mAP), confirming that amplitude-domain processing for brightness, white balance, and noise reduction is particularly crucial for dark image detection. The Phase ISP Branch and Adaptive Fusion Module provide complementary improvements, with their combined effect demonstrating the advantage of our decoupled frequency-domain design. The full model's performance (27.2 mAP) represents a +2.6 mAP improvement over the encoder-only baseline, validating the effectiveness of our complete pipeline.
Due to space constraints, we have included additional experimental results, visualizations, and ablation analysis in the
supplementary material.

\begin{table}[h]
\centering
\caption{Ablation study of the proposed Modules on the LOD-Dark}
\resizebox{0.45\textwidth}{!}{%
\begin{tabular}{ccccc}
\toprule
\begin{tabular}[c]{@{}c@{}}Frequency RAW\\Encoder\end{tabular} & 
\begin{tabular}[c]{@{}c@{}}Phase ISP\\Branch\end{tabular} & 
\begin{tabular}[c]{@{}c@{}}Amplitude ISP\\Branch\end{tabular} & 
\begin{tabular}[c]{@{}c@{}}Adaptive Fusion\\Module\end{tabular} & mAP \\
\midrule
 \checkmark& &  &  & 24.6\\
 \checkmark&  \checkmark& &  & 25.4\\
 \checkmark&  \checkmark&  \checkmark& & 26.9\\
\checkmark & \checkmark & \checkmark & \checkmark & \textbf{27.2}\\
\bottomrule
\end{tabular}%
}
\label{tab:ablation}
\end{table}

\section{Conclusion}
\label{sec:conc}

We present FreqAdapt, a frequency-domain adaptive module for RAW image enhancement in object detection.  Leveraging Fourier properties, FreqAdapt reformulates ISP by processing intensity-related operations (e.g., gamma correction, white balance, brightness) in the amplitude domain and structure-related operations (e.g., color correction matrix, sharpening, detail enhancement) in the phase domain, enabling targeted, efficient computation and mitigating the cumulative information loss of sequential spatial pipelines.  The key insight is that ISP functions naturally align with frequency components—amplitude governs energy distribution and contrast, while phase preserves structural and edge information—so an orthogonal decomposition processes each in its most tractable domain while maintaining physical interpretability.  FreqAdapt is lightweight and plug-and-play, integrating seamlessly into existing detectors without architectural changes or complex training.  Extensive experiments on challenging RAW detection benchmarks show consistent gains over state-of-the-art methods with markedly lower computation.  




Although this study focuses on object detection, the proposed method can be naturally extended to other visual perception tasks, such as segmentation, tracking, and restoration, providing an effective framework to better exploit the rich information contained in RAW sensor data. For future work, we aim to investigate the effectiveness of incorporating RAW-domain inputs into end-to-end autonomous driving models, particularly under extreme real-world conditions such as adverse weather and low-light environments.


{
    \small
    \bibliographystyle{ieeenat_fullname}
    \bibliography{main}
}

\newpage
\appendix
\clearpage
\setcounter{page}{1}
\maketitlesupplementary

\section{Experiments Settings}


In the experiments presented in Section 4, we employed different configurations tailored to each dataset and evaluation scenario.
For the LOD and NOD datasets, we trained Faster R-CNN using SGD optimizer with learning rate of 0.02, momentum of 0.9. The model was trained for 50 epochs with batch size of 4, employing a multi-step learning rate scheduler that decayed the learning rate at epochs 30 and 40.
To evaluate computational efficiency, we used RetinaNet as a single-stage baseline, trained with the same SGD configuration and training schedule as Faster R-CNN.
For the AODRAW dataset, we utilized Cascade R-CNN trained with Adam optimizer (learning rate of 0.001) and batch size of 4. The training lasted 50 epochs with the same multi-step scheduling strategy, applying learning rate decay at epochs 30 and 40. 
Data augmentation strategies were consistently applied across all experiments, including random resizing and horizontal flipping to enhance model generalization. Model performance was evaluated using standard COCO-style metrics, specifically mean Average Precision (mAP) and mAP at IoU threshold of 0.5 (mAP$_{50}$).

\section{Additional experiments}

\subsection{ISP Function Ablation Study}

\begin{table}[h]
\centering
\caption{Quantitative ablation study of individual ISP functions on the LOD-Dark dataset. Each row represents the performance when a specific function is removed while all other functions remain active.}
\label{tab:isp_ablation}
\resizebox{0.48\textwidth}{!}{%
\begin{tabular}{l|ccc}
\toprule
\textbf{Configuration} & \textbf{mAP} & \textbf{Params (K)}& \textbf{GFLOPs} \\
\midrule
Full ISP (Baseline) & \textbf{27.2} & 19.41& 2.26\\
\midrule
\multicolumn{4}{l}{\textit{Amplitude-domain ablations}} \\
w/o White Balance & 27.0(-0.2)& 18.83(-0.58)& 2.26\\
w/o Gamma & 25.2(-2.0)& 18.88(-0.54)& 2.26\\
w/o Brightness & 26.4(-0.8)& 18.87(-0.55)& 2.26\\
w/o Noise Reduction & 26.1(-1.1)& 17.36(-2.06)& 2.03(-0.22)\\
\midrule
\multicolumn{4}{l}{\textit{Phase-domain ablations}} \\
w/o CCM & 26.5(-0.7)& 18.73(-0.68)& 2.26\\
w/o Sharpening & 26.7(-0.5)& 18.24(-1.17)& 2.10(-0.15)\\
w/o Detail Enhancement & 26.9(-0.3)& 16.99(-2.42)& 1.80(-0.46)\\
\bottomrule
\end{tabular}%
}
\end{table}

As shown in the \cref{tab:isp_ablation}, we present the contribution of each ISP function module to the overall performance and parameter efficiency on the LOD-Dark dataset. Among the four amplitude-domain function modules, Gamma correction has the most significant impact, with mAP dropping by 2.0 after removal, indicating its critical role in low-light image processing. This aligns with the importance of Gamma correction in adjusting image brightness and contrast. Noise suppression ranks second, with a performance drop of 1.1 mAP after removal, as noise is more pronounced in low-light environments. This module consumes considerable parameters and computational resources, reflecting the complexity of noise processing. Brightness adjustment shows a 0.8 mAP decrease when removed, demonstrating its contribution to low-light enhancement. White balance has the minimal impact, reducing only 0.2 mAP, possibly because color correction is less important than brightness and noise processing under low-light conditions.

The three phase-domain function modules have relatively smaller overall impacts. CCM removal results in a 0.7 mAP drop, the largest among phase-domain functions, indicating that color transformation still plays a role in feature representation. Sharpening and detail enhancement have relatively minor effects, causing 0.5 and 0.3 mAP drops respectively. Amplitude-domain functions have a greater overall impact on performance than phase-domain functions, indicating that brightness and noise-related processing is more important than frequency-domain detail processing under low-light conditions.



\subsection{Visualization of AROD Challenging Scenarios}

\begin{figure*}[h]
    \centering

    \begin{minipage}[t]{0.01\textwidth}
        \centering
        \rotatebox{90}{\footnotesize Low-Night}
    \end{minipage}
    \begin{minipage}[t]{0.135\textwidth}
        \centering
        \includegraphics[width=\textwidth]{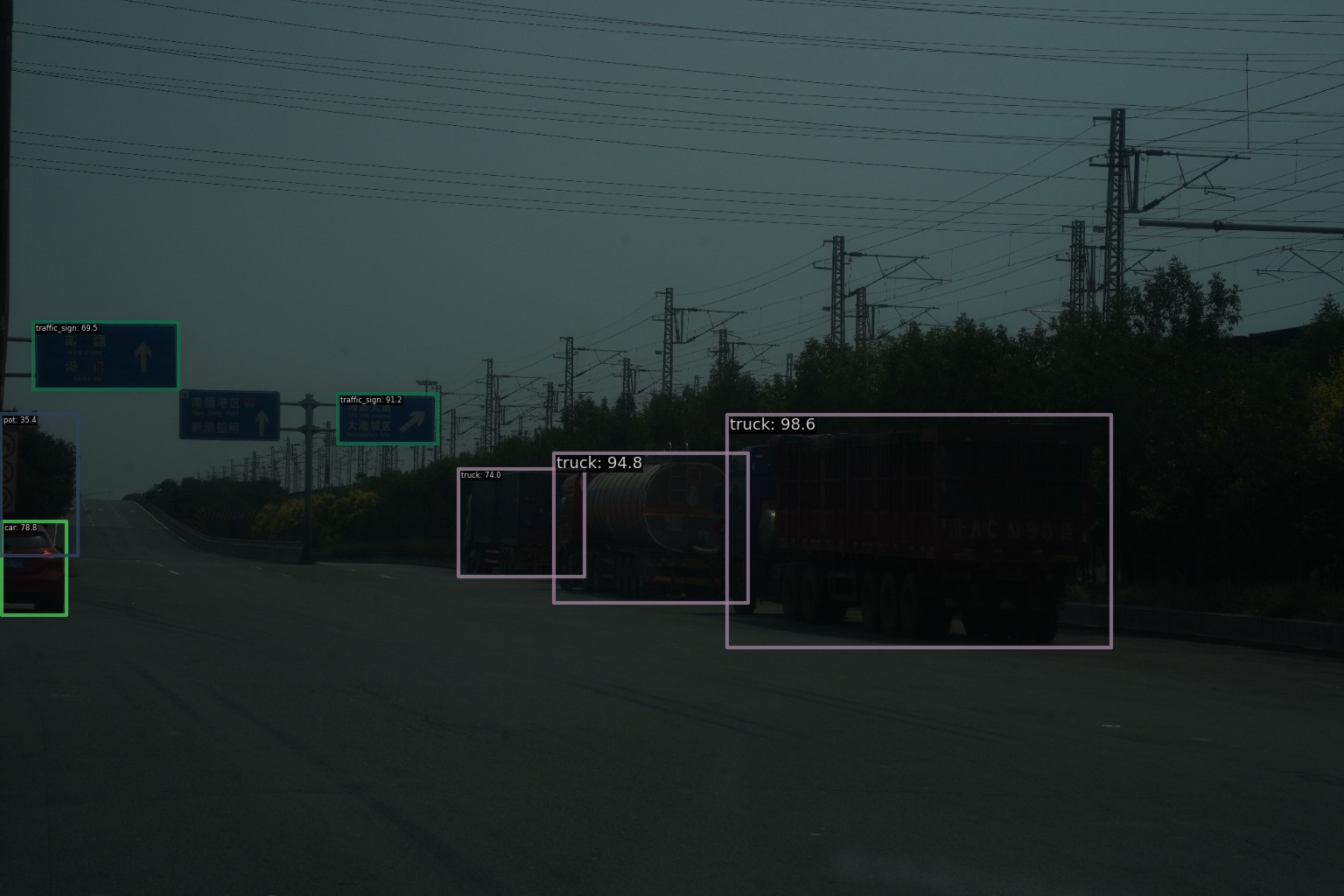}
    \end{minipage}
    \begin{minipage}[t]{0.135\textwidth}
        \centering
        \includegraphics[width=\textwidth]{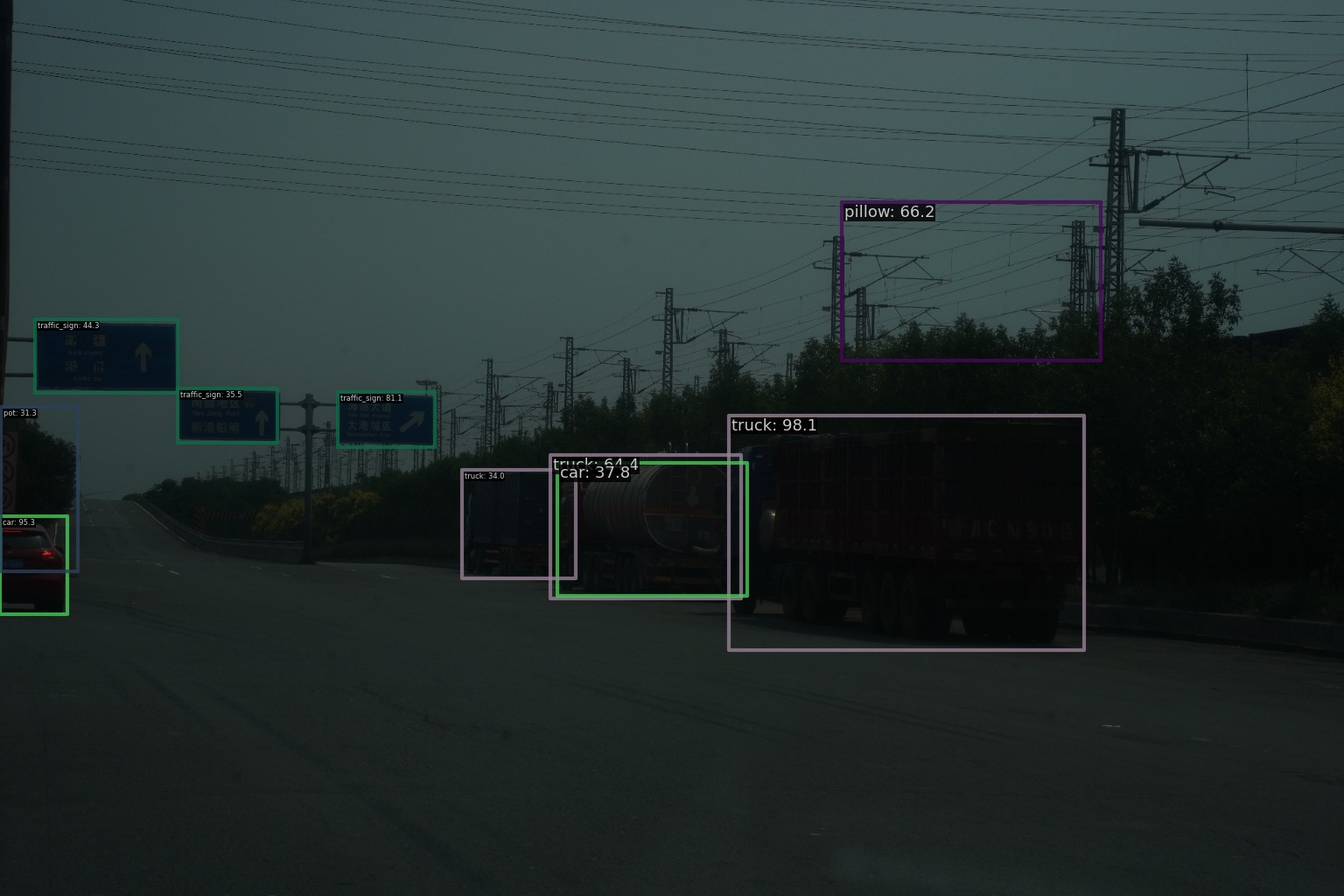}
    \end{minipage}
    \begin{minipage}[t]{0.135\textwidth}
        \centering
        \includegraphics[width=\textwidth]{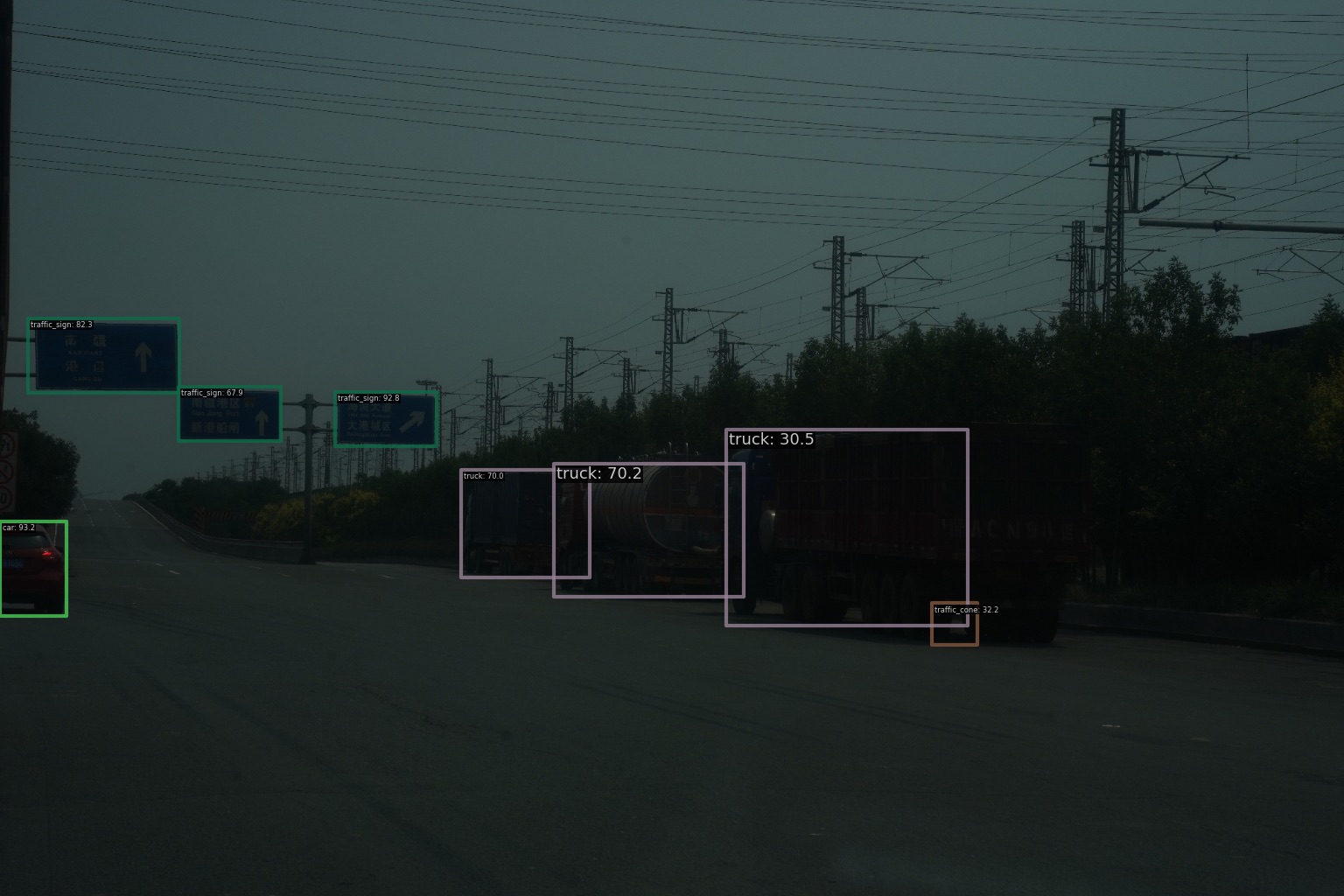}
    \end{minipage}
    \begin{minipage}[t]{0.135\textwidth}
        \centering
        \includegraphics[width=\textwidth]{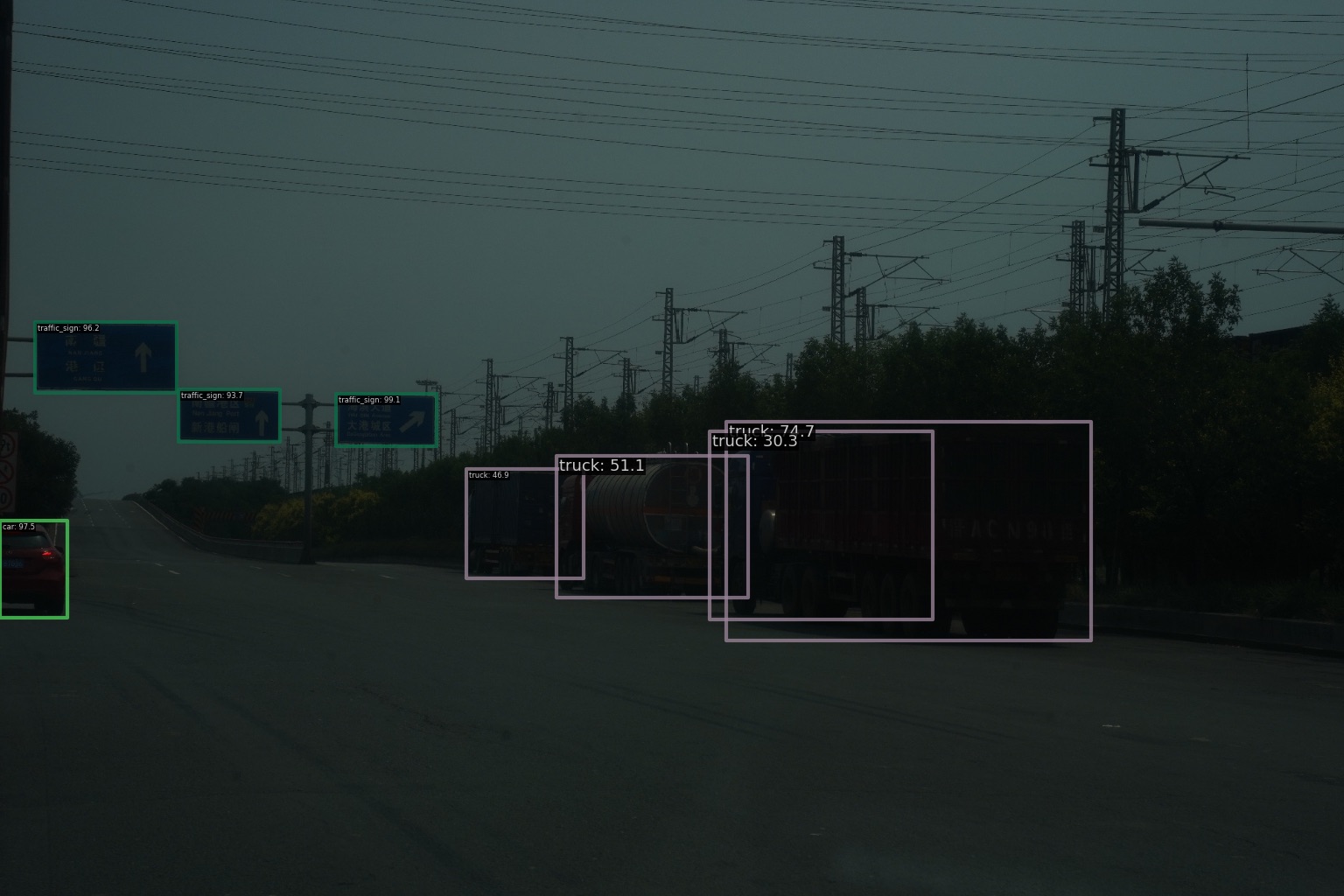}
    \end{minipage}
    \begin{minipage}[t]{0.135\textwidth}
        \centering
        \includegraphics[width=\textwidth]{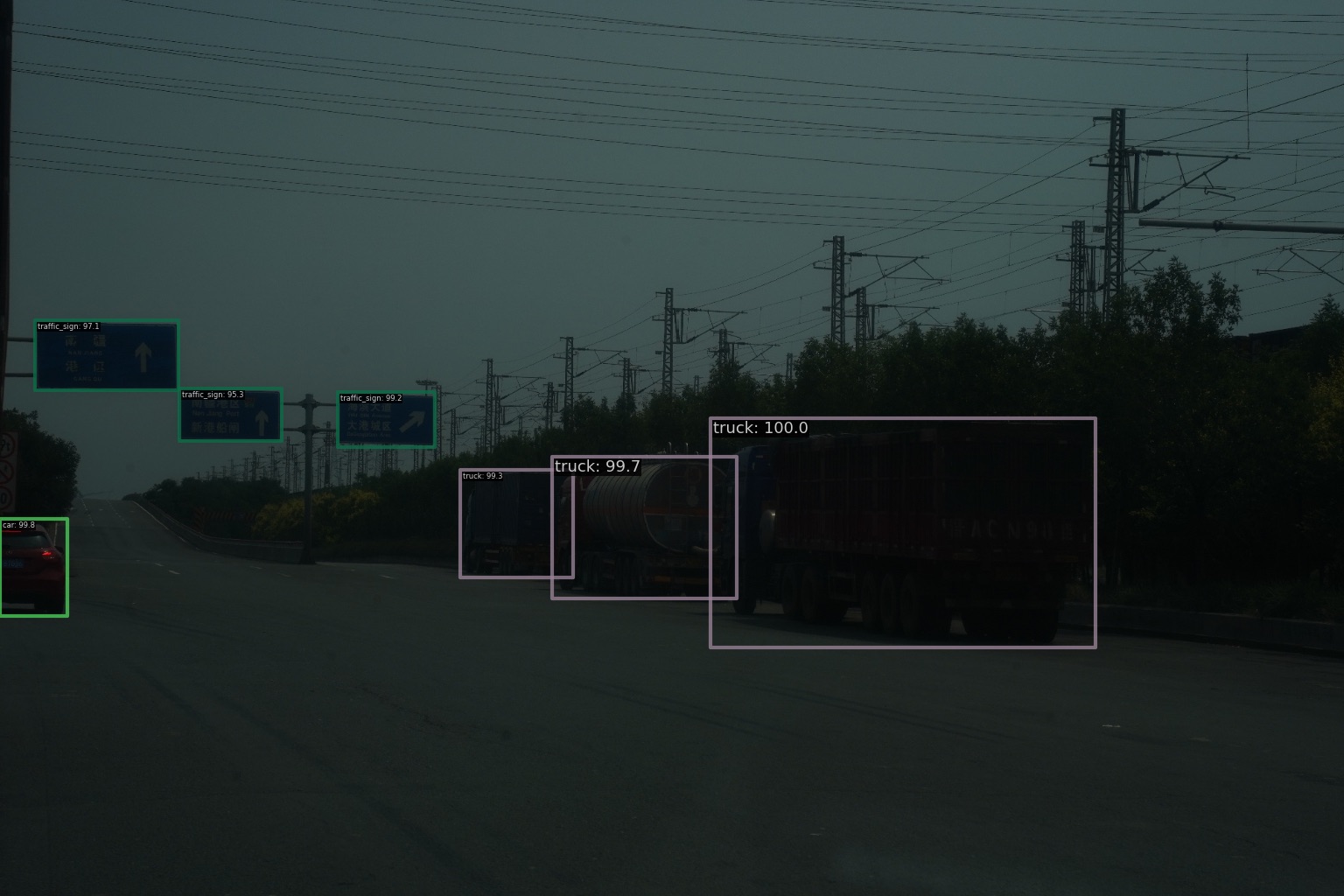}
    \end{minipage}
    \begin{minipage}[t]{0.135\textwidth}
        \centering
        \includegraphics[width=\textwidth]{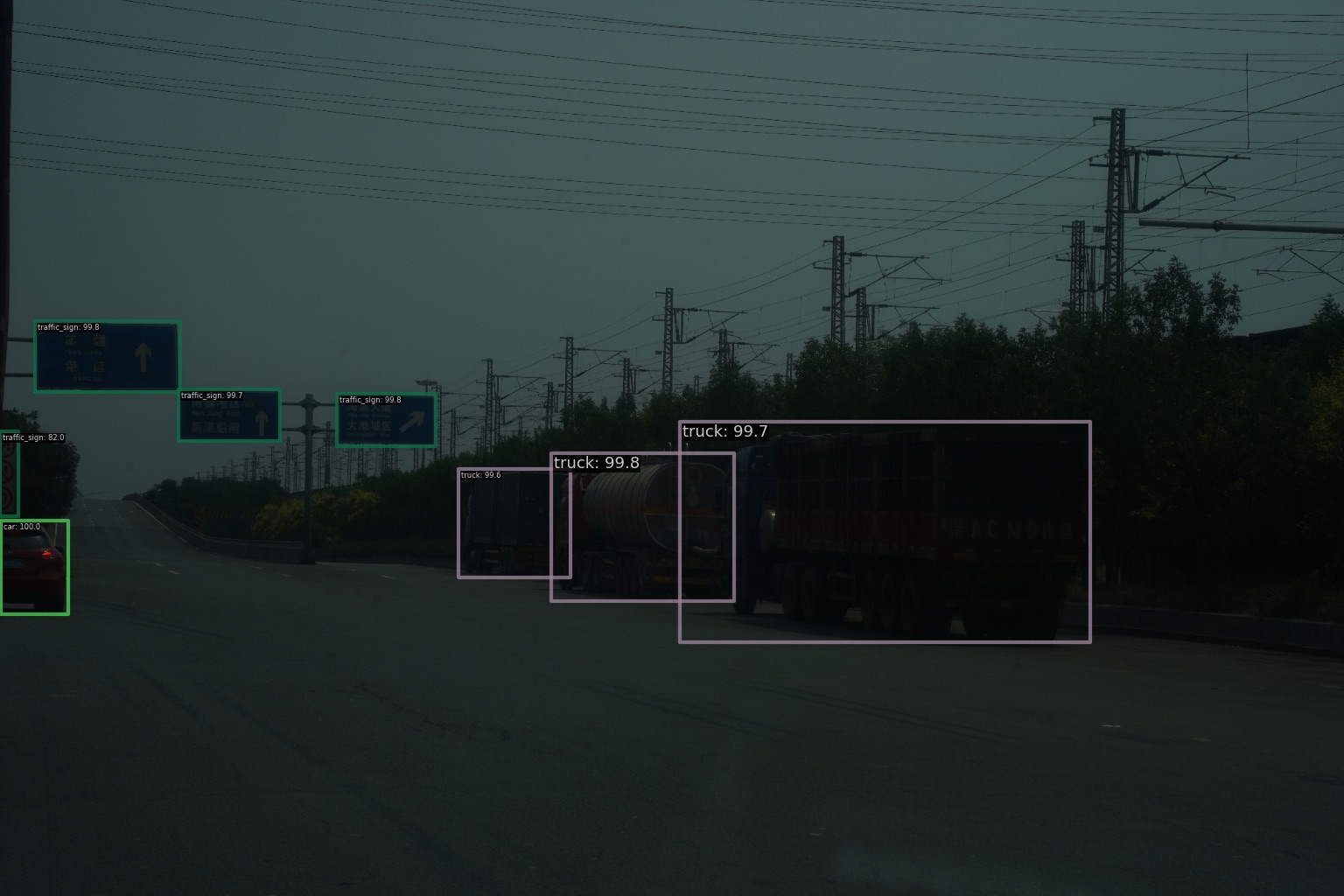}
    \end{minipage}
    \begin{minipage}[t]{0.135\textwidth}
        \centering
        \includegraphics[width=\textwidth]{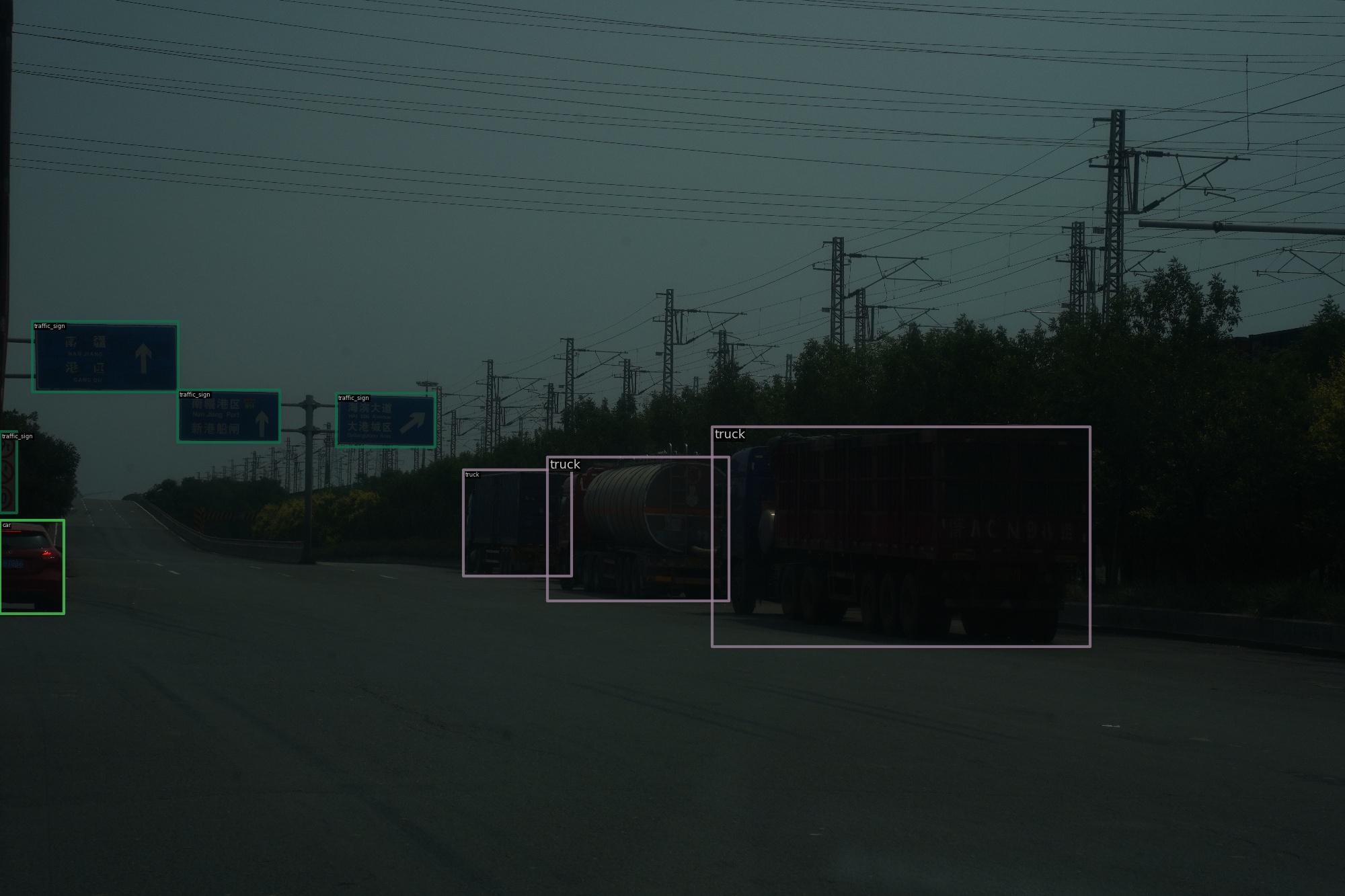}
    \end{minipage}

    \vspace{0.2em}

    \begin{minipage}[t]{0.01\textwidth}
        \centering
        \rotatebox{90}{\footnotesize Fog}
    \end{minipage}
    \begin{minipage}[t]{0.135\textwidth}
        \centering
        \includegraphics[width=\textwidth]{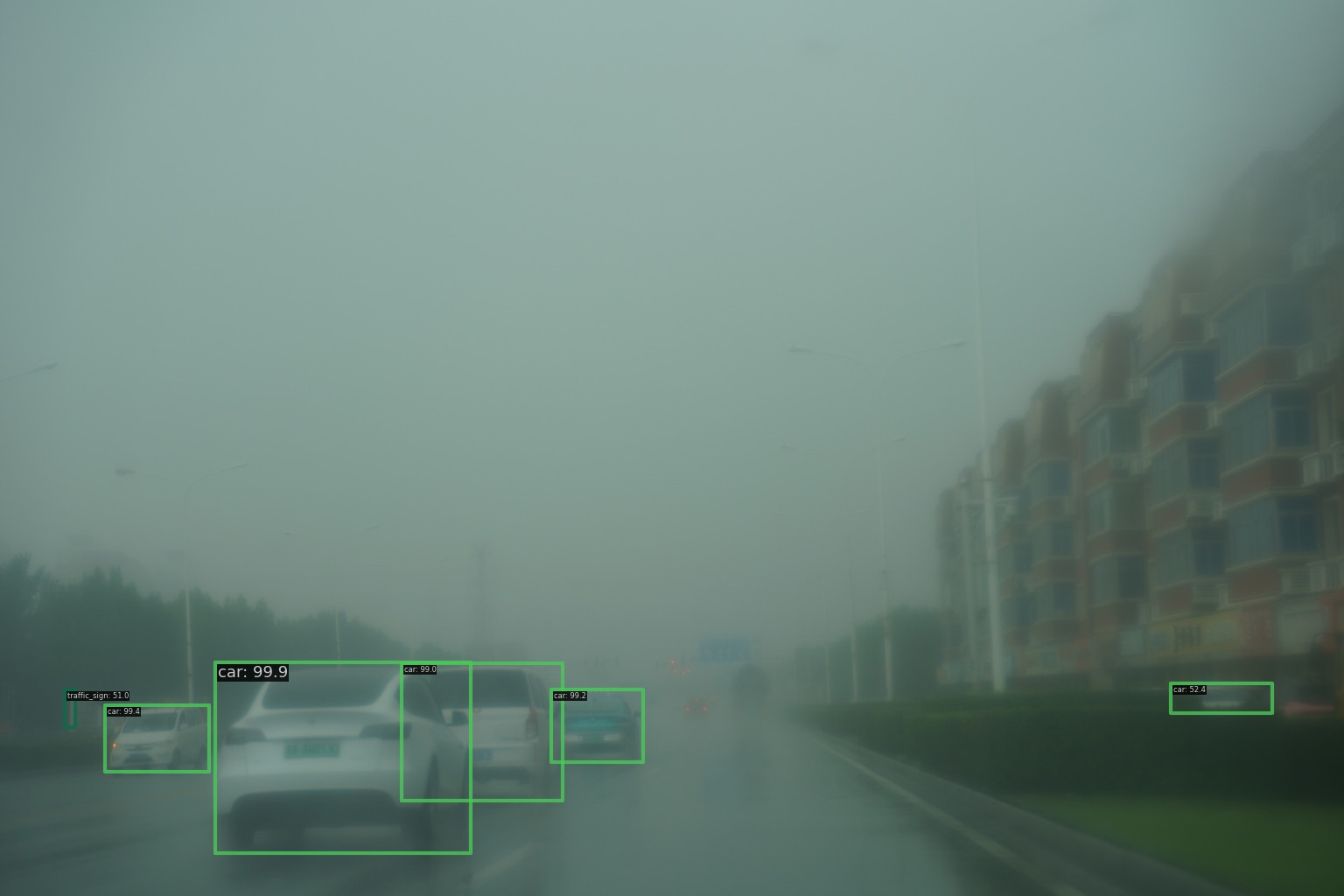}
    \end{minipage}
    \begin{minipage}[t]{0.135\textwidth}
        \centering
        \includegraphics[width=\textwidth]{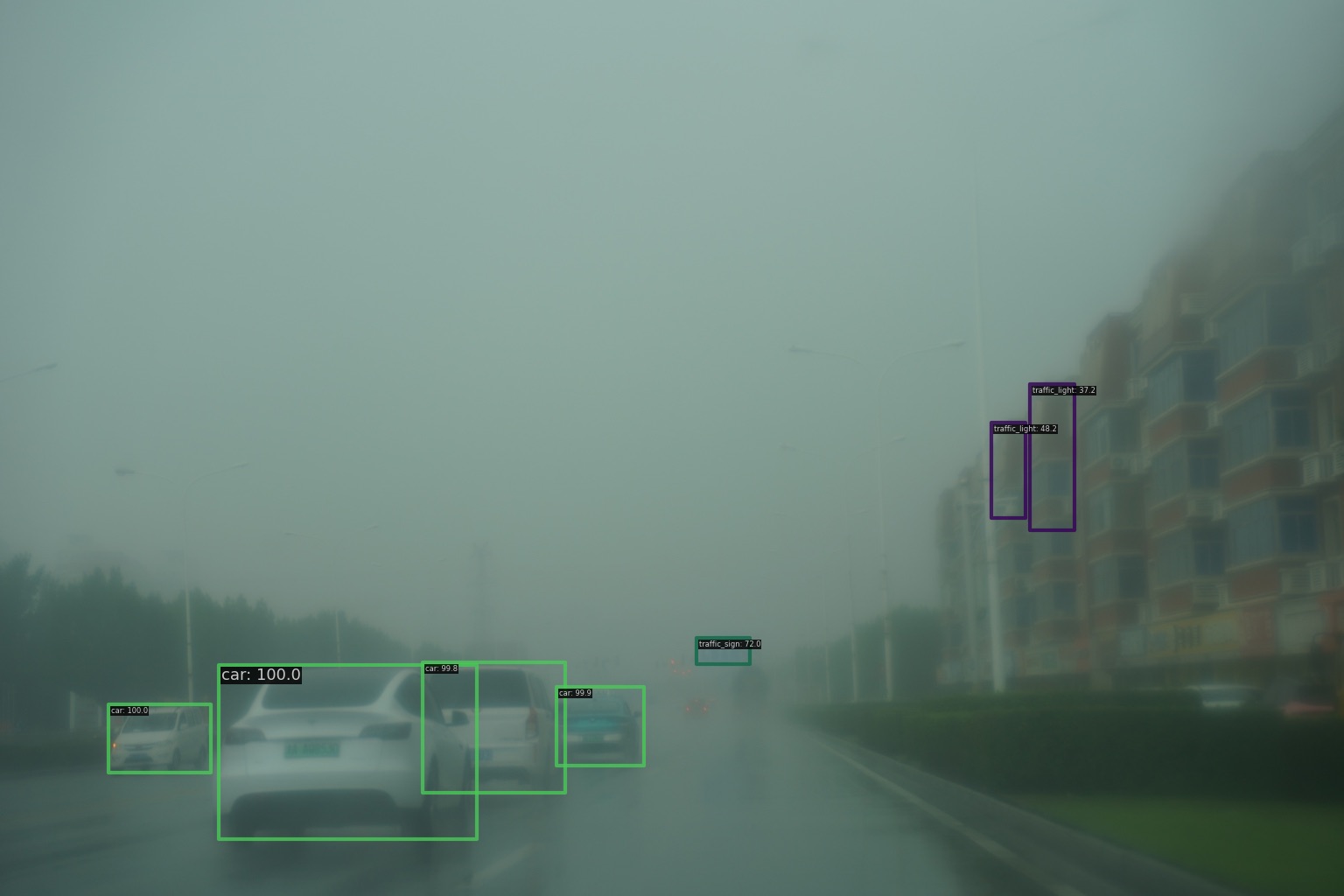}
    \end{minipage}
    \begin{minipage}[t]{0.135\textwidth}
        \centering
        \includegraphics[width=\textwidth]{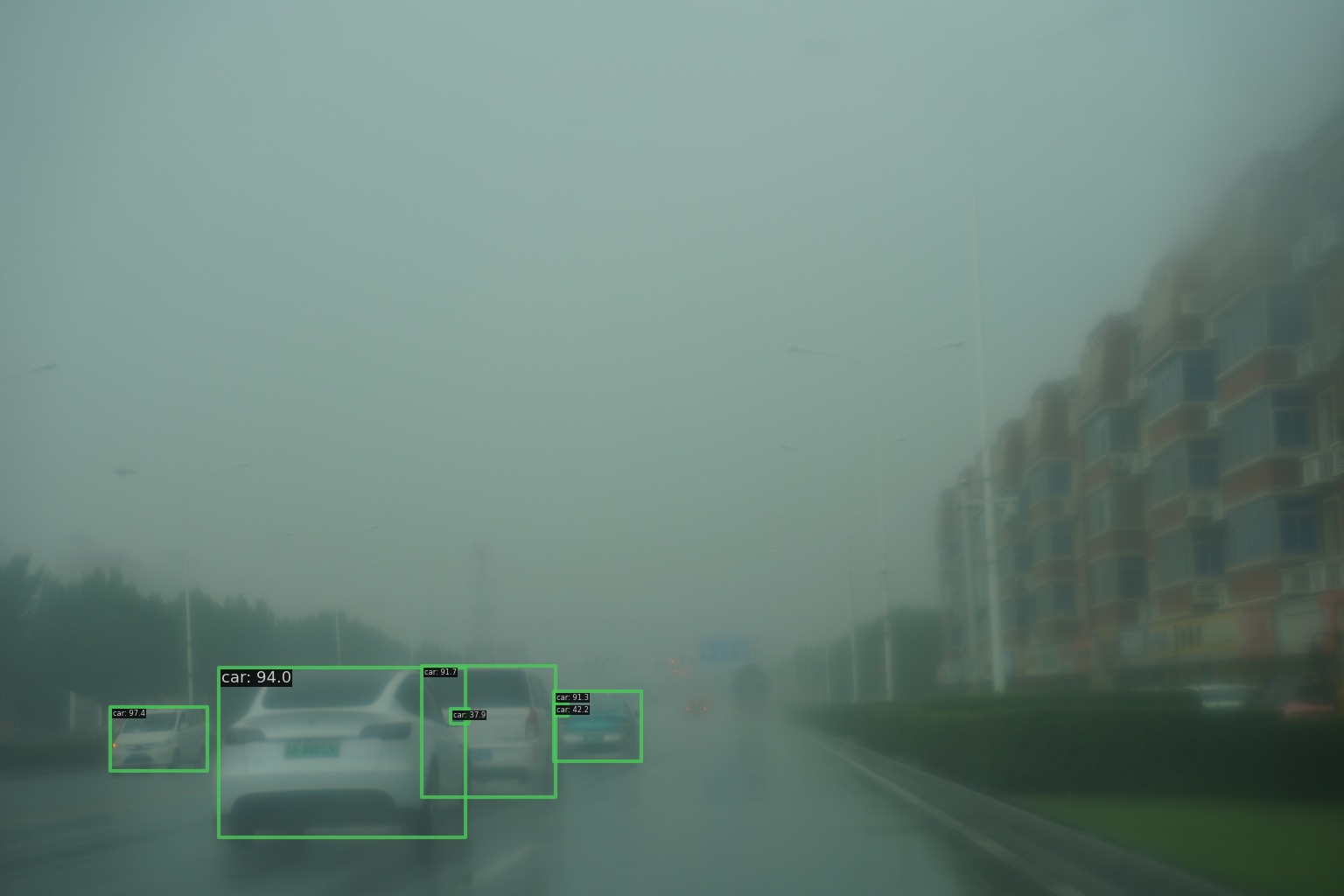}
    \end{minipage}
    \begin{minipage}[t]{0.135\textwidth}
        \centering
        \includegraphics[width=\textwidth]{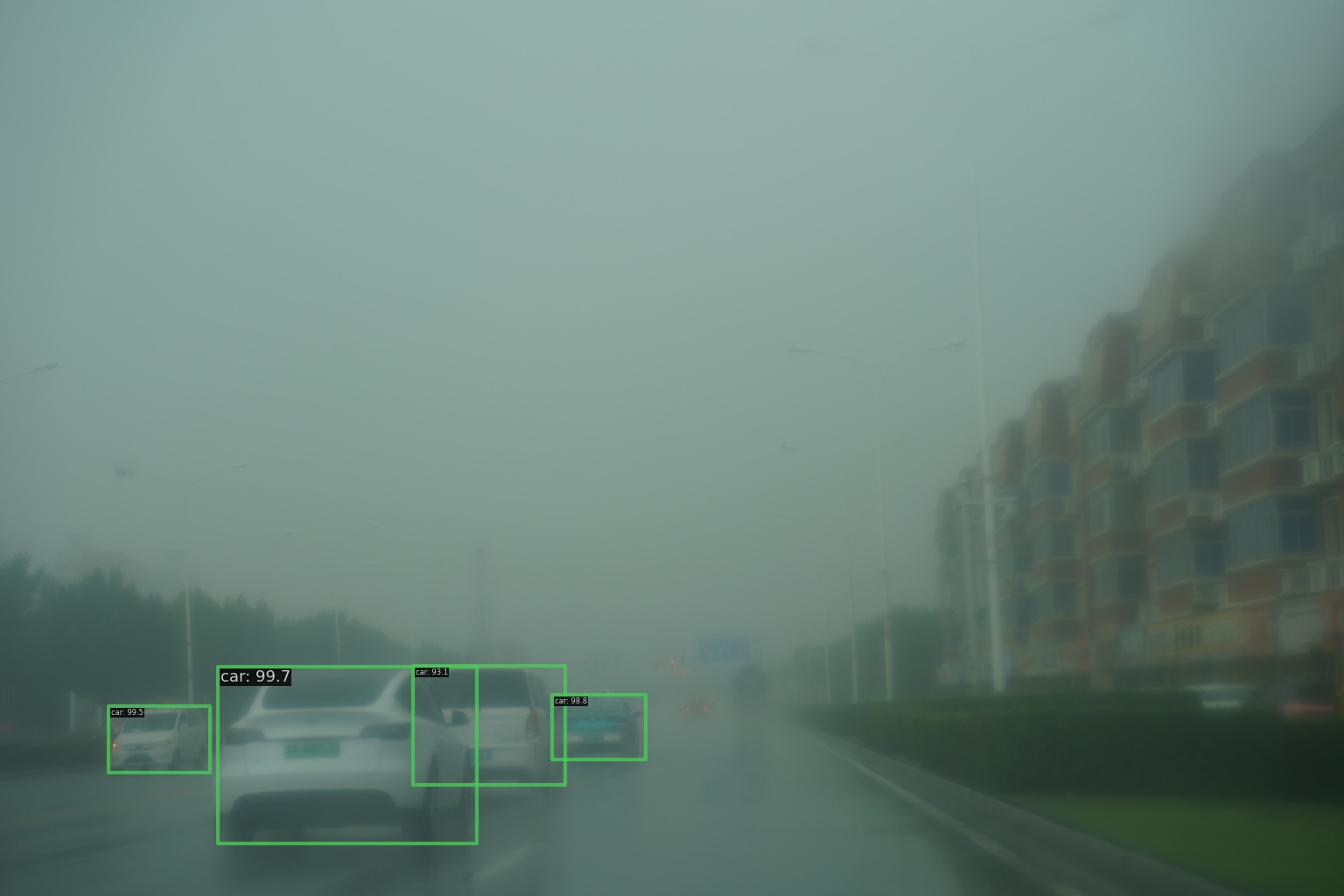}
    \end{minipage}
    \begin{minipage}[t]{0.135\textwidth}
        \centering
        \includegraphics[width=\textwidth]{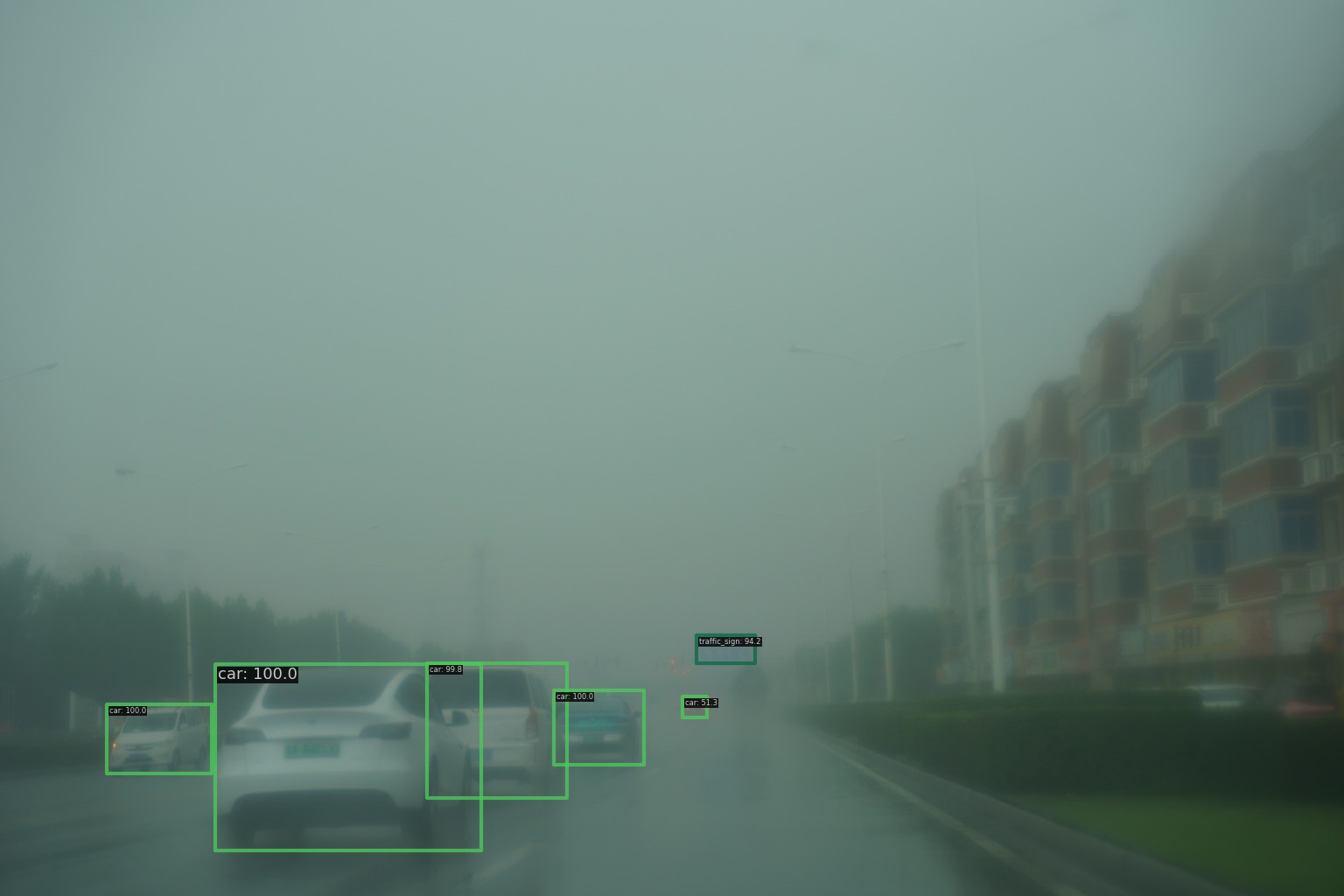}
    \end{minipage}
    \begin{minipage}[t]{0.135\textwidth}
        \centering
        \includegraphics[width=\textwidth]{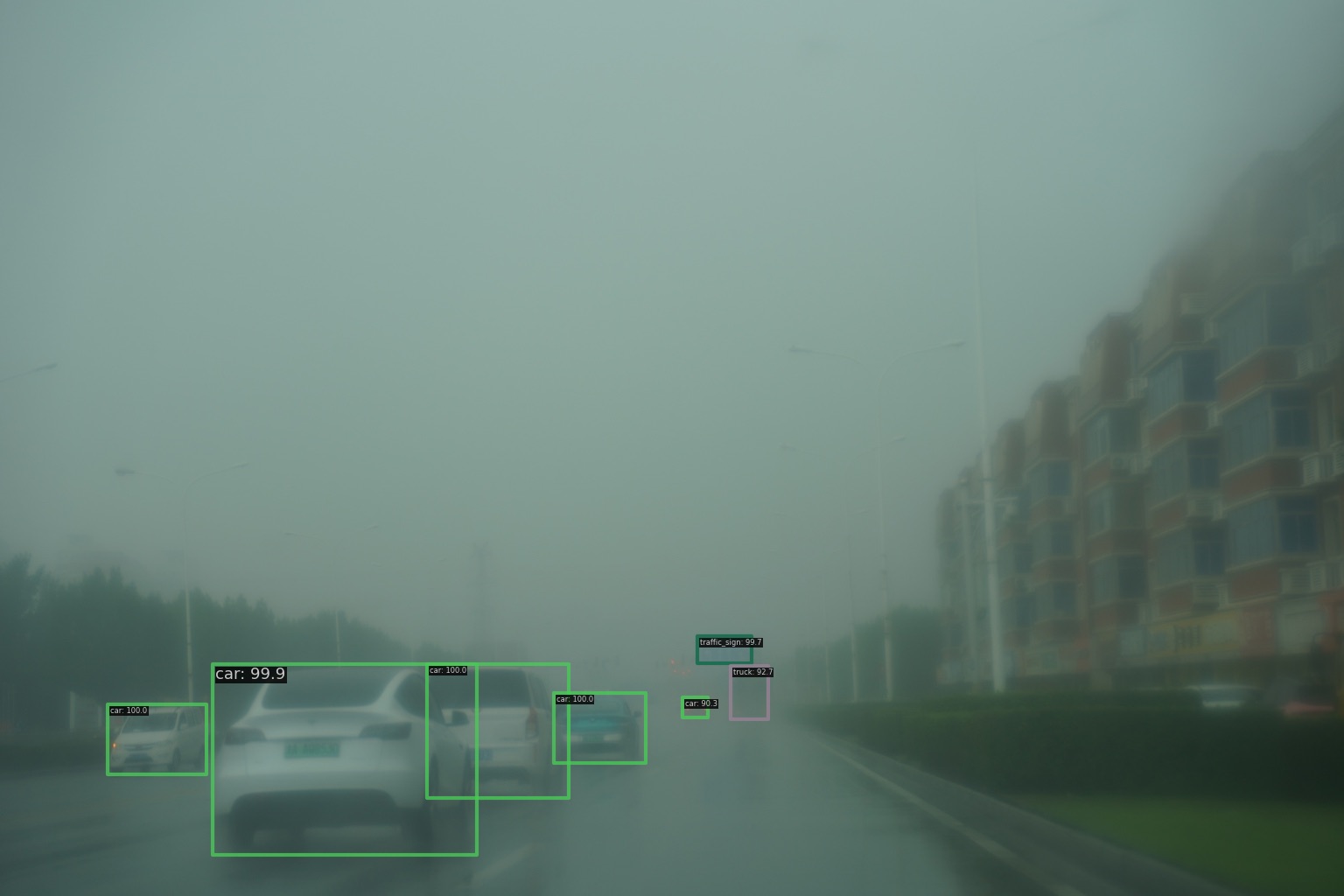}
    \end{minipage}
    \begin{minipage}[t]{0.135\textwidth}
        \centering
        \includegraphics[width=\textwidth]{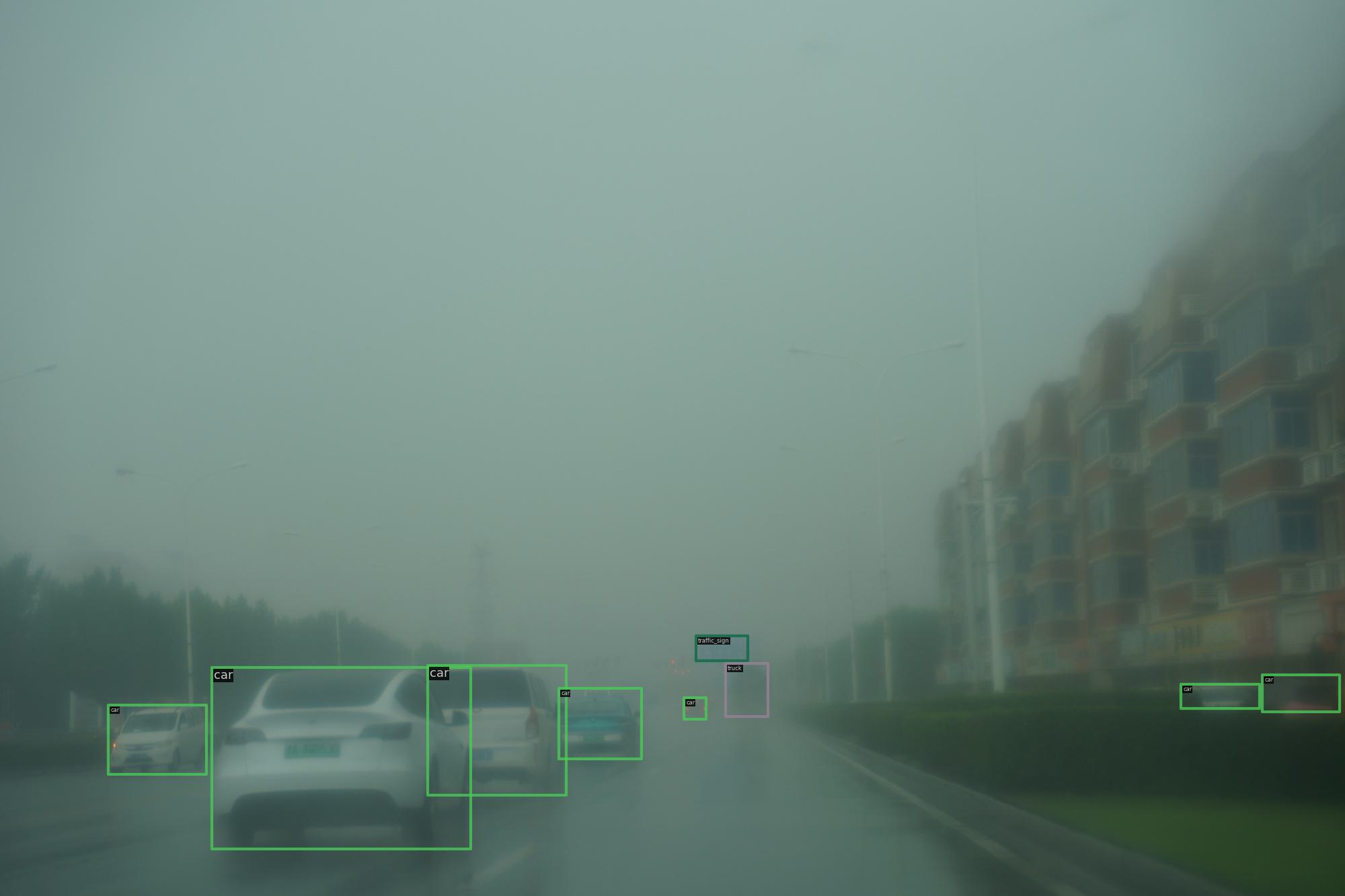}
    \end{minipage}

    \vspace{0.2em}

    \begin{minipage}[t]{0.01\textwidth}
        \centering
        \rotatebox{90}{\footnotesize Rain}
    \end{minipage}
    \begin{minipage}[t]{0.135\textwidth}
        \centering
        \includegraphics[width=\textwidth]{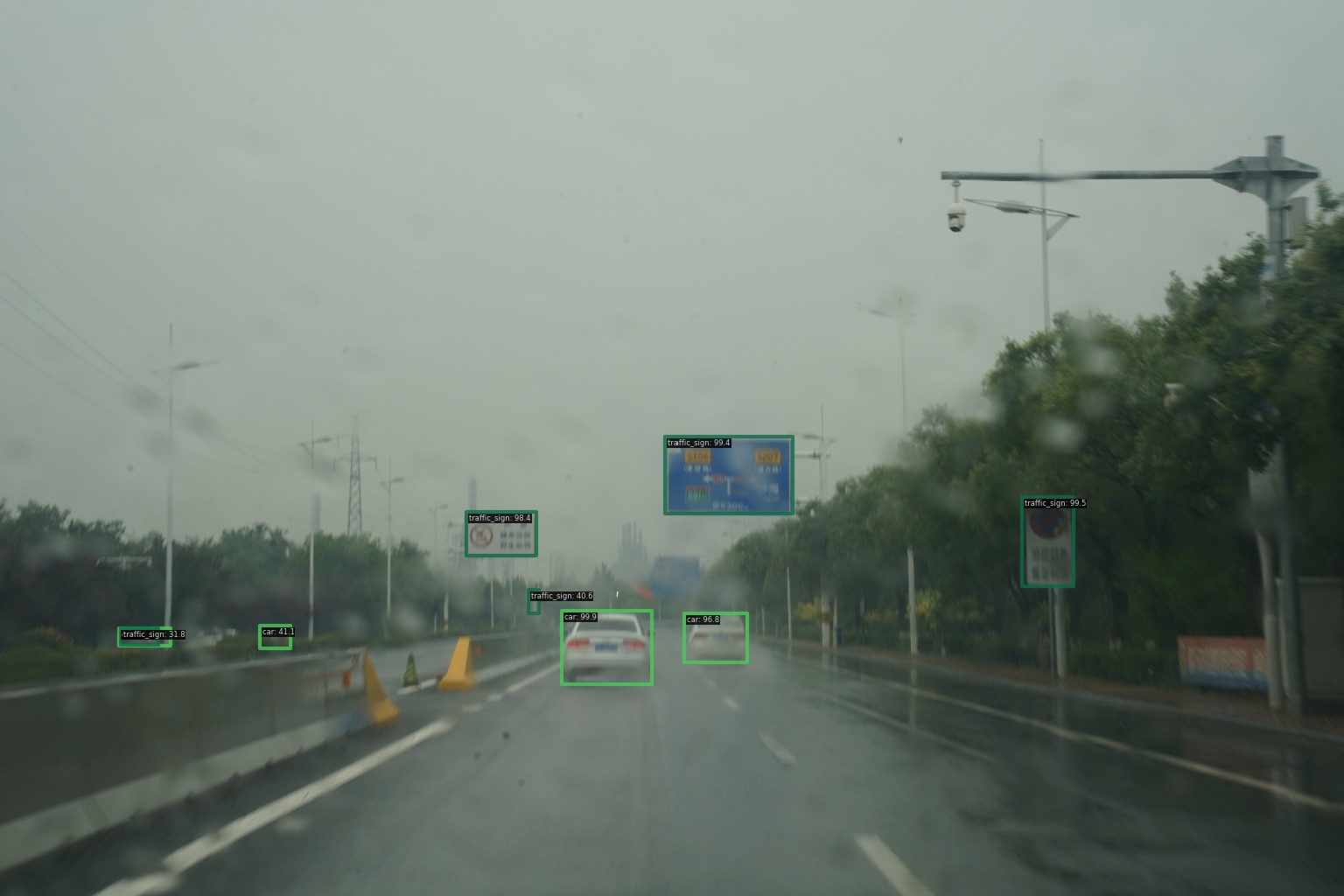}
        {\footnotesize (a) RAW}
    \end{minipage}
    \begin{minipage}[t]{0.135\textwidth}
        \centering
        \includegraphics[width=\textwidth]{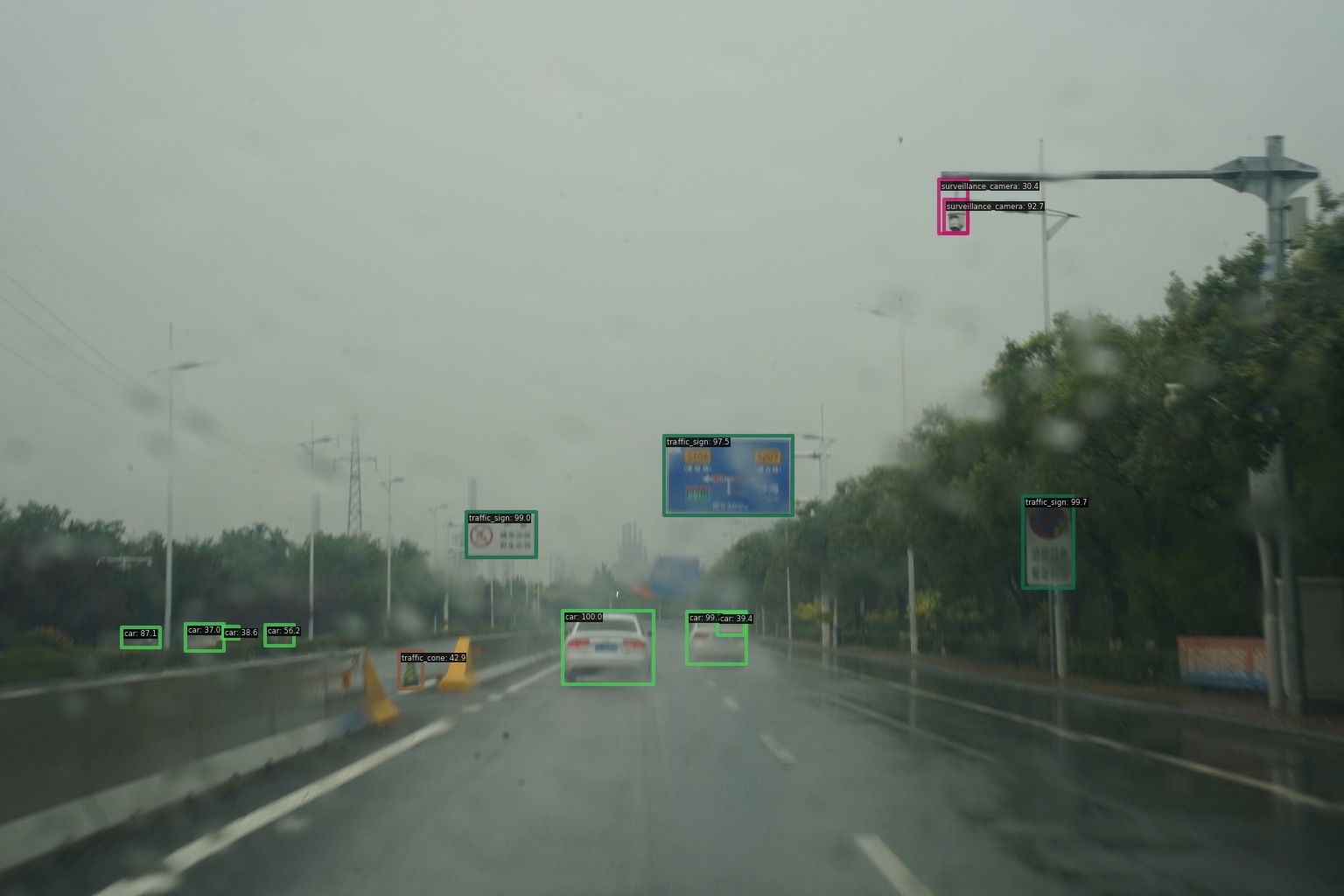}
        {\footnotesize (b) sRGB}
    \end{minipage}
    \begin{minipage}[t]{0.135\textwidth}
        \centering
        \includegraphics[width=\textwidth]{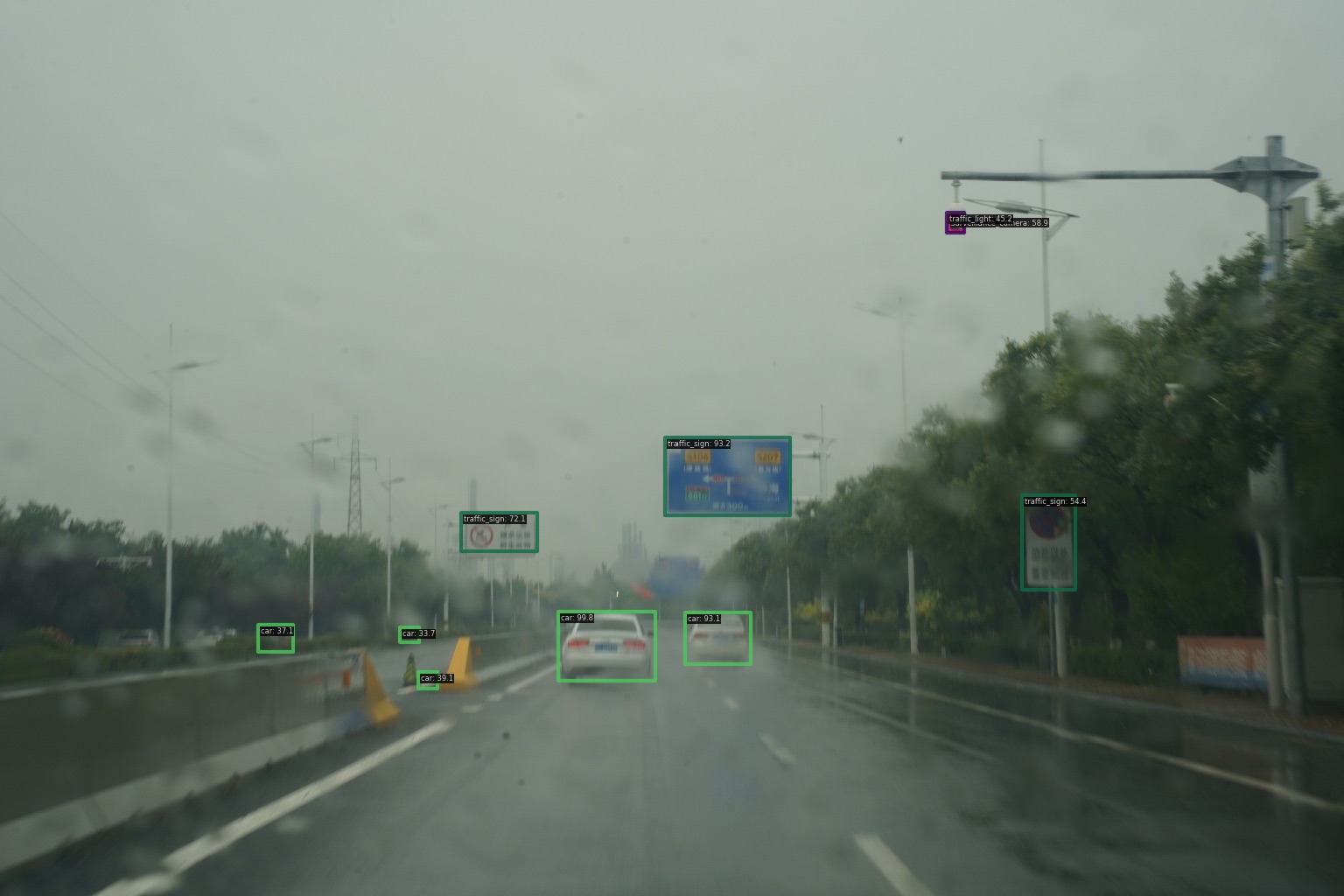}
        {\footnotesize (c) GenISP \cite{morawski2022genisp}}
    \end{minipage}
    \begin{minipage}[t]{0.135\textwidth}
        \centering
        \includegraphics[width=\textwidth]{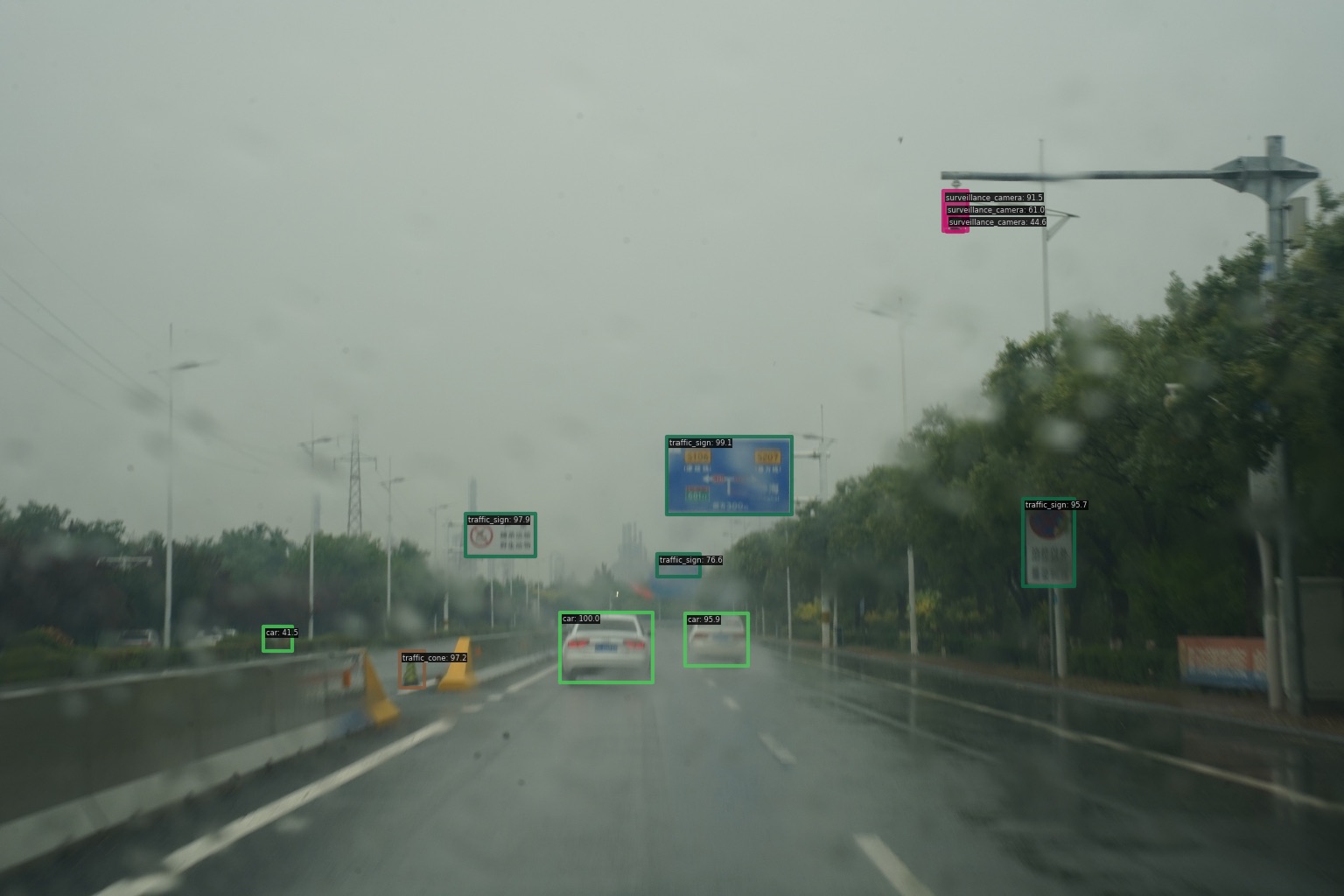}
        {\footnotesize (d) RAOD \cite{xu2023toward}}
    \end{minipage}
    \begin{minipage}[t]{0.135\textwidth}
        \centering
        \includegraphics[width=\textwidth]{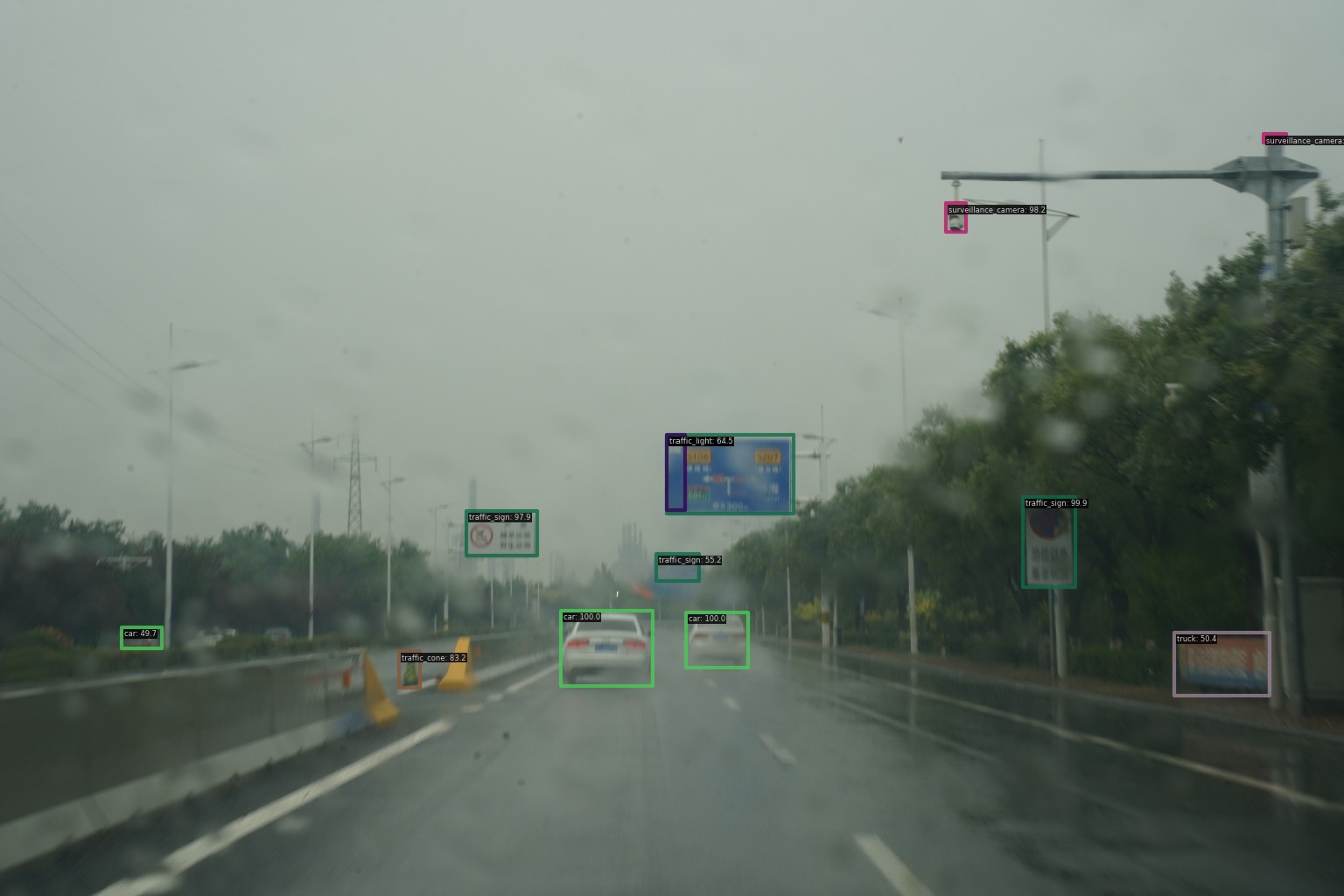}
        {\footnotesize (e) RAM \cite{gamrian2025beyond}}
    \end{minipage}
    \begin{minipage}[t]{0.135\textwidth}
        \centering
        \includegraphics[width=\textwidth]{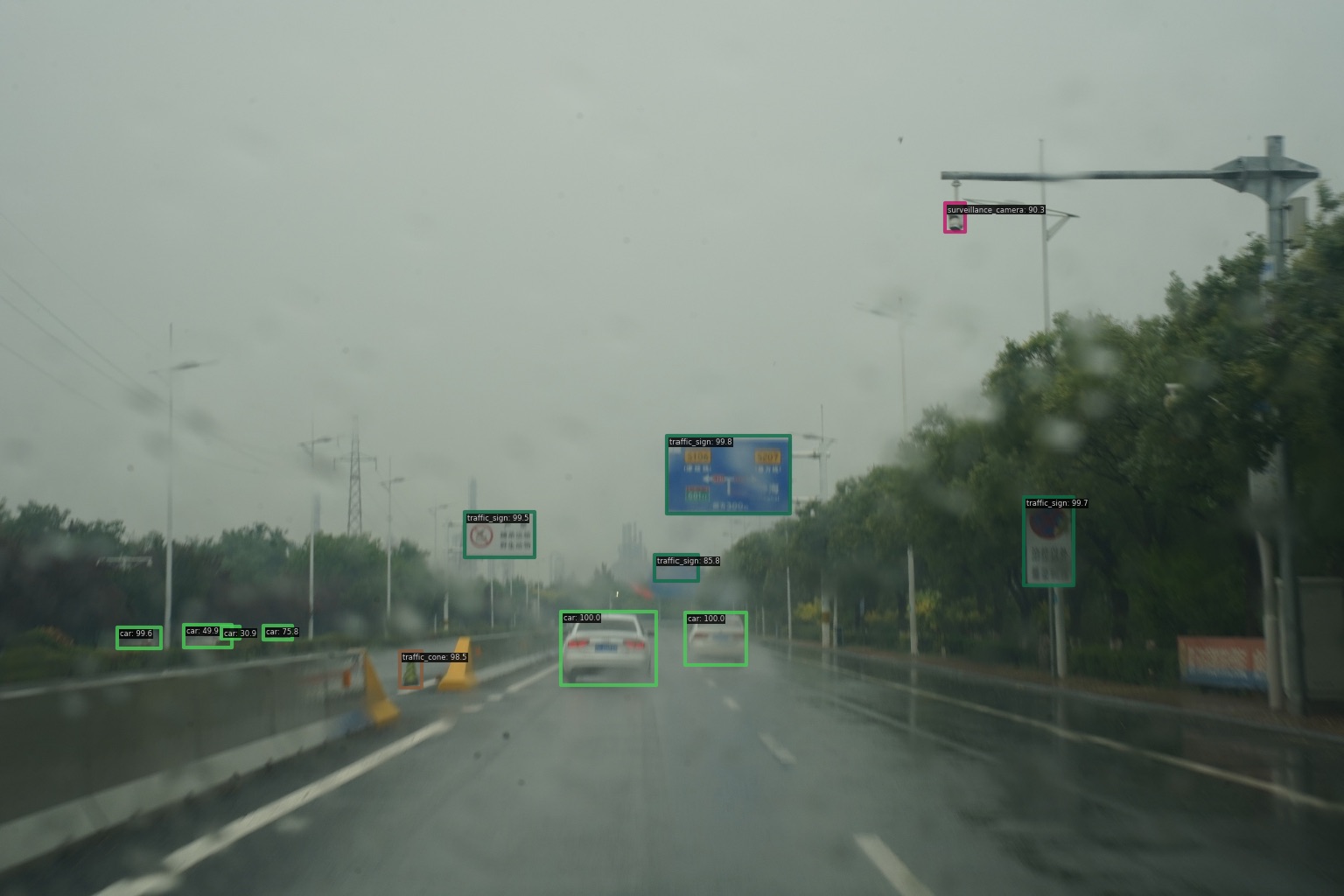}
        {\footnotesize (f) FreqAdapt}
    \end{minipage}
    \begin{minipage}[t]{0.135\textwidth}
        \centering
        \includegraphics[width=\textwidth]{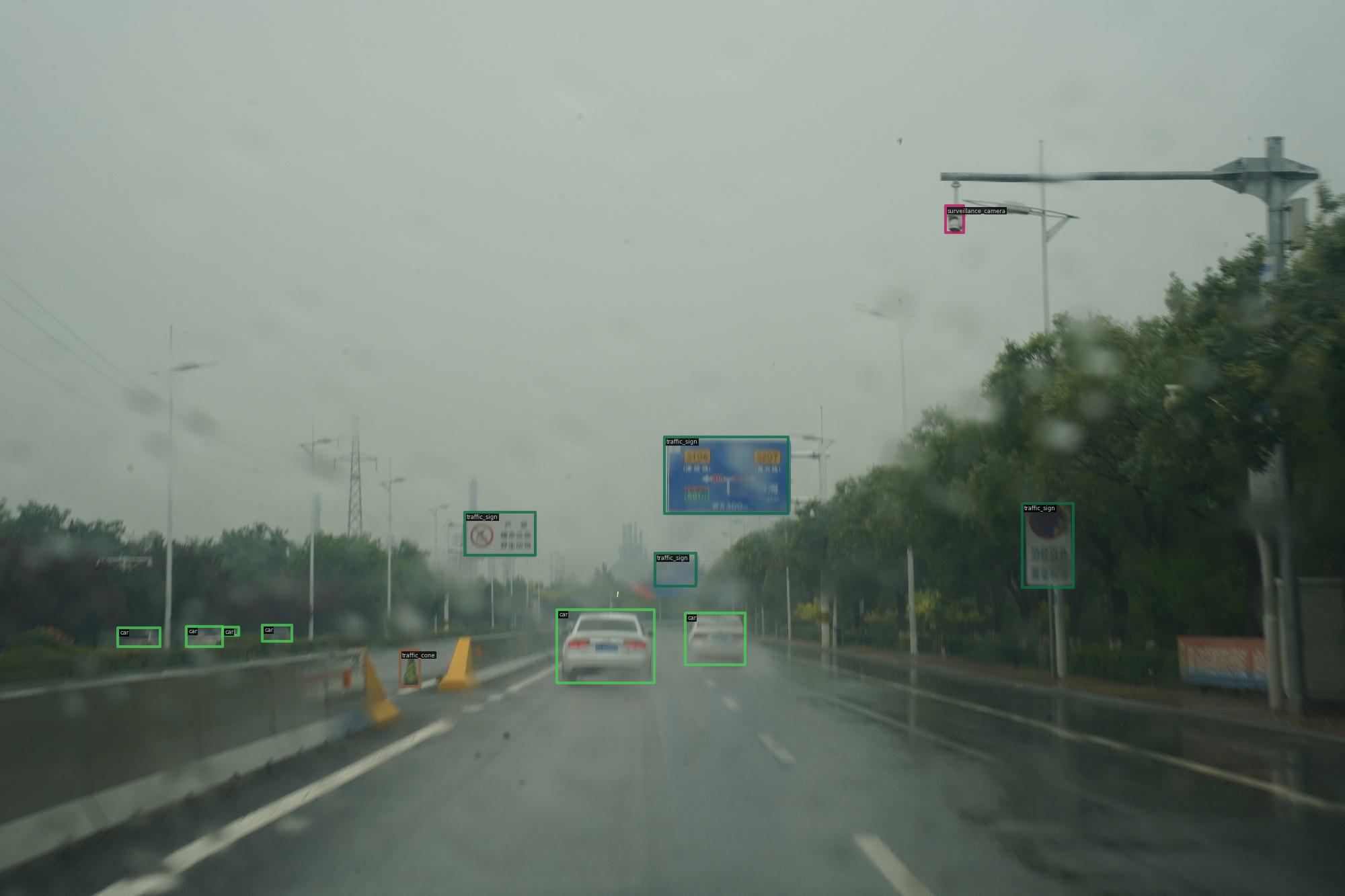}
        {\footnotesize (g) GT}
    \end{minipage}

    \caption{Qualitative comparison on AROD dataset under three challenging weather conditions. From top to bottom: Low-Night, Fog, and Rain scenarios.}

    \label{fig:object}
     \vspace{-4mm}
\end{figure*}





As shown in \cref{fig:object}, we present a comprehensive comparison of different methods under challenging weather conditions including low-light, fog, and rain scenarios. For better visualization, all images are displayed in RGB format, with detection results annotated by respective methods.

The first row illustrates nighttime low-light conditions featuring road infrastructure such as traffic signs and utility poles. The second row presents a foggy urban street scene where severely reduced visibility causes blurred building contours, leading to missed detections of distant objects by some methods. The third row depicts a rainy highway scenario where visual interference and reflections from rainwater result in varying degrees of false detections, producing numerous false positive bounding boxes by other methods.

The detection results clearly demonstrate that directly using sRGB images for detection suffers from severe performance degradation under these extreme weather conditions. While RAOD and RAM show some improvements, they still exhibit notable deficiencies in target localization accuracy and detection completeness. In contrast, our proposed FreqAdapt method achieves detection performance closer to the Ground Truth across all three adverse scenarios, not only reducing missed detections and false positives but also improving the localization accuracy of bounding boxes.

These visualization results comprehensively validate the robustness and effectiveness of FreqAdapt for object detection under diverse adverse weather conditions, demonstrating the crucial role of frequency-domain adaptive strategies in enhancing detection performance.


\subsection{Semantic Segmentation Evaluation}



To further validate the generalization capability of our proposed method across diverse perception tasks, we employed the ADE20K RAW dataset \cite{cui2025raw}, a synthetic variant derived from the original ADE20K dataset \cite{zhou2017scene}. We maintained the same training and testing splits as the original dataset.
For semantic segmentation experiments, we adopted SegFormer with the MIT-B0 backbone as our segmentation framework. We trained all models using the Adam optimizer with a batch size of 4. Training images were cropped to 512 × 512 pixels, and we performed 80,000 training iterations.

\begin{table}[ht]
\centering
\caption{Semantic segmentation performance on the ADE20K RAW dataset. We report mIoU under three exposure conditions using SegFormer with an MIT-B0 backbone.}
\label{tab:ADE20K}
\resizebox{\columnwidth}{!}{%
\begin{tabular}{l|ccc}
\toprule
\multirow{2}{*}{Method}
& \multicolumn{3}{c}{mIoU $\uparrow$} \\
& Dark & Normal & Over-exposed \\
\midrule
GenISP    & 20.09 & 28.71 & 26.26 \\
RAOD      & 19.39 & 28.07 & 26.40 \\
RAM       & 20.08 & 28.70 & 26.47 \\
FreqAdapt & \textbf{20.33} & \textbf{30.17} & \textbf{27.94} \\
\bottomrule
\end{tabular}%
}
\vspace{-2mm}
\end{table}


Table \ref{tab:ADE20K} presents the semantic segmentation results on the ADE20K RAW dataset under varying exposure conditions. Our proposed FreqAdapt method consistently outperforms all baseline approaches across all three exposure scenarios. The superior performance stems from our frequency-domain design: the amplitude branch handles brightness-related adjustments (white balance, gamma, noise reduction) by selectively modulating different frequency components, effectively enhancing dark regions while suppressing noise through frequency-aware thresholding. Meanwhile, the phase branch preserves structural information critical for segmentation through edge-aware sharpening and detail enhancement operations that work directly on frequency coefficients. The significant improvement under normal conditions demonstrates that frequency-domain processing better maintains semantic details compared to spatial-domain ISP methods.

\begin{figure*}[h]
    \centering
    \begin{minipage}[b]{0.16\textwidth}
        \centering
        \includegraphics[width=\textwidth]{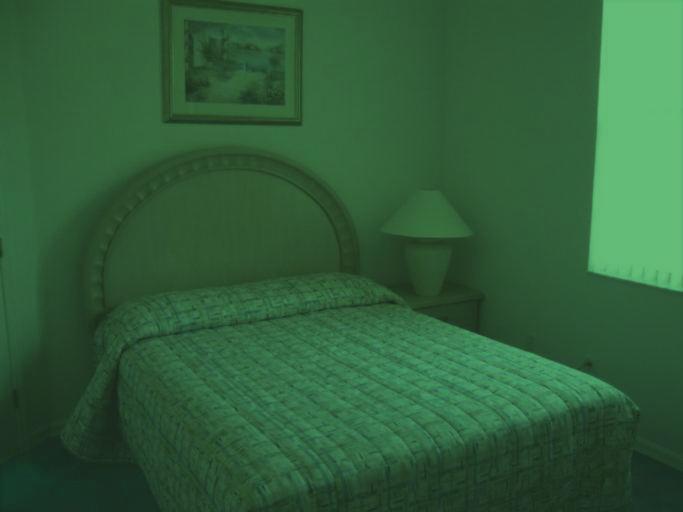}
        {\footnotesize (a) Input}
    \end{minipage}
    \hfill
    \begin{minipage}[b]{0.16\textwidth}
        \centering
        \includegraphics[width=\textwidth]{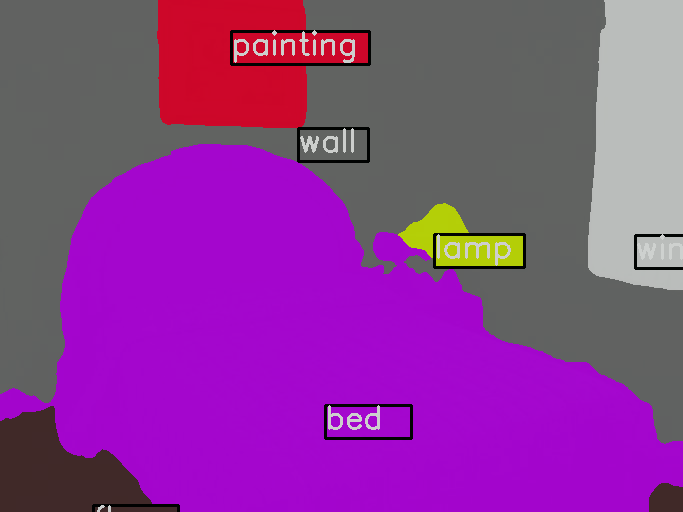}
        {\footnotesize (b) GenISP}
    \end{minipage}
    \hfill
    \begin{minipage}[b]{0.16\textwidth}
        \centering
        \includegraphics[width=\textwidth]{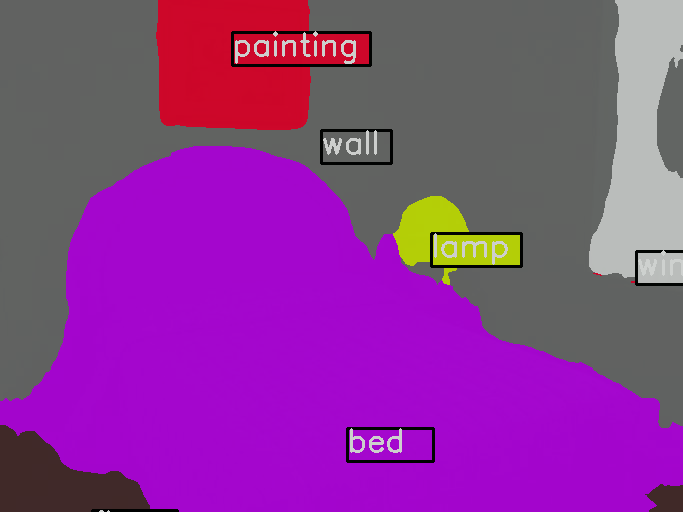}
        {\footnotesize (c) RAOD}
    \end{minipage}
    \hfill
    \begin{minipage}[b]{0.16\textwidth}
        \centering
        \includegraphics[width=\textwidth]{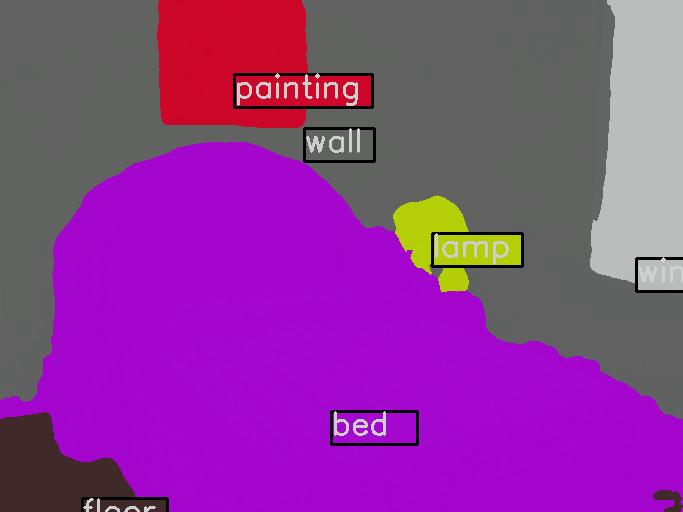}
        {\footnotesize (d) RAM}
    \end{minipage}
    \hfill
    \begin{minipage}[b]{0.16\textwidth}
        \centering
        \includegraphics[width=\textwidth]{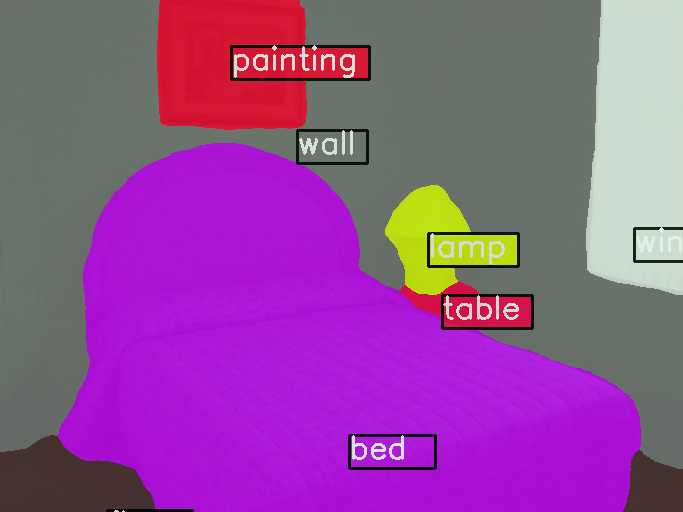}
        {\footnotesize (e) FreqAdapt}
    \end{minipage}
    \hfill
    \begin{minipage}[b]{0.16\textwidth}
        \centering
        \includegraphics[width=\textwidth]{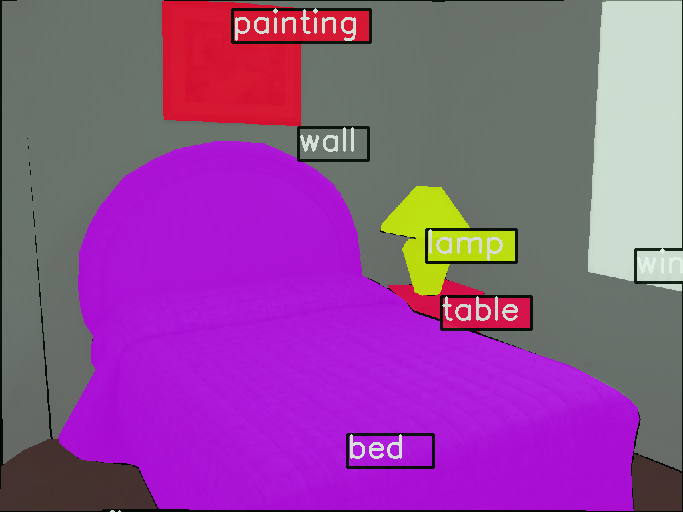}
        {\footnotesize (f) GT}
    \end{minipage}

    \caption{Visualization of semantic segmentation results on ADE20K RAW under normal exposure conditions.}
    \label{fig:segmentation}
\end{figure*}

\cref{fig:segmentation} shows semantic segmentation results on a normal-exposure bedroom scene from ADE20K RAW. FreqAdapt achieves notably more accurate segmentation compared to baselines, with clearer object boundaries and better semantic consistency, particularly visible in the wall-painting transitions and furniture edges. While GenISP, RAOD, and RAM exhibit misclassifications and blurred boundaries, FreqAdapt's segmentation closely matches the ground truth, demonstrating that frequency-domain processing better preserves semantic details during RAW-to-RGB conversion.


\subsection{FreqAdapt apply to RGB}


\begin{table}[t]
\centering
\caption{Comparison of FreqAdapt using RAW and sRGB inputs on the NOD-Nikon and NOD-Sony datasets. FreqAdapt achieves larger improvements with RAW inputs, which retain richer frequency information.}
\label{tab:freqadapt_comparison}
\resizebox{\columnwidth}{!}{%
\begin{tabular}{lcccc}
\toprule
\multirow{2}{*}{Method}
& \multicolumn{2}{c}{NOD-Nikon}
& \multicolumn{2}{c}{NOD-Sony} \\
\cmidrule(lr){2-3}
\cmidrule(lr){4-5}
& $\mathrm{mAP}$
& $\mathrm{mAP}_{50}$
& $\mathrm{mAP}$
& $\mathrm{mAP}_{50}$ \\
\midrule
RAW
& 26.3 & 49.7
& 25.8 & 50.6 \\
sRGB
& 26.6 & 51.5
& 25.5 & 50.3 \\
FreqAdapt (sRGB)
& 27.7 & 53.9
& 26.6 & 52.9 \\
FreqAdapt (RAW)
& \textbf{29.4} & \textbf{54.8}
& \textbf{27.3} & \textbf{53.7} \\
\bottomrule
\end{tabular}%
}
\end{table}

Table~\ref{tab:freqadapt_comparison} compares FreqAdapt's performance on RAW versus sRGB images. The significant performance gap (29.4 vs 27.7 mAP on NOD-Nikon) reveals a fundamental frequency-domain advantage of RAW data. RAW images preserve the complete frequency spectrum from the sensor, enabling our amplitude branch to perform frequency-selective adjustments (noise suppression, brightness modulation) and our phase branch to enhance structural details without interference from prior ISP processing. In contrast, sRGB images have undergone irreversible frequency modifications through demosaicing, tone mapping, and gamma correction, which compress dynamic range and alter phase relationships. These pre-existing frequency distortions limit FreqAdapt's ability to perform targeted frequency-domain operations. The 3.1 mAP improvement on RAW versus only 1.1 mAP on sRGB demonstrates that effective frequency-domain adaptation requires access to unprocessed frequency information, validating our design choice of operating directly on RAW data.

\section{Theoretical Basis for Amplitude-Phase Decoupling}
\label{sec::division}

The proposed dual-branch architecture is not a heuristic design but is deeply rooted in the physical and mathematical properties of the Fourier transform. In signal processing, the amplitude spectrum dictates the energy distribution and global statistics of a signal, whereas the phase spectrum encodes spatial locations, structural geometry, and inter-signal correlations. Based on these principles, we explicitly decouple the ISP pipeline into two specialized branches.

\textbf{Amplitude Domain for Intensity and Noise.} Operations governing global illumination, color temperature, and contrast---namely White Balance, Brightness Enhancement, and Gamma Correction---fundamentally alter the energy distribution of the image. For instance, brightness and white balance scale the DC and low-frequency energy, while Gamma correction redistributes energy across the spectrum. Therefore, they are naturally formulated as amplitude modulations. Furthermore, random sensor noise predominantly manifests as high-frequency energy perturbations. Isolating Noise Reduction in the amplitude domain allows for targeted energy suppression (via thresholding) without displacing the underlying spatial structures, which are protected in the phase domain.

\textbf{Phase Domain for Structure and Correlation.} Conversely, operations managing structural integrity and inter-channel dependencies are assigned to the Phase ISP Branch. 

First, the assignment of the Color Correction Matrix (CCM) to the phase domain is supported by cross-spectral analysis. Mathematically, inter-channel color relationships are encoded in the cross-power spectrum:
\begin{equation}
S_{RG}(\omega) = F_R(\omega) \cdot F_G^*(\omega) = A_R \cdot A_G \cdot e^{j(\phi_R - \phi_G)}
\end{equation}
where the relative phase term $(\phi_R - \phi_G)$ directly encodes the color correlation between channels. When spatial-domain CCM applies off-diagonal elements for color mixing, it induces coupled amplitude-phase variations in the frequency domain. Our approach leverages the key observation that color casts primarily correspond to systematic phase disparities among RGB channels. By adjusting the inter-channel phase relationships $\Delta\phi_{RG}$ and $\Delta\phi_{RB}$, we achieve accurate color correction via phase-induced channel realignment, instead of amplitude scaling or mixing. This mechanism is analogous to phase correlation for image registration, sharing a similar alignment objective through distinct mathematical pathways.

Second, the assignment of Edge Sharpening and Detail Enhancement to the phase domain is guided by the well-established theory of Phase Congruency \cite{morrone1987feature, kovesi1999image}. Traditional spatial sharpening amplifies high-frequency amplitudes, which inevitably exacerbates noise. However, human visual perception of edges and details corresponds precisely to spatial locations where the Fourier components are maximally in phase. By adjusting high-frequency phase components to align structural patterns, we achieve precise edge sharpening and detail enhancement without the noise-amplification artifacts inherent to amplitude manipulation. 

Through this principled decoupling, each ISP function is adaptively learned in its most mathematically and physically appropriate domain.

\section{ Limitations}
\label{sec::limit}

Despite its effectiveness, FreqAdapt presents certain limitations that warrant future investigation. 
First, regarding spatial flexibility, while the inherent global receptive field of the Fourier transform excels at holistic ISP adjustments, it may be less optimal for highly localized, spatially varying modulations required in scenes with complex, non-uniform lighting. 
Second, regarding training stability, the gradient flow through FFT/iFFT operations can be numerically unstable. This is particularly evident when processing RAW images with extreme dynamic ranges, where the massive energy disparity between the DC component and near-zero high-frequency components can lead to unstable gradients (e.g., singularities in phase calculation), necessitating rigorous learning rate scheduling and gradient clipping. 
Finally, regarding computational overhead and deployment, although element-wise spectral operations are efficient, the 2D FFT incurs an $\mathcal{O}(HW \log(HW))$ complexity and necessitates storing complex-valued representations, increasing the memory footprint. Furthermore, frequency-domain operations are highly sensitive to quantization errors, often demanding higher numerical precision (e.g., FP32). This poses potential challenges for deployment on edge devices strictly optimized for low-precision (e.g., INT8) arithmetic.

\section{Model architecture}

\cref{tab:freqadapt_config} presents the detailed layer configurations for FreqAdapt and its lightweight variant FreqAdapt-T. Both models share the same base channel dimension (nf=16) and core architectural design, with the primary difference lying in their ISP functionality coverage. 
FreqAdapt implements comprehensive ISP operations including seven functions: white balance, gamma correction, brightness adjustment, noise reduction in the amplitude branch, and color correction matrix, sharpening, detail enhancement in the phase branch. 
As shown in the \cref{tab:isp_ablation} ablation study, FreqAdapt-T streamlines the architecture by retaining only the essential ISP functions—white balance, gamma, brightness, and CCM—while removing computationally intensive operations such as noise reduction, sharpening, and detail enhancement. 
This selective removal of advanced ISP functions reduces the model complexity by approximately 35\% in terms of parameters while maintaining comparable performance on standard lighting conditions, making FreqAdapt-T suitable for real-time applications where computational efficiency is prioritized.

\section{Traditional Image Signal Processing Pipeline}

The conventional ISP pipeline transforms raw sensor data into RGB images through a series of specialized processing stages. We outline the key operations in their typical execution order:

\textbf{Raw Preprocessing.} Initial sensor corrections involve black level subtraction to remove dark current offset and establish the true zero-light reference point. Linearization maps ADC output values to a linear radiometric space, correcting any sensor non-linearities to ensure subsequent operations work with scene-linear intensity values.

\textbf{Defective Pixel Correction.} Malfunctioning sensor elements, including permanently dead pixels and temporarily hot pixels from thermal noise, are identified and corrected through interpolation from surrounding valid pixels, preventing these point defects from corrupting the final image.

\textbf{Demosaicing.} The sparse color sampling from CFA patterns (typically Bayer RGGB) is reconstructed into full-resolution RGB channels. This interpolation process must estimate missing color values at each pixel location while minimizing artifacts such as false colors and maintaining edge sharpness.

\textbf{Denoising.} Stochastic noise arising from photon statistics and electronic readout is suppressed through spatial or frequency filtering techniques. Modern approaches balance noise reduction with detail preservation using edge-aware filters or learned priors.

\textbf{White Balance.} Channel-specific gain adjustments compensate for the spectral distribution of the scene illuminant, ensuring achromatic objects appear neutral. These gains are determined through automatic algorithms, manual settings, or camera metadata.

\textbf{Color Correction.} 3×3 transformation matrix converts device-dependent RGB values to a standard color space, compensating for differences between the camera's spectral sensitivity functions and the target colorimetry.

\textbf{Tone Mapping.} High dynamic range scene radiance is compressed to match display capabilities through global or local operators that preserve perceptual contrast while avoiding saturation in highlights and shadows.

\textbf{Gamma Correction.} The sRGB or other appropriate transfer function is applied, converting linear intensities to a perceptually uniform encoding that optimizes quantization efficiency and ensures correct display reproduction.

\textbf{Sharpening and Detail Enhancement.} High-frequency components are selectively amplified to compensate for lens blur and demosaicing softness, typically using unsharp masking or gradient-based methods while avoiding overshoot artifacts.

\textbf{Compression and Encoding.} The processed image undergoes quantization to 8-bit depth and lossy compression (JPEG, HEIF) for storage efficiency, potentially including chroma subsampling and metadata preservation.

\begin{table*}[t]
\centering
\caption{Layer configurations of FreqAdapt and FreqAdapt-T. Each ConvBlock consists of Conv2d, BatchNorm2d, and ReLU layers.}
\label{tab:freqadapt_config}
\resizebox{0.95\textwidth}{!}{%
\begin{tabular}{c|c|l|c|c}
\hline
Module & Layer & Type & FreqAdapt & FreqAdapt-T \\
\hline

\multirow{7}{*}{Freq Encoder}
& 1 & Amplitude Conv
& $3{\rightarrow}16$, $3{\times}3$ kernel
& $3{\rightarrow}16$, $3{\times}3$ kernel \\
& 2 & Amplitude Pool
& $2{\times}2$ kernel, stride 2
& $2{\times}2$ kernel, stride 2 \\
& 3 & Phase Conv
& $3{\rightarrow}16$, $3{\times}3$ kernel
& $3{\rightarrow}16$, $3{\times}3$ kernel \\
& 4 & Phase Pool
& $2{\times}2$ kernel, stride 2
& $2{\times}2$ kernel, stride 2 \\
& 5 & RAW Statistics Conv
& $3{\rightarrow}8$, $3{\times}3$ kernel
& $3{\rightarrow}8$, $3{\times}3$ kernel \\
& 6 & Fusion Conv
& $40{\rightarrow}32$, $1{\times}1$ kernel
& $40{\rightarrow}32$, $1{\times}1$ kernel \\
& 7 & Adaptive Average Pooling
& $1{\times}1$ output
& $1{\times}1$ output \\
\hline

\multirow{6}{*}{Amplitude ISP}
& 1 & WB Predictor
& Linear: $32{\rightarrow}16{\rightarrow}3$
& Linear: $32{\rightarrow}16{\rightarrow}3$ \\
& 2 & Brightness Predictor
& Linear: $32{\rightarrow}16{\rightarrow}1$
& Linear: $32{\rightarrow}16{\rightarrow}1$ \\
& 3 & Gamma Predictor
& Linear: $32{\rightarrow}16{\rightarrow}1$
& Linear: $32{\rightarrow}16{\rightarrow}1$ \\
& 4 & Noise Reduction
& Linear + Conv layers
& -- \\
& 5 & Frequency Modulator
& Conv: $3{\rightarrow}16{\rightarrow}3$, $1{\times}1$
& Conv: $3{\rightarrow}16{\rightarrow}3$, $1{\times}1$ \\
& 6 & Local Adapter
& Conv: $3{\rightarrow}16{\rightarrow}3$, $3{\times}3$
& Conv: $3{\rightarrow}16{\rightarrow}3$, $3{\times}3$ \\
\hline

\multirow{5}{*}{Phase ISP}
& 1 & CCM Predictor
& Linear: $32{\rightarrow}16{\rightarrow}9$
& Linear: $32{\rightarrow}16{\rightarrow}9$ \\
& 2 & Sharpening
& Conv + edge detection
& -- \\
& 3 & Detail Enhancement
& Conv: $3{\rightarrow}16{\rightarrow}8{\rightarrow}3$
& -- \\
& 4 & Phase Adjuster
& Conv: $3{\rightarrow}16{\rightarrow}3$, $1{\times}1$
& Conv: $3{\rightarrow}16{\rightarrow}3$, $1{\times}1$ \\
& 5 & Edge Enhancer
& Conv: $3{\rightarrow}8{\rightarrow}1$, $3{\times}3$
& Conv: $3{\rightarrow}8{\rightarrow}1$, $3{\times}3$ \\
\hline

\multirow{3}{*}{Fusion}
& 1 & Weight Predictor
& Linear: $32{\rightarrow}16{\rightarrow}2$
& Linear: $32{\rightarrow}16{\rightarrow}2$ \\
& 2 & Attention
& Conv: $6{\rightarrow}16{\rightarrow}3$, $1{\times}1$
& Conv: $6{\rightarrow}16{\rightarrow}3$, $1{\times}1$ \\
& 3 & Final Refinement
& Conv: $3{\rightarrow}16{\rightarrow}3$, $3{\times}3$
& Conv: $3{\rightarrow}16{\rightarrow}3$, $3{\times}3$ \\
\hline
\end{tabular}%
}
\vspace{-2mm}
\end{table*}

\end{document}